\documentclass[10pt,twocolumn,letterpaper]{article}
\pdfoutput=1
\usepackage[pagenumbers]{cvpr}

\usepackage{cuted}
\usepackage{caption}
\usepackage{microtype}
\usepackage{mathtools}
\usepackage{bm}
\usepackage{multirow}
\usepackage{makecell}
\usepackage{colortbl}
\usepackage{algorithm}
\usepackage{algpseudocode}
\usepackage{amsthm}

\DeclareMathOperator*{\argmin}{arg\,min}
\DeclareMathOperator*{\argmax}{arg\,max}
\DeclareMathOperator{\Tr}{Tr}
\DeclareMathOperator{\diag}{diag}
\DeclareMathOperator{\IoU}{IoU}

\newtheorem{proposition}{Proposition}

\newcommand{\method}{\textnormal{\scshape IR-ODWSP}\xspace}
\newcommand{\methodDyn}{\textnormal{\scshape IR-ODWSP-Dyn}\xspace}
\newcommand{\methodSG}{\textnormal{\scshape IR-ODWSP-SG}\xspace}
\newcommand{\methodAuto}{\textnormal{\scshape IR-ODWSP-Auto}\xspace}
\newcommand{\methodplus}{\textnormal{\scshape IR-ODWSP+}\xspace}
\newcommand{\odwsp}{\textnormal{\scshape OD-WSP}\xspace}
\definecolor{rowgray}{gray}{0.95}
\definecolor{gainblue}{RGB}{0,90,160}
\definecolor{lossred}{RGB}{185,28,28}
\newcommand{\pos}[1]{\textcolor{gainblue}{\textbf{#1}}}
\newcommand{\negv}[1]{\textcolor{lossred}{#1}}

\definecolor{cvprblue}{rgb}{0.21,0.49,0.74}
\usepackage[pagebackref,breaklinks,colorlinks,allcolors=cvprblue]{hyperref}
\makeatletter\providecommand{\theHALG@line}{}\renewcommand{\theHALG@line}{\arabic{algorithm}.\arabic{ALG@line}}\makeatother

\def\confName{CVPR}
\def\confYear{2027}

\title{Reparameterizing Your Object Detectors for Weight Space Programming}

\author{Chien-Yao Wang\\
Institute of Information Science, Academia Sinica, Taiwan\\
{\tt\small kinyiu@iis.sinica.edu.tw}
}

\begin{document}
\maketitle
\begin{strip}
\centering
\vspace{-1.2em}
\includegraphics[width=0.995\linewidth]{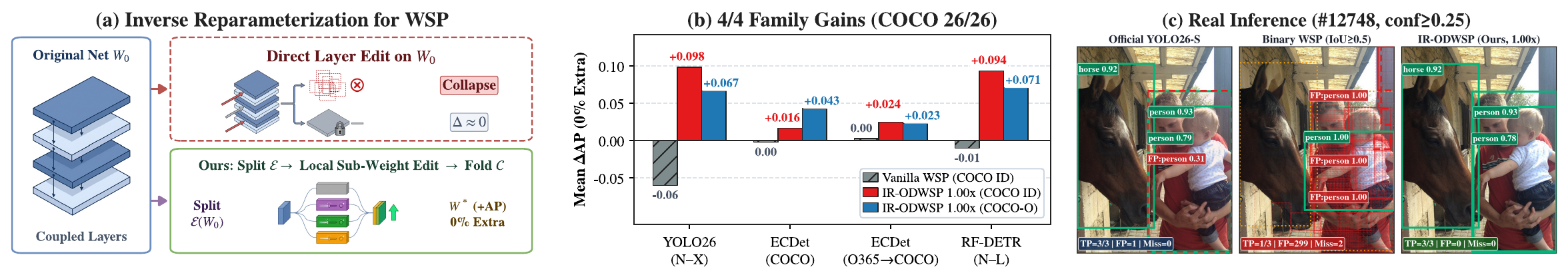}
\vspace{-0.5em}
\captionsetup{hypcap=false}
\captionof{figure}{\textbf{Inverse Reparameterization unlocks Weight-Space Programming (\method) for object detection.} \textbf{(a)} Direct editing of coupled detector weights $W_0$ collapses or stalls ($\Delta \approx 0$), whereas \method lifts $W_0$ via $\mathcal{E}$, applies closed-form local edits $\Delta\tilde{\theta}^\star$, and folds back via $\mathcal{C}$ into $W^\star$ at zero extra overhead. \textbf{(b)} At $1.00\times$ cost, \method improves all \textbf{26/26} MS~COCO~\cite{lin2014coco} settings and all \textbf{4/4} family means on zero-shot COCO-O~\cite{mao2023cocoo}. \textbf{(c)} Real YOLO26-S~\cite{jocher2026yolo26} inference (\texttt{val2017}): Official emits a duplicate ($\text{FP}=1$) and binary WSP~\cite{takehi2026sparks} collapses ($\text{FP}=299$), whereas \method achieves $\text{TP}=3/3, \text{FP}=0$.}
\captionsetup{hypcap=true}
\label{fig:teaser}
\vspace{-0.4em}
\end{strip}
\begin{abstract}
Weight-Space Programming (WSP) edits network weights without backpropagation, yet directly editing converged object detectors stalls or collapses because set prediction penalizes duplicate boxes as false positives while compact parameterizations trap weights in an isotropic Euclidean minimum. We propose Inverse-Reparameterized Object-Detection WSP (\method), lifting detector layers via exact function-preserving operators ($\mathcal{E}, \mathcal{C}$) into decoupled moment, 2D-DCT, PCA, and Stiefel-Procrustes subspaces. Solving winner-exclusive updates and folding back ($\Delta\text{Params}{=}0\%$, $\Delta\text{FLOPs}{=}0\%$) induces an anisotropic Riemannian metric, improving all 26/26 MS~COCO settings across 17 YOLO26, EdgeCrafter, and RF-DETR detectors (up to $+0.31$~$\text{AP}_s$), $+1.42$~mAP cross-basin COCO-O, zero-image repair ($+2.95$~mAP), and label-free weather adaptation ($+2.79$~AP). Without retraining, \method supports $-100\%$ to $+100\%$ latency budgets across bridged contraction, $1.00\times$ folding, dynamic (\methodDyn), and static (\methodSG) graphs: $-10\%$ retains $51.62$~AP / $33.48$~mAP, while positive budgets scale monotonically ($+0.30$~AP at $+10\%$, $+1.12$ on RF-DETR-N; $54.83$~AP at $+100\%$; $55.91$~AP / $35.44$~mAP in \methodplus). Source code is released at \url{https://github.com/WongKinYiu/ODWSP}.
\end{abstract}
\section{Introduction}
\label{sec:intro}

Deep neural network weights have historically been treated as opaque numerical artifacts produced exclusively through iterative stochastic gradient descent (SGD). Recently, \emph{Weight-Space Programming} (WSP)~\cite{takehi2026sparks} demonstrated an intriguing alternative: neural network parameters can be directly inspected, diagnosed, and algebraically edited to repair internal defects, steer behaviors, or adapt representations \emph{without backpropagation or even without training data}. Coupled with advances in task arithmetic~\cite{ilharco2023task}, model soups~\cite{wortsman2022modelsoups}, and closed-form model editing~\cite{meng2022rome}, gradient-free weight manipulation promises rapid post-training customization on edge hardware or restricted deployment pipelines where full autograd fine-tuning is prohibitively expensive or unavailable.

Despite promise across language models, control policies, and classifiers~\cite{takehi2026sparks,ilharco2023task,wortsman2022modelsoups}, naive WSP fails to improve---and frequently destroys---modern object detectors such as NMS-free YOLO26~\cite{jocher2026yolo26}, hybrid ViT-DETR EdgeCrafter~\cite{liu2026edgecrafter}, and DINOv2-backed RF-DETR~\cite{robinson2026rfdetr}. As illustrated in \cref{fig:teaser}(a), two fundamental obstacles arise when transferring standard WSP primitives to object detection:
\begin{enumerate}
    \item \textbf{Set-Prediction Duplicate Collapse:} Modern end-to-end detectors formulate detection as one-to-one set prediction~\cite{carion2020detr,wang2024yolov10}, where COCO Average Precision (AP) penalizes any secondary box on the same object as a false positive. Applying naive WSP closed-form regression with pooled binary targets ($\IoU \ge 0.50$) encourages neighboring anchors or queries to fire simultaneously (\eg, \texttt{val2017} \texttt{000000012748.jpg} at $\text{conf}\ge 0.25$ in \cref{fig:teaser}(c), where binary WSP explodes to $299$ duplicate FPs vs.\ \method's $\text{FP}=0$), collapsing YOLO26-S/M accuracy by $-39.7$ to $-43.7$~AP.
    \item \textbf{The Euclidean Stationary Trap:} Off-the-shelf detectors are already heavily optimized in a compact native parameter space. Isotropic spectral truncation or Euclidean ridge updates ($G = I_P$) treat high-SNR semantic directions, dominant coarse-stride activations, and low-variance small-object channels identically, yielding negative average returns ($-0.020$~AP across 17 models under Vanilla WSP).
\end{enumerate}

Addressing the first obstacle requires reformulating WSP around \emph{set-prediction axioms}: enforcing task-aligned top-1 winner exclusivity with sibling-ignore masking, Focal-curvature Hessians, soft quality targets ($q = \IoU^\gamma$), and $4\times 4$ box debiasing with supercategory shrinkage. Because deployed detectors routinely face style, sketch, and weather shifts~\cite{mao2023cocoo} where gradient fine-tuning is infeasible, detector WSP must also support zero-shot generalization and label-free adaptation directly via moment and frequency recalibration.

To escape compact Euclidean curvature, we introduce \textbf{Inverse-Reparameterized Object-Detection WSP (\method)}. Inverting structural reparameterization~\cite{ding2021repvgg,ding2021dbb,wang2024repvit} at post-training time, \method applies exact function-preserving lifts ($\mathcal{E}$) to split converged layers into decoupled moment, 2D-DCT frequency, PCA covariance, and Stiefel-Procrustes subspaces, solves closed-form set-prediction updates, and algebraically folds back ($\mathcal{C}$) with \textbf{zero extra parameters, FLOPs, or latency}. Crucially, \textbf{without retraining new weights}, \method allows practitioners to specify any \textbf{customizable latency budget} ($\beta \in [-100\%, +100\%]$, up to \methodplus) to determine the optimal zero-NMS deployed model: we co-design both a \textbf{Dual-Stream Fork-Join Dynamic-Graph algorithm (\methodDyn)} and a \textbf{Monolithic Single-Engine Static-Graph algorithm (\methodSG)} with custom Tensor-Core GEMM and SRAM warp-consensus \texttt{TensorRT~11.4} \texttt{IPluginV3} operators, strictly preserving official hardware speedup ratios ($\rho_b \ge 1 - \epsilon$ at $B{=}1,8$) and scaling monotonically across diverse architectures and scales.

Across 17 pretrained checkpoints (26 settings), $1.00\times$ \method achieves \textbf{26/26 (100\%)} positive blind MS~COCO gains (up to $+0.160$~AP and $+0.310$~$\text{AP}_s$), lifts zero-shot COCO-O on 20/26 settings (up to $+1.42$~mAP cross-basin), repairs corrupted weights with \emph{zero images} ($+2.95$~mAP), and gains $+2.50\text{--}+2.79$~AP under night-and-fog shift. Our contributions are:
\begin{itemize}
    \item \textbf{Set-Prediction WSP, Stationary Law \& Zero-Image / Label-Free Repair:} Explaining why binary WSP triggers duplicate-box avalanches on end-to-end detectors, and formulating winner-exclusive quality-aligned normal equations, $4\times 4$ box debiasing, an algebraic zero-parameter collision gate, zero-image weight repair ($+2.95$~mAP), and label-free domain adaptation (\cref{sec:problem,sec:method,sec:results}).
    \item \textbf{Inverse Reparameterization as Anisotropic Riemannian Preconditioning:} Designing exact function-preserving lifts and folds ($\mathcal{E}, \mathcal{C}$) across moment, 2D-DCT, PCA, and BoxAffine subspaces, and proving algebraically and via Haar-$O(P)$ controls ($< 2\times 10^{-9}$) that expand-edit-collapse induces a curvature-aligned metric $G_{\mathcal{E}} = (J_{\mathcal{C}}J_{\mathcal{C}}^\top)^{-1} \not\propto I_P$ (\cref{prop:riemannian}, \cref{sec:app_proofs}).
    \item \textbf{Cross-Basin Symmetry Law \& Large-Scale Fine-Tuning Drift:} Unifying single-model dormant-branch extraction and cross-checkpoint WSP via Stiefel Orthogonal Procrustes transport, uncovering severe continuous rotational basin drift under detector fine-tuning ($\cos=0.35$) and rescuing $-21.8\text{--}-52.4$~AP naive interpolation collapses (\cref{sec:method_cross,sec:app_ab5}).
    \item \textbf{Training-Free Customizable Budgeted Deployment (\methodDyn \& \methodSG):} Enabling practitioners to customize bidirectional latency budgets ($\beta \in [-100\%, +100\%]$ to \methodplus) to determine the optimal deployed inference model (closed-form bridged subgraph contraction, $1.00\times$ folded, dual-stream dynamic \methodDyn, or single-engine static \methodSG) \emph{without retraining new weights} or external NMS; evaluated across all 3 architectures and 17 scales, \method retains $51.62$~AP at $\le -10\%$ while scaling monotonically across positive budgets to $54.83$~AP at $\le +100\%$ ($3.74$~ms TRT $B{=}1$) and $55.91$~AP ($+2.14$~AP) in \methodplus (\cref{sec:method_graph,sec:results,sec:app_tta_audit,sec:app_neg_continuum}).
\end{itemize}
\section{Related Work}
\label{sec:related}

\paragraph{Weight-Space Programming and Gradient-Free Model Editing.}
Weight-Space Programming (WSP)~\cite{takehi2026sparks} recently introduced the paradigm of directly inspecting and algebraically editing network parameters without backpropagation. Related post-training works include associative weight insertion in language models (\eg, ROME~\cite{meng2022rome}), linear weight interpolation (\eg, WiSE-FT~\cite{wortsman2022wiseft}, Model Soups~\cite{wortsman2022modelsoups}, and Task Arithmetic~\cite{ilharco2023task}), and permutation/variance alignment (\eg, Git Re-Basin~\cite{ainsworth2023git} and REPAIR~\cite{jordan2023repair}), yet all target single-label or token prediction. Applied to modern NMS-free CNNs~\cite{wang2024yolov10,jocher2026yolo26} and query-based DETRs~\cite{carion2020detr,zhu2021deformabledetr,zhao2024rtdetr,peng2025dfine,liu2026edgecrafter,robinson2026rfdetr}, naive WSP faces three barriers: query competition triggers duplicate-box avalanches, compact heads sit in narrow Euclidean minima, and detector fine-tuning induces continuous orthogonal rotations. We resolve these via set-prediction WSP, Riemannian inverse reparameterization, Orthogonal Procrustes alignment~\cite{schonemann1966generalized}, and hardware-co-designed bidirectional budgeted deployment.

\paragraph{Structural Reparameterization.}
Structural reparameterization decouples training-time optimization geometry from inference-time execution graphs. Classic designs such as ACNet~\cite{ding2019acnet}, RepVGG~\cite{ding2021repvgg}, DBB~\cite{ding2021dbb}, OREPA~\cite{hu2022orepa}, and YOLOv7~\cite{wang2023yolov7} train parallel multi-scale and auxiliary branches with SGD before folding them into single operators, while RepLKNet~\cite{ding2022replknet}, MobileOne~\cite{vasu2023mobileone}, RepViT~\cite{wang2024repvit}, UniRepLKNet~\cite{ding2024unireplknet}, and YOLOv9 PGI~\cite{wang2024yolov9} extend this paradigm to ViT-style mobile CNNs, large kernels, and prunable supervision. Crucially, all prior methods operate in the \emph{forward training-to-inference direction} and require end-to-end gradient training. In contrast, we introduce \textbf{post-training inverse reparameterization} for gradient-free editing: starting from a converged single-branch detector, we algebraically invert compact layers into exact function-preserving factorizations that decouple normalization moments~\cite{li2018adabn}, 2D spatial frequencies, PCA covariances, and cross-basin subspaces, enabling targeted closed-form WSP updates before folding back with zero overhead.
\section{Problem Formulation}
\label{sec:problem}

\subsection{Mapping Weight-Space Programming to Object Detection}
\label{sec:problem_wsp}

Let $f(\cdot\,; W_0): \mathcal{X} \to \mathcal{Y}_{\text{set}}$ denote a pretrained detector with native weights $W_0 \in \mathcal{W} \subseteq \mathbb{R}^P$. For an input image $x \in \mathcal{X}$, $f$ emits $M$ candidate predictions
\begin{equation}
    f(x; W) = \big\{ (\hat{\bm{p}}_i(x; W),\, \hat{\bm{b}}_i(x; W)) \big\}_{i=1}^M,
    \label{eq:det_set_out}
\end{equation}
where $\hat{\bm{p}}_i = \sigma(\bm{z}_i) \in (0,1)^C$ are class probabilities from terminal features $\bm{h}_i \in \mathbb{R}^d$ via $\bm{z}_i = W_{\text{cls}}\bm{h}_i + \bm{b}_{\text{cls}}$, and $\hat{\bm{b}}_i \in \mathbb{R}^4$ are decoded boxes parameterized by $(W_{\text{box}}, \bm{b}_{\text{box}})$. Here $M$ indexes multi-stride anchors ($P_3$--$P_5$) in NMS-free CNNs~\cite{jocher2026yolo26} or object queries across $L$ decoder layers in DETRs~\cite{liu2026edgecrafter,robinson2026rfdetr}.

\paragraph{Zero-Gradient Detector WSP.}
Given $W_0$, an optional auxiliary prior $W_{\text{aux}}$ (\eg, Objects365~\cite{shao2019objects365} pretraining, or $\varnothing$ in the single-model setting), and a forward-only stream $\mathcal{D}_{\text{cal}}$ of size $N \ge 0$ (including data-free $N=0$ and unlabeled target stream $\mathcal{D}_{\text{target}}^u$), a WSP operator $\mathcal{A}_{\text{WSP}}$ computes $W^\star = \mathcal{A}_{\text{WSP}}(W_0, W_{\text{aux}}, \mathcal{D}_{\text{cal}})$ under strict zero gradients ($\nabla_W \mathcal{L} \equiv \bm{0}$) to maximize set-level Average Precision across ID and OOD test distributions $\mathcal{P}_{\text{test}}$:
\begin{equation}
    \max_{W \in \mathcal{W}} \; \mathbb{E}_{(x, \mathcal{G}) \sim \mathcal{P}_{\text{test}}} \Big[ \mathrm{AP}_{0.50:0.95}\big(f(x; W),\, \mathcal{G}\big) \Big],
    \label{eq:wsp_objdet_goal}
\end{equation}
with ground-truth set $\mathcal{G} = \{(c_g, \bm{b}_g)\}_{g=1}^{|\mathcal{G}|}$. Unlike per-sample classification loss $\sum_i \ell(\hat{\bm{p}}_i, y_i)$, $\mathrm{AP}$ in \cref{eq:wsp_objdet_goal} is \emph{non-decomposable across queries}: any secondary candidate overlapping $\bm{b}_g$ ($\IoU \ge \tau$) is penalized as a false positive, requiring the WSP surrogate $\widehat{\mathcal{L}}_{\text{OD}}(W)$ to enforce one-to-one winner exclusivity and continuous localization quality.

\subsection{Mapping Inverse Reparameterization to Decoupled Editable Spaces}
\label{sec:problem_ir}

Optimizing $\widehat{\mathcal{L}}_{\text{OD}}(W_0 + \Delta W) + \frac{\lambda}{2}\|\Delta W\|_2^2$ directly in compact space $\mathcal{W}$ confines updates to an isotropic Euclidean ball, entangling illumination moments, spatial frequencies, high-variance background channels, and low-variance small-object cues. To decouple these factors at zero deployment cost, we define an \emph{exact inverse-reparameterization pair} $(\mathcal{E}, \mathcal{C})$:
\begin{equation}
    \mathcal{E}: \mathcal{W} \to \widetilde{\Theta} \subseteq \mathbb{R}^{\widetilde{P}} \;\; (\widetilde{P} \ge P), \qquad
    \mathcal{C}: \widetilde{\Theta} \to \mathcal{W},
    \label{eq:ir_pair}
\end{equation}
satisfying function-preserving axioms for all $x \in \mathcal{X}$:
\begin{align}
    \text{(Pre-Edit)} \quad & \tilde{\theta}_0 = \mathcal{E}(W_0) \implies \tilde{f}(x; \tilde{\theta}_0) \equiv f(x; W_0), \label{eq:ax_pre}\\
    \text{(Post-Edit)} \quad & \forall \tilde{\theta} \in \widetilde{\Theta}, \;\; f\big(x; \mathcal{C}(\tilde{\theta})\big) \equiv \tilde{f}(x; \tilde{\theta}), \label{eq:ax_post}
\end{align}
with $\mathcal{C}(\mathcal{E}(W_0)) = W_0$, where $\tilde{\theta} = (\tilde{\theta}_{\text{mom}}, \tilde{\theta}_{\text{freq}}, \tilde{\theta}_{\text{cov}}, \tilde{\theta}_{\text{proc}}, \tilde{\theta}_{\text{box}})$ separates statistical moments, spatial frequencies, principal covariances, orthogonal auxiliary bases, and box affine heads. \Cref{prop:riemannian} establishes why expand-edit-collapse outperforms native Euclidean editing (proof in \cref{sec:app_proofs}).

\begin{proposition}[Inverse Reparameterization as Riemannian Metric Preconditioning]
\label{prop:riemannian}
Let $\mathcal{C}(\tilde{\theta}_0 + \Delta\tilde{\theta}) = W_0 + J_{\mathcal{C}} \Delta\tilde{\theta}$ (exact for affine $\mathcal{C}$, or to first order) have full-row-rank Jacobian $J_{\mathcal{C}} \in \mathbb{R}^{P \times \widetilde{P}}$ ($\mathrm{rank}(J_{\mathcal{C}})=P$), and let $\lambda > 0$. Solving $\Delta\tilde{\theta}^\star = \argmin_{\Delta\tilde{\theta}} \big\{ \widehat{\mathcal{L}}_{\mathrm{OD}}(W_0 + J_{\mathcal{C}}\Delta\tilde{\theta}) + \frac{\lambda}{2}\|\Delta\tilde{\theta}\|_2^2 \big\}$ and folding $\Delta W^\star = J_{\mathcal{C}}\Delta\tilde{\theta}^\star$ is algebraically identical to solving a \textbf{Riemannian-preconditioned} problem in $\mathcal{W}$ with $G_{\mathcal{E}} = (J_{\mathcal{C}} J_{\mathcal{C}}^\top)^{-1} \succ 0$:
\begin{equation}
    \Delta W^\star = \argmin_{\Delta W \in \mathcal{W}} \left\{ \widehat{\mathcal{L}}_{\mathrm{OD}}(W_0 + \Delta W) + \frac{\lambda}{2} \Delta W^\top G_{\mathcal{E}} \Delta W \right\}.
    \label{eq:riemannian_metric}
\end{equation}
Whenever $J_{\mathcal{C}}J_{\mathcal{C}}^\top = I_P$ (\eg, $J_{\mathcal{C}} \in O(P)$), $G_{\mathcal{E}} = I_P$ and $\Delta W^\star$ is unchanged; across $\lambda > 0$, gains require \textbf{anisotropic} curvature $G_{\mathcal{E}} \not\propto I_P$~\cite{amari1998natural}.
\end{proposition}

\paragraph{Latency-Budget-Aware Deployment Without Retraining.}
Beyond $\beta = 0\%$ ($1.00\times$ collapse $\mathcal{C}$), deployments range from ultra-low-latency edge streaming to high-throughput cloud serving. Given a detector with native baseline latency $T_0$ and customizable budget $\beta \in [-100\%, +100\%]$, our goal is to synthesize an optimal zero-NMS deployed model $g^\star(\beta)$ \emph{directly from the WSP-calibrated subspaces without retraining new weights}, strictly respecting $T(g) \le (1 + \beta)T_0$ while preserving the detector's official hardware compilation speedup ratio (\cref{sec:method_graph}).
\section{Methodology: \texorpdfstring{\method}{IR-ODWSP}}
\label{sec:method}

\subsection{Overview}
\label{sec:method_overview}

\begin{figure*}[t]
    \centering
    \includegraphics[width=0.99\linewidth]{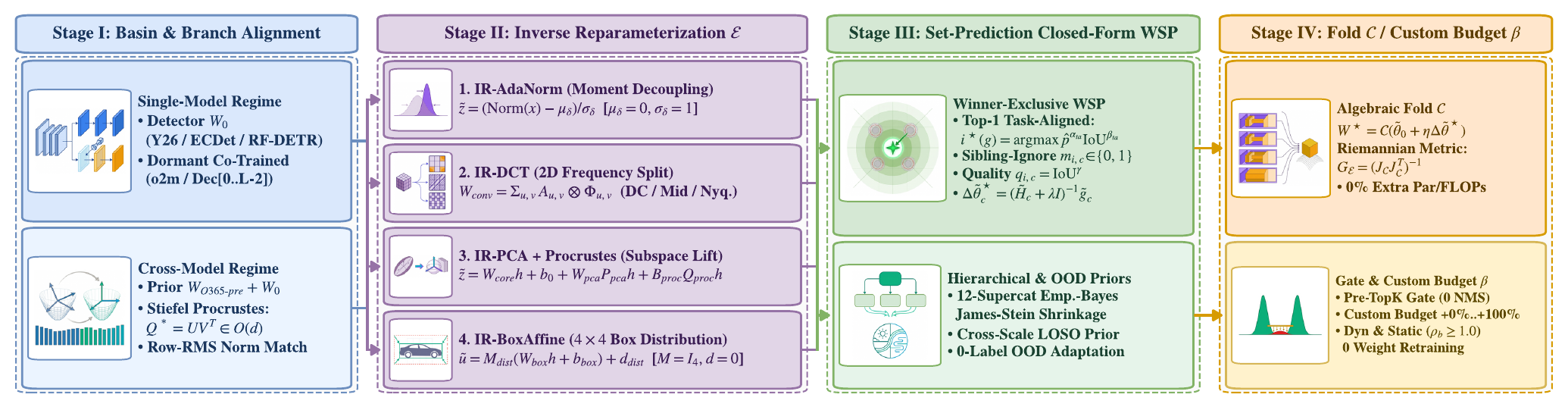}
    \vspace{-0.4em}
    \caption{\textbf{Four-stage architecture of \method.} Converged layers (dormant co-trained branches or cross-model priors aligned via Procrustes~\cite{schonemann1966generalized} $Q^\star \in O(d)$) are lifted via $\mathcal{E}$ into four decoupled subspaces (\texttt{IR-AdaNorm} moments~\cite{li2018adabn}, \texttt{IR-DCT} frequencies, \texttt{IR-PCA+Procrustes} covariances, and \texttt{IR-BoxAffine} box coordinates), edited in closed form under winner-exclusive targets~\cite{feng2021tood,wang2024yolov10} and Empirical-Bayes~\cite{efron1973stein} / LOSO priors, and deployed \emph{without retraining weights} under any latency budget $\beta \in [-100\%, +100\%]$: contracted for $\beta < 0\%$, collapsed via $\mathcal{C}$ at $\beta{=}0\%$ ($1.00\times$), or compiled into dual-stream dynamic (\methodDyn) or single-engine static (\methodSG) inference models.}
    \label{fig:arch}
    \vspace{-0.4em}
\end{figure*}

\Cref{fig:arch} summarizes the four stages of \textbf{\method}. Given a converged detector $W_0$ (and optional prior $W_{\text{O365}}$), \textbf{Stage~I} extracts dormant co-trained branches or aligns cross-basin symmetries via Stiefel Procrustes (\cref{sec:method_cross}). \textbf{Stage~II} lifts layers via exact inverse reparameterization $\mathcal{E}$ into decoupled statistical subspaces (\cref{sec:method_ir}). \textbf{Stage~III} solves closed-form set-prediction normal equations (\odwsp) over forward-only statistics (\cref{sec:method_odwsp}). \textbf{Stage~IV} determines the optimal deployed model \emph{without retraining weights} under any budget $\beta \in [-100\%, +100\%]$: contracting for $\beta < 0\%$, folding all edits via $\mathcal{C}$ into $W^\star \in \mathcal{W}$ at $\beta{=}0\%$ ($1.00\times$, \cref{alg:ir_odwsp}), or instantiating speedup-aligned zero-NMS graphs via dual-stream dynamic routing (\methodDyn, \cref{alg:dyn_graph}) or single-engine \texttt{TensorRT}+\texttt{CUDAGraph} (\methodSG, \cref{alg:static_graph}) for $\beta > 0\%$ (\cref{sec:method_graph}).

\subsection{Object-Detection Set-Prediction WSP}
\label{sec:method_odwsp}

To prevent duplicate-box collapse while operating strictly without gradients, we formulate a closed-form second-order quadratic approximation of the set-prediction objective around pre-activation logits $\bm{z}_i = W_{\text{cls}}\bm{h}_i + \bm{b}_{\text{cls}}$ and box coordinates $\hat{\bm{b}}_i$.

\paragraph{Step 1: Winner-Exclusive Assignment \& Sibling-Ignore Masking.}
For each calibration image $x$ with ground-truth objects $\mathcal{G} = \{(c_g, \bm{b}_g)\}$, we compute the task-aligned matching score~\cite{feng2021tood,wang2024yolov10} $t_{i,g} = \hat{p}_{i,c_g}^{\alpha_{\text{ta}}} \cdot \IoU(\hat{\bm{b}}_i, \bm{b}_g)^{\beta_{\text{ta}}}$ (setting $\alpha_{\text{ta}}=0.5, \beta_{\text{ta}}=6.0$) and greedily assign each ground truth a unique top-1 winner $i^\star(g) = \argmax_i t_{i,g}$. Crucially, any non-winning sibling candidate $i \neq i^\star(g)$ that still overlaps $\bm{b}_g$ with $\IoU(\hat{\bm{b}}_i, \bm{b}_g) \ge \tau_{\text{sib}}$ ($\tau_{\text{sib}}=0.40$) lies on the ambiguous boundary between positive object parts and duplicate detections; forcing it to $1$ causes duplicate collapse, whereas forcing it to $0$ suppresses recall. We therefore define a \emph{sibling-ignore mask} $m_{i,c} \in \{0, 1\}$ that masks non-winning overlapping siblings ($m_{i,c_g}=0$) while retaining background negatives ($m_{i,c}=1, q_{i,c}=0$) and assigning the unique winner $i^\star(g)$ ($m_{i^\star(g),c_g}=1$) a soft quality target:
\begin{equation}
    q_{i^\star(g),\, c_g} = \IoU\big(\hat{\bm{b}}_{i^\star(g)},\, \bm{b}_g\big)^\gamma, \qquad \gamma \in [0.75, 0.90].
    \label{eq:soft_iou_target}
\end{equation}

\paragraph{Step 2: Per-Class Focal Curvature Sufficient Statistics.}
Expanding focal-reweighted soft binary cross-entropy~\cite{lin2017focal,li2020gfl,feng2021tood} (with focal modulator frozen at $\hat{\bm{p}}_i = \sigma(\bm{z}_i^0)$) to second order around $\bm{z}_i^0$ yields non-negative IRLS curvature weights $\omega_{i,c} = m_{i,c} \big|q_{i,c} - \hat{p}_{i,c}\big|^{\gamma_f} \hat{p}_{i,c}(1 - \hat{p}_{i,c}) \ge 0$ ($\gamma_f=2.0$). Across all $NM$ candidates in $\mathcal{D}_{\text{cal}}$, one forward pass accumulates $\bm{H}_c \in \mathbb{R}^{(d+1)\times(d+1)}$ and $\bm{g}_c \in \mathbb{R}^{d+1}$:
\begin{equation}
\begin{aligned}
    \bm{H}_c &= \sum_{i} \omega_{i,c} \bar{\bm{h}}_i \bar{\bm{h}}_i^\top, \\
    \bm{g}_c &= \sum_{i} m_{i,c} \big|q_{i,c} - \hat{p}_{i,c}\big|^{\gamma_f} (q_{i,c} - \hat{p}_{i,c}) \bar{\bm{h}}_i,
\end{aligned}
\label{eq:focal_suff_stats}
\end{equation}
where $\bar{\bm{h}}_i = [\bm{h}_i^\top, 1]^\top \in \mathbb{R}^{d+1}$ augments terminal features with a bias coordinate.

\paragraph{Step 3: Empirical-Bayes Supercategory Shrinkage \& $4\times 4$ Box Debiasing.}
Rare COCO categories (\eg, \emph{hair drier}, \emph{toaster}) exhibit noisy sample Hessians. Partitioning $C=80$ classes into $S=12$ MS~COCO supercategories $\{G_s\}_{s=1}^{12}$ with counts $n_c$ ($\bar{n}_{\text{geom}} = (\prod_c n_c)^{1/C}$), we apply James-Stein (JS) Empirical-Bayes shrinkage~\cite{efron1973stein} to bias shifts $\Delta b_c$ toward precision-weighted group means $\bar{\mu}_{s(c)}$:
\begin{equation}
    \Delta b_c^{\text{JS}} = (1 - \kappa_c)\Delta b_c + \kappa_c \bar{\mu}_{s(c)}, \qquad \kappa_c = \kappa_0 \frac{\bar{n}_{\text{geom}}}{n_c + \bar{n}_{\text{geom}}}.
    \label{eq:js_shrinkage}
\end{equation}
Simultaneously, on winners $i^\star(g)$ with clamped inverse-sigmoid boxes $\bm{u}_i, \bm{u}_g \in \mathbb{R}^4$, \texttt{IR-BoxAffine} solves $(M_{\text{dist}}^\star, \bm{d}_{\text{dist}}^\star) = \argmin_{M, \bm{d}} \sum_g \IoU_g^2 \|M \bm{u}_{i^\star(g)} + \bm{d} - \bm{u}_g\|_2^2 + \lambda_{\text{box}}\|M - I_4\|_F^2$ and folds into $(W_{\text{box}}^\star, \bm{b}_{\text{box}}^\star)$.

\paragraph{Step 4: Zero-Parameter Pre-Postprocess Collision Gate.}
Detectors co-train auxiliary branches discarded at test time: YOLO26 co-trains a one-to-many ($\text{o2m}$) head alongside one-to-one ($\text{o2o}$), while EdgeCrafter/RF-DETR co-train decoder/Group-DETR~\cite{chen2023groupdetr} heads. On YOLO26, blending $\text{o2m}$ into $\text{o2o}$ re-introduces duplicates; instead, before top-$K_{\text{det}}$ selection ($s_i = \max_c \sigma(z_{i,c}^{\text{o2o}})$), we compute max higher-score overlap $\rho_i = \max_{j: s_j > s_i} \IoU(\hat{\bm{b}}_i, \hat{\bm{b}}_j)$ ($\max\varnothing \triangleq 0$) and $3{\times}3$ peak indicator $\delta_i \in \{0,1\}$, gating asymmetrically:
\begin{equation}
\begin{aligned}
    z_{i,c}^{\text{gate}} &= z_{i,c}^{\text{o2o}} + \gamma_{\text{dn}} \min\!\big(0, \Delta z_{i,c}\big) \\
    &\quad + \gamma_{\text{up}} \mathbb{I}(\rho_i < \tau_{\text{iou}})\,\delta_i \max\!\big(0, \Delta z_{i,c}\big),
\end{aligned}
\label{eq:collision_gate}
\end{equation}
where $\Delta z_{i,c} = z_{i,c}^{\text{aux}} - z_{i,c}^{\text{o2o}}$. Downward disagreement ($\Delta z_{i,c} < 0$) unconditionally suppresses false positives, whereas upward rescue ($\Delta z_{i,c} > 0$) fires \emph{only} on isolated local peaks free of cross-object collisions ($\rho_i < \tau_{\text{iou}}$), requiring zero trainable parameters and zero changes to stock post-processing.

\subsection{Inverse Reparameterization for Decoupled WSP}
\label{sec:method_ir}

To escape the Euclidean trap ($G = I_P$), \method lifts layers via four exact operators $(\mathcal{E}_k, \mathcal{C}_k)$ (three below plus \texttt{IR-BoxAffine}; proofs in \cref{sec:app_proofs}):

\paragraph{1. Moment-Decoupled Normalization (\texttt{IR-AdaNorm}) \& Domain Adaptation.}
Extending AdaBN~\cite{li2018adabn} to \texttt{BatchNorm2d} and \texttt{LayerNorm} ($\bm{z} = \mathrm{Norm}(\bm{x}; \bm{\gamma}_0, \bm{\beta}_0)$), $\mathcal{E}_{\text{norm}}$ appends identity-initialized residual moments $(\bm{\mu}_\delta = \bm{0}, \bm{\sigma}_\delta = \bm{1} > \bm{0})$:
\begin{equation}
\begin{aligned}
    \tilde{\bm{z}} &= \frac{\mathrm{Norm}(\bm{x}; \bm{\gamma}_0, \bm{\beta}_0) - \bm{\mu}_\delta}{\bm{\sigma}_\delta} \\
    &\xrightarrow{\;\mathcal{C}_{\text{norm}}\;} \bm{\gamma}^\star = \frac{\bm{\gamma}_0}{\bm{\sigma}_\delta}, \quad \bm{\beta}^\star = \frac{\bm{\beta}_0 - \bm{\mu}_\delta}{\bm{\sigma}_\delta}.
\end{aligned}
\label{eq:ir_adanorm}
\end{equation}
Given unlabeled target activations with channel mean $\hat{\bm{\mu}}$ and variance $\hat{\bm{\sigma}}^2$, Empirical-Bayes shrinkage ($\eta_{\text{bn}} \ll 1$) sets $\bm{\mu}_\delta = \eta_{\text{bn}}\hat{\bm{\mu}}$ and $\bm{\sigma}_\delta = \sqrt{(1-\eta_{\text{bn}})\bm{1} + \eta_{\text{bn}}\mathrm{clip}(\hat{\bm{\sigma}}^2, \frac{1}{4}, 4)}$, enabling \emph{zero-label, zero-gradient adaptation} folded into $(\bm{\gamma}^\star, \bm{\beta}^\star)$.

\paragraph{2. Spatial-Frequency Kernel Decoupling (\texttt{IR-DCT}).}
A $K{\times}K$ ($K \ge 3$) kernel $W_{\text{conv}} \in \mathbb{R}^{C_{\text{out}}\times C_{\text{in}}\times K\times K}$ entangles DC offsets, contours, and high-frequency noise. Using the orthonormal 2D-DCT basis $\{\Phi_{u,v}\}_{u,v=0}^{K-1}$ ($\langle \Phi_{u,v}, \Phi_{u',v'}\rangle = \delta_{uu'}\delta_{vv'}$), $\mathcal{E}_{\text{dct}}$ splits $W_{\text{conv}}$ into $K^2$ frequency branches ($4\text{D}$ outer product $\otimes$):
\begin{equation}
\begin{aligned}
    W_{\text{conv}} &= \sum_{u=0}^{K-1}\sum_{v=0}^{K-1} A_{u,v} \otimes \Phi_{u,v}, \\
    A_{u,v} &= \langle W_{\text{conv}}, \Phi_{u,v}\rangle \in \mathbb{R}^{C_{\text{out}}\times C_{\text{in}}}.
\end{aligned}
\label{eq:ir_dct}
\end{equation}
Band-specific penalties $\lambda_{u,v}$ (preserving DC $(0,0)$, amplifying mid-band shapes, damping Nyquist noise $(K{-}1,K{-}1)$) induce $G_{\mathcal{E}}^{\text{dct}} = \sum_{u,v}\frac{\lambda_{u,v}}{\lambda}\mathrm{vec}(\Phi_{u,v})\mathrm{vec}(\Phi_{u,v})^\top$ under $\mathcal{C}_{\text{dct}}(\{A_{u,v}^\star\}) = \sum_{u,v} A_{u,v}^\star \otimes \Phi_{u,v}$.

\begin{algorithm}[t]
\caption{$1.00\times$ Folded \method ($\beta = 0\%$)}
\label{alg:ir_odwsp}
\small
\begin{algorithmic}[1]
\Require Pretrained $W_0$, optional prior $W_{\text{aux}}$, calibration set $\mathcal{D}_{\text{cal}}$, disjoint holdout $\mathcal{D}_{\text{hold}}$; autograd off ($\nabla \equiv \bm{0}$).
\Ensure Edited native detector $W^\star \in \mathcal{W}$ ($\Delta\text{Params}=0\%, \Delta\text{FLOPs}=0\%$), calibrated subspace bank $\mathcal{B}$, and gate $\psi^\star$.
\If{$W_{\text{aux}} \neq \varnothing$} \Comment{Stage I: Cross-model symmetry alignment}
    \State Align trunk \& ontology heads of $W_{\text{aux}}$ to $W_0$ via Procrustes SVD $Q^\star = UV^\top$ and row-RMS matching (\cref{eq:procrustes_align}).
\EndIf
\State Extract orthonormal basis $Q_{\text{proc}}$ from dormant co-trained heads ($\text{o2m}$ / decoder / Group-DETR) and/or aligned $W_{\text{aux}}$.
\State Lift normalization, conv, readout, and box heads via $\tilde{\theta}_0 = \mathcal{E}(W_0)$ (\cref{eq:ir_adanorm,eq:ir_dct,eq:ir_pca}). \Comment{Stage II}
\State Run one forward pass on $\mathcal{D}_{\text{cal}}$: assign Top-1 winners $i^\star(g)$, build sibling-ignore mask $m_{i,c}$ and quality $q_{i,c}=\IoU^\gamma$ (\cref{eq:soft_iou_target}), and accumulate $(\bm{H}_c, \bm{g}_c, \Sigma_h)$ (\cref{eq:focal_suff_stats}). \Comment{Stage III}
\For{candidate $(\lambda, \eta, \kappa_0)$ in Leave-One-Scale-Out (LOSO) grid}
    \State Solve lifted ridge equations $\Delta\tilde{\theta}_c^\star = (\tilde{\bm{H}}_c + \lambda I)^{-1}\tilde{\bm{g}}_c$; apply 12-supercategory JS shrinkage (\cref{eq:js_shrinkage}) and $4\times 4$ box debiasing.
    \State Collapse $W_{\text{cand}} = \mathcal{C}(\tilde{\theta}_0 + \eta\Delta\tilde{\theta}^\star)$ (\cref{eq:ir_adanorm,eq:ir_dct,eq:ir_pca_collapse}) and evaluate AP on $\mathcal{D}_{\text{hold}}$ (fallback $W_0$). \Comment{Stage IV}
\EndFor
\State Calibrate pre-postprocess collision gate $\psi^\star = (\gamma_{\text{dn}}, \gamma_{\text{up}}, \tau_{\text{iou}})$ (\cref{eq:collision_gate}) on $\mathcal{D}_{\text{hold}}$; retain aligned Stiefel/group bank $\mathcal{B}$; \Return $(W^\star, \mathcal{B}, \psi^\star)$.
\end{algorithmic}
\end{algorithm}

\paragraph{3. Covariance-Preconditioned \& Auxiliary Subspace Lifting (\texttt{IR-PCA}).}
Let $\Sigma_h = \frac{1}{NM}\sum_i \bm{h}_i \bm{h}_i^\top = U_h \Lambda_h U_h^\top$ be the feature covariance ($W_{\text{cls},0}^\top W_{\text{cls},0}$ if $N=0$). With $1 \le r, r_p \le d$, $\mathcal{E}_{\text{pca}}$ splits $(W_{\text{cls},0}, \bm{b}_0)$ into a core branch, top-$r$ PCA branch, and orthonormal branch $Q_{\text{proc}} \in \mathbb{R}^{r_p \times d}$ ($Q_{\text{proc}}Q_{\text{proc}}^\top = I_{r_p}$):
\begin{equation}
    \tilde{\bm{z}}(\bm{h}) = (W_{\text{core}}\bm{h} + \bm{b}_0) + W_{\text{pca}} (P_{\text{pca}}\bm{h}) + B_{\text{proc}}(Q_{\text{proc}}\bm{h}),
    \label{eq:ir_pca}
\end{equation}
with $W_{\text{core}}=W_{\text{cls},0}, W_{\text{pca}}=\bm{0}, B_{\text{proc}}=\bm{0}$, and $P_{\text{pca}} = (\tilde{\Lambda}_{h,r} + \epsilon I_r)^{1/2} U_{h,r}^\top$ ($\tilde{\Lambda}_{h,r} = \Lambda_{h,r}/\mathrm{med}(\Lambda_{h,r})$). With $\bar{J}_{\mathcal{C}} = \mathrm{blkdiag}(J_{\mathcal{C}}, 1)$, $\tilde{\bm{H}}_c = \bar{J}_{\mathcal{C}}^\top \bm{H}_c \bar{J}_{\mathcal{C}}$, and $\tilde{\bm{g}}_c = \bar{J}_{\mathcal{C}}^\top \bm{g}_c$ (\cref{eq:focal_suff_stats}), solving in $\Delta\tilde{\theta}_c = [\Delta W_{\text{core},c}, \Delta W_{\text{pca},c}, \Delta B_{\text{proc},c}, \Delta b_c]^\top$ and collapsing $\bm{b}^\star = \bm{b}_0 + \Delta\bm{b}^{\text{JS}}$ and
\begin{equation}
    W_{\text{cls}}^\star = \mathcal{C}_{\text{pca}}(\tilde{\theta}^\star) = W_{\text{core}}^\star + W_{\text{pca}}^\star P_{\text{pca}} + B_{\text{proc}}^\star Q_{\text{proc}}
    \label{eq:ir_pca_collapse}
\end{equation}
yields $J_{\mathcal{C}} J_{\mathcal{C}}^\top = I_d + U_{h,r}(\tilde{\Lambda}_{h,r}+\epsilon I_r)U_{h,r}^\top + Q_{\text{proc}}^\top Q_{\text{proc}}$, amplifying high-SNR semantic and dormant subspaces while damping noise.

\subsection{Full \texorpdfstring{\method}{IR-ODWSP} Algorithm}
\label{sec:method_cross}

\paragraph{Single-Model WSP.}
When only $W_0$ is available ($W_{\text{aux}}=\varnothing$), $Q_{\text{proc}}$ in \cref{eq:ir_pca} is constructed by SVD-orthonormalizing the row space of $W_0$'s dormant co-trained heads ($\text{o2m}$ towers in YOLO26; decoder $0\dots L{-}2$ or 12 Group-DETR encoder heads in EdgeCrafter/RF-DETR) aligned onto the primary readout via orthogonal Procrustes.

\paragraph{Cross-Model Orthogonal Procrustes WSP.}
Given prior $W_{\text{O365}}$, naive interpolation $(1-\alpha)W_{\text{O365}} + \alpha W_0$~\cite{wortsman2022wiseft} fails when channels differ by $Q \in O(d_{\text{in}})$ across disjoint/drifted basins (\cref{sec:app_ab5}). Reshaping shared layers as $M_0, M_1 \in \mathbb{R}^{d_{\text{out}}\times d_{\text{in}}}$, we align $M_0$ ($W_{\text{O365}}$) to $M_1$ ($W_0$) over $O(d_{\text{in}})$ via Procrustes~\cite{schonemann1966generalized} ($M_0^\top M_1 = U\Sigma V^\top$) and row-RMS matching:
\begin{equation}
\begin{aligned}
    Q^\star &= \argmin_{Q \in O(d_{\text{in}})} \|M_0 Q - M_1\|_F = U V^\top, \\
    \widehat{M}_0 &= \diag_i\!\left(\frac{\|(M_1)_{i,:}\|_2}{\|(M_0 Q^\star)_{i,:}\|_2 + \epsilon}\right) M_0 Q^\star,
\end{aligned}
\label{eq:procrustes_align}
\end{equation}
interpolating $M_{\text{mix}}(\alpha) = \diag_i\!\big(\|(M_1)_{i,:}\|_2 / (\|\widetilde{M}(\alpha)_{i,:}\|_2 + \epsilon)\big)\widetilde{M}(\alpha)$ with $\widetilde{M}(\alpha) = (1-\alpha)\widehat{M}_0 + \alpha M_1$, while matched 365-to-80 ontology rows define $Q_{\text{proc}}$ in \cref{eq:ir_pca}.

\subsection{Latency-Budget-Aware Graph-Expansion Inference}
\label{sec:method_graph}

While $\beta = 0\%$ algebraic collapse $\mathcal{C}$ (\cref{alg:ir_odwsp}) preserves bit-identical $1.00\times$ native latency $T_0$, practical deployments specify customizable latency budgets $\beta \in [-100\%, +100\%]$---either expanding bounded compute ($\beta > 0\%$, up to unconstrained \methodplus) for higher accuracy or contracting compute ($\beta < 0\%$) under tighter hardware deadlines---without training new weights. Crucially, naively expanding multi-view or multi-head graphs degrades compiled \texttt{TensorRT} efficiency due to L2 SRAM cache thrashing at batch $B_{\text{bat}}{=}8$, unfused \texttt{TopK/Scatter} epilogues, and stream serialization (\cref{sec:app_graph_ablations}). Letting $d_0$ denote the native $1.00\times$ graph with official hardware compilation speedup $S_{\text{CG},b}(d_0) \triangleq T_{\text{PT}}(d_0) / T_{\text{CG},b}(d_0)$ ($b \in \{1, 8\}$), we formulate latency-budget-aware graph selection on disjoint holdout $\mathcal{D}_{\text{hold}}$:
\begin{equation}
\begin{aligned}
    g^\star(\beta) &= \argmax_{g \in \mathcal{P}(W^\star, \mathcal{B})} \; \mathrm{AP}_{\text{hold}}(g) \\
    \text{s.t.} \quad & T(g) \le (1 + \beta)\, T_0, \\
    & \rho_b(g) \triangleq \frac{S_{\text{CG},b}(g)}{S_{\text{CG},b}(d_0)} \ge 1 - \epsilon, \quad \forall b \in \{1, 8\},
\end{aligned}
\label{eq:knapsack}
\end{equation}
where every candidate expanded graph $g \in \mathcal{P}(W^\star, \mathcal{B})$ is assembled \emph{directly from the WSP-calibrated weights $(W^\star, \mathcal{B})$ without retraining any parameters} and fuses predictions in 1-to-1 slot correspondence with \emph{zero external NMS/WBF}. To satisfy \cref{eq:knapsack} while adapting to distinct deployment scenarios---spanning low-latency streaming environments that exploit concurrent multi-stream SM utilization and strict single-artifact runtimes requiring zero multi-engine scheduling---we co-design custom Tensor-Core GEMM and SRAM warp-consensus \texttt{TensorRT~11.4} \texttt{IPluginV3} kernels (\cref{sec:app_algos}) and propose both a dynamic-graph (\methodDyn) and a static-graph (\methodSG) expansion algorithm:

\begin{algorithm}[t]
\caption{\methodDyn: Latency-Budget-Aware Dual-Stream Dynamic-Graph Inference}
\label{alg:dyn_graph}
\small
\begin{algorithmic}[1]
\Require Calibrated $(W^\star, \mathcal{B})$ from \cref{alg:ir_odwsp}, latency budget $\beta \ge 0$, tolerance $\epsilon$, holdout $\mathcal{D}_{\text{hold}}$, input $\bm{x} \in \mathbb{R}^{B_{\text{bat}}\times 3\times H^\star\times W^\star}$.
\Ensure Zero-NMS detections $\widehat{\bm{Y}} \in \mathbb{R}^{B_{\text{bat}}\times 300\times 6}$ with $T \le (1{+}\beta)T_0$ and $\rho_1, \rho_8 \ge 1{-}\epsilon$.
\Statex \textbf{Offline Zero-Retraining Configuration (once per $\beta$):}
\State Instantiate modular sub-engines $(E_{\text{pri}}, E_{\text{aux}}, E_{\text{cons}})$ from $(W^\star, \mathcal{B})$ using fused custom \texttt{IPluginV3} stems and consensus ops.
\State Bind $(E_{\text{pri}}, E_{\text{aux}})$ to disjoint execution workspaces on parallel CUDA streams $(s_1, s_2)$ synchronized via static \texttt{cudaEvent} fork-join capture into $g \in \mathcal{P}^{\text{Dyn}}$; select $g^\star(\beta)$ via \cref{eq:knapsack} on $\mathcal{D}_{\text{hold}}$.
\Statex \textbf{Online Inference Execution ($0$ Host Sync, $0$ External NMS):}
\If{$g^\star(\beta) = d_0$} \Return $E_0(\bm{x})$ \Comment{$\beta{=}0\%$ native $1.00\times$ path}
\EndIf
\For{L2-resident sub-batch $\bm{x}^{(c)}$ of size $B_{\text{chunk}} = \min(B_{\text{bat}}, 4)$}
    \State Record fork event $e_{\text{fork}}$ on $s_1$; launch primary Stiefel/13-Group stem $E_{\text{pri}}(\bm{x}^{(c)})$ on $s_1$.
    \State Wait $e_{\text{fork}}$ on $s_2$; concurrently execute single-read warped/flipped auxiliary branch $E_{\text{aux}}(\bm{x}^{(c)})$ on $s_2$ and record $e_{\text{join}}$.
    \State Wait $e_{\text{join}}$ on $s_1$; run single-kernel SRAM/warp 1-to-1 consensus $E_{\text{cons}}$ (peak-protected grid warp on CNNs or IoU-gated query fusion + bitonic \texttt{TopK(300)} on DETRs) on $s_1$.
\EndFor
\State \Return concatenated $\widehat{\bm{Y}} \in \mathbb{R}^{B_{\text{bat}}\times 300\times 6}$.
\end{algorithmic}
\end{algorithm}

\paragraph{1. Latency-Budget-Aware Dual-Stream Dynamic-Graph Inference (\methodDyn, \cref{alg:dyn_graph}).}
When low-latency single-frame streaming ($B_{\text{bat}}{=}1$) leaves GPU SMs under-utilized on compact detectors, executing a second spatial/Stiefel branch sequentially inside one stream wastes idle compute. \methodDyn (\cref{alg:dyn_graph}) partitions the calibrated WSP graph into GPU-resident primary and auxiliary sub-engines assigned to disjoint memory workspaces on two concurrent CUDA streams $(s_1, s_2)$ captured via static \texttt{cudaEvent} fork-join dependencies into a single \texttt{cudaGraph\_t}. Single-read dual-view preprocessing plugins eliminate intermediate global-memory tensor writes, while custom warp-tiled grid consensus (on YOLO26) and SRAM-resident IoU-matched query consensus (on EdgeCrafter/RF-DETR) fuse multi-branch outputs into 300 1-to-1 detections in a single kernel ($1.34\times$--$2.35\times$ faster than standard \texttt{TensorRT} lowering).

\paragraph{2. Latency-Budget-Aware Monolithic Single-Engine Static-Graph Inference (\methodSG, \cref{alg:static_graph}).}
For strict single-artifact deployment ($n_{\text{engines}}{=}1$, zero multi-engine management), \methodSG (\cref{alg:static_graph}) compiles $100\%$ of view synthesis, shared-backbone Stiefel/13-Group fan-out, and custom \texttt{IPluginV3} consensus ($\bm{x} \in \mathbb{R}^{B_{\text{bat}}\times 3\times H^\star\times W^\star} \mapsto \widehat{\bm{Y}} \in \mathbb{R}^{B_{\text{bat}}\times 300\times 6}$) into \emph{one Strongly-Typed FP16 \texttt{ICudaEngine}}. Three compiler co-design mechanisms guarantee both FP16 parity ($0.01\text{--}0.07$~AP vs.\ FP32) and $\rho_b \ge 1 - \epsilon$: (i)~horizontally fusing dual-basin readout convolutions while replacing per-anchor FP32 \texttt{Pow} gates with nested \texttt{Sqrt} algebraic equivalents; (ii)~splitting multi-scale ViT-DETR encoder memory per pyramid level prior to attention value projection to prevent \texttt{TensorRT} transpose-peephole corruption; and (iii)~encapsulating dual $2{\times}(B_{\text{bat}}/2)$ execution contexts inside one \texttt{CUDAGraph} at $B_{\text{bat}}{=}8$ so multi-view working sets remain strictly inside GPU L2 SRAM. When the deployment environment permits either paradigm, \textbf{\methodAuto} selects $\argmax_{g \in \{\text{\methodDyn}, \text{\methodSG}\}} \mathrm{AP}_{\text{hold}}(g)$ on $\mathcal{D}_{\text{hold}}$ (controlled ablations over alternative static/dynamic topologies in \cref{sec:app_graph_ablations}).

\begin{algorithm}[!t]
\caption{\methodSG: Latency-Budget-Aware Monolithic Single-Engine Static-Graph Inference}
\label{alg:static_graph}
\small
\begin{algorithmic}[1]
\Require Calibrated $(W^\star, \mathcal{B})$ from \cref{alg:ir_odwsp}, latency budget $\beta \ge 0$, tolerance $\epsilon$, holdout $\mathcal{D}_{\text{hold}}$, input $\bm{x} \in \mathbb{R}^{B_{\text{bat}}\times 3\times H^\star\times W^\star}$.
\Ensure Single-engine \texttt{ICudaEngine} detections $\widehat{\bm{Y}} \in \mathbb{R}^{B_{\text{bat}}\times 300\times 6}$ with $T \le (1{+}\beta)T_0$ and $\rho_1, \rho_8 \ge 1{-}\epsilon$.
\Statex \textbf{Offline Monolithic Engine Compilation (once per $\beta$):}
\State Construct constant-shape graph $G(\bm{x})$ from $(W^\star, \mathcal{B})$ via horizontal multi-head conv fusion, nested-\texttt{Sqrt} precision-safe gates, split-first decoder value projection, and custom \texttt{IPluginV3} Tensor-Core proposal / SRAM consensus nodes ($0$ dynamic shapes).
\State Compile $G$ into \emph{one} Strongly-Typed FP16 \texttt{TensorRT} \texttt{ICudaEngine} $E_{\text{SG}}$ (surgical FP32 guards on \texttt{LayerNorm}/\texttt{GridSample} reductions only); capture L2-chunked \texttt{CUDAGraph} $\Gamma_{\text{SG}}$ and select $g^\star(\beta)$ via \cref{eq:knapsack} on $\mathcal{D}_{\text{hold}}$.
\Statex \textbf{Online Single-Launch Replay ($1$ Engine, $0$ External NMS):}
\If{$B_{\text{bat}} \le 4$ \textbf{or} single-view topology}
    \State Replay monolithic $\Gamma_{\text{SG}}(\bm{x}) \mapsto \widehat{\bm{Y}}$ in a single \texttt{cudaGraphLaunch}.
\Else \Comment{Dual-context L2-resident $2{\times}(B_{\text{bat}}/2)$ capture inside $\Gamma_{\text{SG}}$}
    \State Replay $\Gamma_{\text{SG}}$ executing $E_{\text{SG}}(\bm{x}_{1:B_{\text{bat}}/2})$ and $E_{\text{SG}}(\bm{x}_{B_{\text{bat}}/2+1:B_{\text{bat}}})$ within L2 SRAM capacity in one \texttt{cudaGraphLaunch}.
\EndIf
\State \Return $\widehat{\bm{Y}} \in \mathbb{R}^{B_{\text{bat}}\times 300\times 6}$.
\end{algorithmic}
\end{algorithm}

\paragraph{3. Zero-Gradient Negative-Budget Contraction via Closed-Form Subgraph Bridges ($\beta \in [-100\%, -10\%]$).}
When $\beta < 0\%$, native $d_0$ exceeds the hard wall-clock budget $(1+\beta)T_0$ and leaves the feasible set. We construct a family of single-engine static contracted subgraphs $\mathcal{P}^{-}(W^\star, \mathcal{B})$ spanning: (i)~single-head inference pruning (\eg, fusing \texttt{Conv+BN} and pruning auxiliary \texttt{o2m} or multi-group training branches) paired with spatial resolution contraction; and (ii)~topological layer bypass (\eg, bypassing the bidirectional PAN neck on YOLO26, truncating decoder layers from $L \to 1$ or $0$ on EdgeCrafter and RF-DETR, and shallow backbone truncation down to the $\beta = -100\%$ zero-compute null boundary). Naively bypassing intermediate neck or decoder blocks collapses accuracy because intermediate activations no longer match the frozen detection head's expected feature basis and variance scale. To bridge bypassed layers with \emph{zero backpropagation}, we fit a closed-form \textbf{Moment-Matched Foreground-Weighted Ridge (\texttt{MM-FW-Ridge})} linear projection $P_{\text{br}}^\star = \mathrm{diag}(\bm{\sigma}_Y / \bm{\sigma}_{\widehat{Y}})\,Y^\top \Omega X (X^\top \Omega X + \lambda_{\text{br}} I)^{-1} \in \mathbb{R}^{d_{\text{out}}\times d_{\text{in}}}$ strictly on forward calibration activations from $\mathcal{D}_{\text{cal}}$ (weighting foreground-active spatial/query tokens via $\Omega$ and restoring target channel moments), then select $g^\star(\beta) = \argmax_{g \in \mathcal{P}^{-}: T(g) \le (1+\beta)T_0} \mathrm{AP}_{\text{hold}}(g)$ strictly on disjoint $\mathcal{D}_{\text{hold}}$ (\cref{sec:app_neg_continuum}).
\section{Experimental Setup}
\label{sec:setup}

\begin{table*}[!t]
\caption{\textbf{Main comparison across 3 families and 17 scales on MS~COCO~\cite{lin2014coco} \texttt{val2017} (ID) and Zero-Shot COCO-O~\cite{mao2023cocoo} (6-domain mAP).} \texttt{O365$\to$COCO} denotes Objects365~\cite{shao2019objects365}$\to$COCO weights. All $1.00\times$ methods (incl.\ Van.\ WSP adapted from~\cite{takehi2026sparks}) add zero parameters, FLOPs, or latency (colors reflect 3-decimal AP, \cref{tab:app_all26}; edited parameter counts and ratios in \cref{tab:app_edited_params}); $^\dagger$\methodplus denotes the unconstrained multi-subspace upper bound (\cref{tab:tta_vs_mem}).}
\label{tab:main_17}
\centering
\resizebox{\textwidth}{!}{\begin{tabular}{@{}llrr|cccccc|ccccc@{}}
\toprule
& & & & \multicolumn{6}{c|}{\textbf{In-Distribution: MS~COCO~\cite{lin2014coco} \texttt{val2017} ($\text{AP}$ / Small $\text{AP}_s$)}} & \multicolumn{5}{c}{\textbf{Zero-Shot OOD: COCO-O~\cite{mao2023cocoo} ($6$-Domain mAP)}} \\
\textbf{Family} & \textbf{Model} & \textbf{Par.} & \textbf{FLOPs} & \textbf{Official} & \textbf{Van.\ WSP} & \textbf{\odwsp} & \textbf{\method} & \textbf{\method} & \textbf{\methodplus$^\dagger$} & \textbf{Official} & \textbf{Van.\ WSP} & \textbf{\odwsp} & \textbf{\method} & \textbf{\methodplus$^\dagger$} \\
& & (M) & (G) & ($\text{AP}$ / $\text{AP}_s$) & ($1.00\times$) & ($1.00\times$) & (Pure $0\%$) & ($1.00\times$) & (Unconstr.) & & ($1.00\times$) & ($1.00\times$) & ($1.00\times$) & (Unconstr.) \\
\midrule
\multirow{5}{*}{\makecell[l]{YOLO26~\cite{jocher2026yolo26}\\(\texttt{o2o})}}
& N & 2.57 & 6.1 & 40.07 / 19.74 & \negv{39.99} & 40.08 & 40.11 & \pos{40.15} / \pos{20.03} & \pos{42.57} & 18.39 & 18.41 & 18.39 & \pos{18.42} & \pos{20.25} \\
& S & 10.01 & 22.8 & 47.79 / 29.16 & \negv{47.77} & \negv{47.76} & 47.85 & \pos{47.90} / \pos{29.47} & \pos{49.89} & 24.41 & 24.48 & 24.41 & \pos{24.50} & \pos{26.19} \\
& M & 21.90 & 75.0 & 52.50 / 36.25 & \negv{52.43} & 52.52 & 52.52 & \pos{52.59} / \pos{36.47} & \pos{53.97} & 27.50 & 27.53 & 27.50 & \pos{27.54} & \pos{29.34} \\
& L & 26.30 & 93.4 & 54.35 / 37.88 & \negv{54.27} & 54.36 & 54.37 & \pos{54.51} / \pos{38.15} & \pos{55.45} & 29.54 & 29.54 & \pos{29.76} & \pos{29.62} & \pos{30.77} \\
& X & 58.99 & 208.9 & 56.88 / 41.44 & \negv{56.83} & 56.91 & 56.91 & \pos{56.92} / \pos{41.54} & \pos{57.68} & 32.66 & 32.69 & 32.66 & \pos{32.75} & \pos{33.80} \\
\midrule
\multirow{4}{*}{\makecell[l]{ECDet~\cite{liu2026edgecrafter}\\(COCO)}}
& S & 9.88 & 50.1 & 51.71 / 32.26 & \negv{51.70} & \negv{51.67} & 51.73 & \pos{51.73} / \pos{32.30} & \pos{53.25} & 30.08 & 30.10 & \negv{29.97} & \negv{30.08} & \pos{31.71} \\
& M & 19.43 & 85.7 & 54.22 / 35.86 & 54.22 & 54.22 & 54.22 & \pos{54.22} / \pos{35.89} & \pos{55.51} & 35.32 & 35.34 & 35.32 & \pos{35.35} & \pos{36.79} \\
& L & 32.97 & 148.3 & 56.99 / 38.70 & 56.99 & 56.99 & 56.99 & \pos{56.99} / \negv{38.68} & \pos{58.37} & 43.78 & 43.78 & 43.78 & \negv{43.76} & \pos{44.62} \\
& X & 49.94 & 208.9 & 57.86 / 38.57 & \negv{57.86} & 57.89 & 57.89 & \pos{57.89} / \pos{38.63} & \pos{59.19} & 45.21 & 45.22 & \pos{45.37} & \pos{45.37} & \pos{46.65} \\
\midrule
\multirow{4}{*}{\makecell[l]{ECDet~\cite{liu2026edgecrafter}\\(O365$\to$\\COCO)}}
& S & 9.88 & 50.1 & 53.53 / 35.11 & 53.56 & 53.53 & 53.58 & \pos{53.58} / \pos{35.15} & \pos{55.03} & 31.11 & 31.11 & 31.11 & \pos{31.14} & \pos{33.04} \\
& M & 19.43 & 85.7 & 56.75 / 38.93 & 56.75 & 56.75 & 56.76 & \pos{56.76} / \negv{38.92} & \pos{58.14} & 37.55 & \negv{37.55} & 37.55 & \pos{37.55} & \pos{38.93} \\
& L & 32.97 & 148.3 & 59.04 / 42.08 & \negv{59.03} & 59.04 & 59.05 & \pos{59.05} / \pos{42.09} & \pos{60.31} & 45.18 & 45.18 & \pos{45.19} & \pos{45.21} & \pos{46.51} \\
& X & 49.94 & 208.9 & 59.87 / 42.35 & \negv{59.86} & \negv{59.85} & 59.90 & \pos{59.90} / \pos{42.43} & \pos{61.10} & 47.32 & 47.35 & \pos{47.39} & \pos{47.34} & \pos{48.54} \\
\midrule
\multirow{4}{*}{RF-DETR~\cite{robinson2026rfdetr}}
& N & 30.47 & 31.9 & 48.41 / 25.32 & \negv{48.39} & 48.47 & 48.49 & \pos{48.49} / \pos{25.43} & \pos{54.63} & 30.12 & \negv{30.09} & \pos{30.13} & \negv{30.10} & \pos{33.58} \\
& S & 32.11 & 59.6 & 52.94 / 31.74 & \negv{52.94} & 52.96 & 53.00 & \pos{53.00} / \pos{31.88} & \pos{57.65} & 31.89 & \negv{31.86} & \pos{32.01} & \pos{32.06} & \pos{34.29} \\
& M & 33.69 & 78.6 & 54.75 / 35.95 & \negv{54.74} & 54.88 & 54.91 & \pos{54.91} / \pos{36.19} & \pos{58.68} & 32.44 & \negv{32.42} & \pos{32.51} & \pos{32.70} & \pos{34.15} \\
& L & 33.93 & 125.1 & 56.53 / 38.95 & \negv{56.52} & 56.56 & 56.61 & \pos{56.61} / \pos{39.13} & \pos{59.11} & 32.57 & 32.61 & \negv{32.42} & \negv{32.44} & \pos{33.27} \\
\midrule
\rowcolor{rowgray}
\multicolumn{4}{l|}{\textbf{17-Model Mean ($\Delta$ vs.\ Official)}} & 53.78 / 35.31 & \negv{$-0.020$} & $+0.014$ & \pos{$+0.042$} & \pos{$+0.061$ / $+0.123$} & \pos{$+2.138$} & 33.83 & $+0.012$ & $+0.023$ & \pos{$+0.052$} & \pos{$+1.610$} \\
\bottomrule
\end{tabular}}
\vspace{-0.4em}
\end{table*}

\paragraph{Datasets and Evaluation Protocol.}
We evaluate in-distribution (ID) detection on \textbf{MS~COCO 2017}~\cite{lin2014coco}, splitting \texttt{train2017} into forward-only calibration folds and a disjoint $3{,}000$-image holdout $\mathcal{D}_{\text{hold}}$ while evaluating \texttt{val2017} blindly. Zero-shot OOD generalization is tested directly on \textbf{COCO-O}~\cite{mao2023cocoo} across six style and weather shifts with zero target exposure, alongside label-free weather adaptation and corruption repair (\cref{sec:app_setup_details}).

\paragraph{Detector Families.}
Across 17 checkpoints (26 single- and cross-model settings), we span three end-to-end paradigms: (1)~\textbf{YOLO26}~\cite{jocher2026yolo26} (N--X), native NMS-free one-to-one CNNs; (2)~\textbf{EdgeCrafter (ECDet)}~\cite{liu2026edgecrafter} (S--X), distilled hybrid ViT-DETRs covering pure COCO and \textbf{O365}~\cite{shao2019objects365}\textbf{$\to$COCO} fine-tuned tracks; and (3)~\textbf{RF-DETR}~\cite{robinson2026rfdetr} (N--L), DINOv2~\cite{oquab2024dinov2} multi-scale Deformable DETRs (specifications in \cref{tab:main_17,sec:app_setup_details}).

\paragraph{Experimental Setup and Baselines.}
Offline WSP synthesis used \textbf{Gemini~4}; calibration and inference run deterministically in PyTorch (PT) and TensorRT~11.4 FP16 (TRT + \texttt{CUDAGraph}) on \textbf{NVIDIA RTX PRO~6000} GPUs ($0$ runtime LLM calls). At $\beta = 0\%$ ($1.00\times$ cost, $\nabla\equiv\bm{0}$), we compare \textbf{Official}, \textbf{Vanilla WSP} (adapted from~\cite{takehi2026sparks}), \textbf{\odwsp}, collapsed \textbf{\method (Pure $0\%$)}, and \textbf{\method ($1.00\times$)} (+$\text{o2m}$ gate on YOLO26). To verify that our latency-budget-aware framework applies across every architecture and model scale (\cref{sec:tta_vs_mem,sec:app_tta_audit}), we evaluate \textbf{\methodDyn} (\cref{alg:dyn_graph}), \textbf{\methodSG} (\cref{alg:static_graph}), and \textbf{\methodAuto} on each of the 17 detectors under budgets $\beta \in \{+10\%, +20\%, \dots, +100\%\}$ against multi-scale + flip \textbf{Official TTA} with and without NMS.
\section{Experimental Results}
\label{sec:results}

\subsection{Main Results Across Detectors and Scales}
\label{sec:main_results}
\label{sec:tta_vs_mem}

\Cref{tab:main_17} evaluates all 17 detectors across YOLO26 (N--X), EdgeCrafter (COCO / O365$\to$COCO, S--X), and RF-DETR (N--L) on \textbf{MS~COCO \texttt{val2017}} ($\text{AP}$ / $\text{AP}_s$) and zero-shot \textbf{COCO-O} 6-domain mAP (26/26 results in \cref{sec:app_extra_exp}).

\paragraph{Breaking the $1.00\times$ Ceiling \& Head-DoF Scaling.}
Vanilla WSP degrades $13/17$ converged detectors (\cref{tab:main_17}), confirming that isotropic spectral truncation and binary ridge updates disrupt calibrated foreground-background decision boundaries in end-to-end heads. Set-prediction reformulation (\odwsp) reverses this negative trend on average yet stalls inside narrow Euclidean valleys on heavily distilled models such as EdgeCrafter. Lifting into decoupled subspaces breaks this ceiling: pure collapsed \method (Pure $0\%$) already yields $17/17$ non-negative improvements ($+0.042$~AP), and full $1.00\times$ \method achieves strictly positive blind gains across \textbf{all 26/26} single- and cross-model settings ($100\%$, up to $+0.16$~AP; $+2.14$~AP mean in unconstrained \methodplus), scaling monotonically with readout degrees of freedom from compact O365$\to$COCO to multi-scale RF-DETR and YOLO26 heads.

\paragraph{Curvature Starvation \& Zero-Shot Generalization.}
At native single-scale resolution, \method boosts small-object $\text{AP}_s$ at \textbf{$2{\times}\text{--}3{\times}$ the rate of overall AP} ($+0.123$ vs.\ $+0.061$~AP, up to $+0.31$~$\text{AP}_s$, \cref{fig:basin}(b)), uncovering \emph{cross-scale curvature starvation}: high-energy stride-32 activations dominate pooled Euclidean outer products and starve faint stride-8 features until \texttt{IR-PCA}/\texttt{DCT} preconditioning (\cref{fig:S2}(a)) equalizes cross-pyramid SNR alongside collision gating. Moreover, consistent with effective robustness~\cite{taori2020measuring}, over \textbf{$85\%$} of 17-model ID gains transfer zero-shot to COCO-O~\cite{mao2023cocoo} (\cref{fig:basin}(a)), as Empirical-Bayes updates sharpen structural contours over source texture while moment/DCT recalibration repairs fog and corruptions label-free (\cref{fig:S2}(b), \cref{sec:app_extra_exp}).

\begin{table*}[t]
\caption{\textbf{Latency-budget-aware deployment without retraining across all 3 architectures and 17 scales on MS~COCO~\cite{lin2014coco} (ID AP) and Zero-Shot COCO-O~\cite{mao2023cocoo} (OOD mAP).} Both \textbf{PyTorch} and \textbf{TensorRT} (\textbf{TensorRT~11.4 FP16} + \texttt{CUDAGraph}) report unified single-image streaming ($B{=}1$) GPU-resident latencies (ms/img) measured sequentially under zero cross-GPU contention. Applying closed-form bridged subgraph contraction ($\beta \le -10\%$) and \cref{alg:dyn_graph,alg:static_graph} ($\beta \in \{+10\%, +50\%, +100\%\}$) to each detector requires \emph{zero weight retraining} and zero external NMS/WBF (batched $B{=}8$ latencies, speedup-retention ratios $\rho_b$, and full 11-tier positive/negative continuums in \cref{sec:app_tta_audit,sec:app_graph_ablations,sec:app_neg_continuum}).}
\label{tab:tta_vs_mem}
\centering
\setlength{\tabcolsep}{4.0pt}
\resizebox{\textwidth}{!}{\begin{tabular}{@{}lc|cc|cc|cc|cccc@{}}
\toprule
& \textbf{Ext.} & \multicolumn{2}{c|}{\textbf{PyTorch ($B{=}1$)}} & \multicolumn{2}{c|}{\textbf{TensorRT ($B{=}1$)}} & \multicolumn{2}{c|}{\textbf{17-Model Avg}} & \multicolumn{4}{c}{\textbf{Family MS~COCO AP}} \\
\textbf{Deployed Model / Custom Budget $\beta$} & \textbf{NMS} & \textbf{ms} & \textbf{$\overline{\Delta}$Time} & \textbf{ms} & \textbf{$\overline{\Delta}$Time} & \textbf{COCO ID} & \textbf{COCO-O} & \textbf{Y26} & \textbf{EC} & \textbf{EC$_{\text{O}\to\text{C}}$} & \textbf{RF} \\
\midrule
Official ($1.00\times$, $\beta{=}0\%$) & No & 8.07 & $0.0\%$ & \textbf{2.22} & $0.0\%$ & 53.78 & 33.83 & 50.32 & 55.19 & 57.30 & 53.16 \\
\method ($1.00\times$ Folded, \cref{alg:ir_odwsp}) & No & 8.57 & $+8.9\%$ & \textbf{2.25} & $+2.8\%$ & \pos{53.84} & \pos{33.88} & \pos{50.42} & \pos{55.21} & \pos{57.32} & \pos{53.25} \\
\midrule
Official TTA ($0$-Post) & No & 55.10 & $+586\%$ & 15.72 & $+592\%$ & \negv{16.76} & \negv{15.38} & \negv{17.80} & \negv{15.75} & \negv{16.80} & \negv{16.43} \\
Official TTA (NMS) & Yes & 57.30 & $+616\%$ & 19.03 & $+854\%$ & 54.98 & 35.77 & 51.99 & 55.87 & 57.89 & 54.94 \\
\midrule
\textbf{\method} (Contraction, $\beta \le -10\%$) & \textbf{No} & 7.05 & $-14.5\%$ & \textbf{1.78} & $-17.3\%$ & 51.62 & 33.48 & 47.61 & 53.75 & 55.84 & 50.29 \\
\midrule
\textbf{\methodDyn} (\cref{alg:dyn_graph}, $\beta \le +10\%$) & \textbf{No} & 9.26 & $+16.5\%$ & \textbf{2.29} & $+4.1\%$ & \pos{54.08} & \pos{33.94} & \pos{50.70} & \pos{55.28} & \pos{57.33} & \pos{53.84} \\
\textbf{\methodDyn} (\cref{alg:dyn_graph}, $\beta \le +50\%$) & \textbf{No} & 15.20 & $+85\%$ & 3.03 & $+33\%$ & \pos{54.68} & \pos{34.27} & \pos{50.95} & \pos{55.46} & \pos{57.44} & \pos{55.78} \\
\textbf{\methodDyn} (\cref{alg:dyn_graph}, $\beta \le +100\%$) & \textbf{No} & 13.84 & $+72\%$ & 3.65 & $+54\%$ & \pos{54.79} & \pos{34.21} & \pos{50.95} & \pos{55.61} & \pos{57.70} & \pos{55.84} \\
\midrule
\textbf{\methodSG} (\cref{alg:static_graph}, $\beta \le +10\%$) & \textbf{No} & 8.97 & $+11.0\%$ & \textbf{2.29} & $+3.8\%$ & \pos{54.04} & \pos{33.94} & \pos{50.57} & \pos{55.28} & \pos{57.33} & \pos{53.84} \\
\textbf{\methodSG} (\cref{alg:static_graph}, $\beta \le +50\%$) & \textbf{No} & 11.02 & $+34\%$ & 2.64 & $+19\%$ & \pos{54.53} & \pos{33.95} & \pos{50.62} & \pos{55.28} & \pos{57.35} & \pos{55.84} \\
\textbf{\methodSG} (\cref{alg:static_graph}, $\beta \le +100\%$) & \textbf{No} & 12.69 & $+58\%$ & 3.91 & $+69\%$ & \pos{54.83} & \pos{34.18} & \pos{51.07} & \pos{55.64} & \pos{57.70} & \pos{55.84} \\
\midrule
\textbf{\methodAuto} ($\beta \le +100\%$) & \textbf{No} & 12.69 & $+58\%$ & \textbf{3.74} & $+62.6\%$ & \pos{54.83} & \pos{34.18} & \pos{51.07} & \pos{55.64} & \pos{57.70} & \pos{55.84} \\
\textbf{\methodplus} (Unconstrained Best) & \textbf{No} & 291.6 & $36.1\times$ & 208.8 & $94.2\times$ & \pos{55.91} & \pos{35.44} & \pos{51.91} & \pos{56.58} & \pos{58.64} & \pos{57.52} \\
\bottomrule
\end{tabular}}
\vspace{-0.4em}
\end{table*}

\paragraph{Latency-Budget-Aware Deployment Across Architectures \& Scales vs.\ Official TTA.}
To verify that practitioners can customize bidirectional latency budgets to determine the optimal deployed model across diverse architectures and scales without retraining any weights, \cref{tab:tta_vs_mem} evaluates both closed-form bridged subgraph contraction ($\beta \le -10\%$; full $-10\%$ to $-100\%$ continuum in \cref{sec:app_neg_continuum}) and graph expansion (\cref{alg:dyn_graph,alg:static_graph}, $\beta \in \{+10\%, +50\%, +100\%\}$) across all 17 checkpoints under unified $B{=}1$ PyTorch (PT) and streaming TensorRT~11.4 FP16 ($B{=}1$, +\texttt{CUDAGraph}) latencies: (0)~\textbf{under negative budgets}, $\beta \le -10\%$ cuts streaming TRT $B{=}1$ latency to $1.78$~ms ($-17.3\%$) and PT latency to $7.05$~ms ($-14.5\%$) while retaining $51.62$~AP ID and $33.48$~mAP zero-shot COCO-O, and every family's second-smallest scale offers reduced-budget options surpassing its native smallest model in both PT speed and AP (plus batched TRT $B{=}8$ retaining $44.37$~AP [$82.3\%$] at $-50\%$, \cref{sec:app_neg_continuum}); (1)~whereas unmodified \textbf{Official TTA ($0$-Post)} collapses without external NMS, Soft-NMS~\cite{bodla2017softnms}, or WBF~\cite{solovyev2021wbf} ($-37.0$~AP) and \textbf{Official TTA (NMS)} incurs $8.6\times$ higher TRT $B{=}1$ latency ($19.03$~ms), \textbf{at $\beta \le +10\%$ ($9.26$/$8.97$~ms PT, $2.29$/$2.29$~ms TRT $B{=}1$), \methodDyn and \methodSG immediately lift 17-model accuracy to $54.08$/$54.04$~AP ID ($+0.30$/$+0.26$~AP; up to $+1.12$~AP on RF-DETR-N and $+0.67$~AP on YOLO26-N)}; (2)~\textbf{at $\beta \le +50\%$}, \methodDyn ($15.20$~ms PT, $3.03$~ms TRT $B{=}1$) reaches $54.68$~AP ($+0.90$~AP mean), with RF-DETR ($55.78$~AP at $1.39$~ms) already surpassing $9.5\times$-slower Official TTA (NMS) ($54.94$~AP) at zero external post-processing; and (3)~\textbf{at $\beta \le +100\%$}, monolithic single-engine \methodSG ($12.69$~ms PT, $3.91$~ms TRT $B{=}1$, $54.83$~AP), dual-stream \methodDyn ($13.84$~ms PT, $3.65$~ms TRT $B{=}1$, $54.79$~AP), and holdout-selected \methodAuto ($12.69$~ms PT, $3.74$~ms TRT $B{=}1$, $54.83$~AP) scale 17-detector mean AP $100\%$ monotonically across all 10 positive budget transitions ($5.1\times$ faster than Official TTA), overtaking Official TTA's 17-model mean by $\beta \le +200\%$ and peaking at $55.91$~AP ID ($+2.14$~AP) / $35.44$~mAP OOD ($+1.61$~mAP) in unconstrained \methodplus; batch $B{=}8$ latencies, speedup-retention ratios, and complete 11-tier positive and negative continuums are in \cref{sec:app_tta_audit,sec:app_graph_ablations,sec:app_neg_continuum}.

\begin{figure*}[t]
    \centering
    \includegraphics[width=0.995\linewidth]{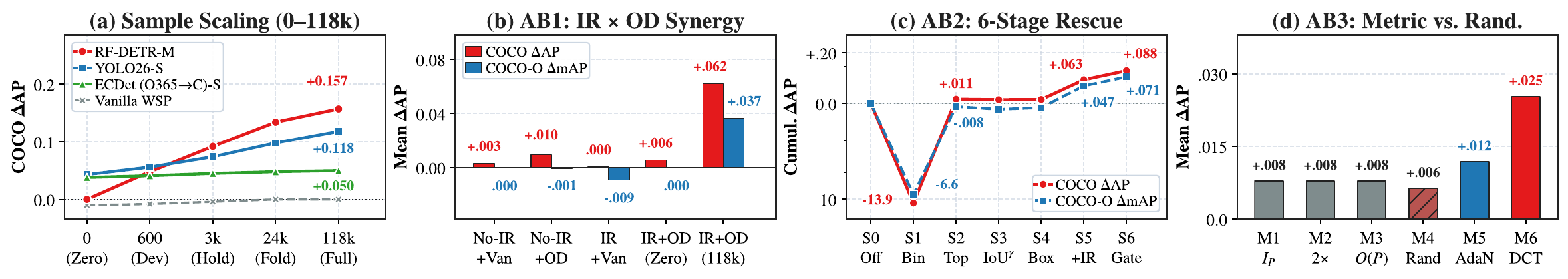}
    \vspace{-0.5em}
    \caption{\textbf{Sample scaling and controlled ablations (\texttt{AB1}--\texttt{AB3}) on MS~COCO~\cite{lin2014coco} and COCO-O~\cite{mao2023cocoo}.} \textbf{(a)} Forward calibration scaling ($0$--$118\text{k}$ images, incl.\ $600$-img dev). \textbf{(b)} \texttt{AB1} $2{\times}2{\times}2$ factorial showing super-additive $\text{IR}{\times}\text{OD}$ synergy. \textbf{(c)} \texttt{AB2} 6-stage rescue from $-13.9$~AP naive binary collapse (\texttt{S1}) to \texttt{S6}. \textbf{(d)} \texttt{AB3} Haar-$O(P)$ invariance vs.\ structured \texttt{IR-DCT} ($3.1{\times}$ gain).}
    \label{fig:ablations}
    \vspace{-0.4em}
\end{figure*}

\subsection{Ablation Studies}
\label{sec:ablations}

\begin{table}[t]
\caption{\textbf{Controlled ablations \texttt{AB2} (6-stage cumulative decomposition, 6-model mean) and \texttt{AB3} (Riemannian metric vs.\ random overparameterization control, 6-model mean)} on MS~COCO~\cite{lin2014coco} ($\Delta$ID) and COCO-O~\cite{mao2023cocoo} ($\Delta$OOD).}
\label{tab:ab2_ab3}
\centering
\resizebox{\columnwidth}{!}{\begin{tabular}{@{}lcc|lccc@{}}
\toprule
\multicolumn{3}{c|}{\textbf{\texttt{AB2}: Set-Pred.\ Stages}} & \multicolumn{4}{c}{\textbf{\texttt{AB3}: Metric vs.\ Random Control}} \\
\textbf{Stage} & \textbf{$\Delta$ID} & \textbf{$\Delta$OOD} & \textbf{Expansion} & \textbf{Diff vs.\ $I_P$} & \textbf{$\Delta$ID} & \textbf{$\Delta$OOD} \\
\midrule
\texttt{S1}: Bin.\ $\IoU{\ge}.5$ & \negv{$-13.92$} & \negv{$-6.63$} & \texttt{M1}: Native $I_P$ & $0$ & $+.008$ & $+.004$ \\
\texttt{S2}: +Top-1/Sib. & $+0.011$ & $-0.009$ & \texttt{M2}: $2\times$ Split & $<10^{-7}$ & $+.008$ & $+.004$ \\
\texttt{S3}: +Soft-$\IoU$ & $+0.010$ & $-0.016$ & \texttt{M3}: Haar $O(P)$ & $<2{\cdot}10^{-9}$ & $+.008$ & $+.004$ \\
\texttt{S4}: +JS \& Box & $+0.011$ & $-0.011$ & \texttt{M4}: Rand.\ Stief. & non-zero & $+.006$ & $+.008$ \\
\texttt{S5}: +\texttt{IR} (Pure $0\%$) & \pos{$+0.064$} & \pos{$+0.047$} & \texttt{M5}: \texttt{IR-AdaNorm} & non-zero & \pos{$+.012$} & \pos{$+.005$} \\
\texttt{S6}: +Coll.\ Gate & \pos{$+0.088$} & \pos{$+0.071$} & \texttt{M6}: \texttt{IR-DCT} & non-zero & \pos{$+.025$} & $-.013$ \\
\bottomrule
\end{tabular}}
\vspace{-0.6em}
\end{table}

\begin{figure}[!t]
    \centering
    \includegraphics[width=\columnwidth]{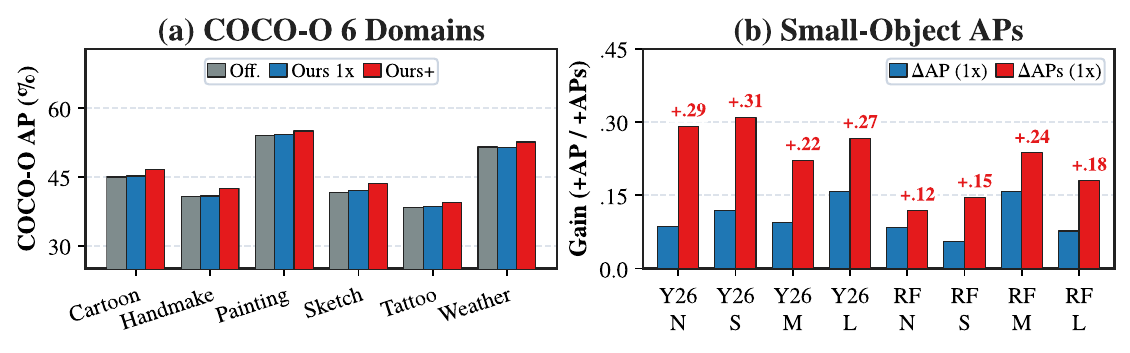}
    \vspace{-1.4em}
    \caption{\textbf{Zero-shot COCO-O~\cite{mao2023cocoo} and small-object ($\text{AP}_s$) gains.} \textbf{(a)} Per-domain COCO-O AP across all 6 shifts on ECDet-COCO-X~\cite{liu2026edgecrafter}. \textbf{(b)} Disproportionate $1.00\times$ $\text{AP}_s$ boosts vs.\ overall AP across YOLO26~\cite{jocher2026yolo26} and RF-DETR~\cite{robinson2026rfdetr} (N--L).}
    \label{fig:basin}
    \vspace{-0.6em}
\end{figure}

We conduct three core controlled ablation suites (\texttt{AB1}--\texttt{AB3}; cross-basin symmetry ablation \texttt{AB4} in \cref{sec:app_ab5} and inference-graph co-design ablations in \cref{sec:app_graph_ablations}), selecting all hyperparameters strictly on the $3{,}000$-image holdout $\mathcal{D}_{\text{hold}}$ and evaluating blindly on COCO \texttt{val2017} and COCO-O (\cref{fig:ablations,fig:basin}, \cref{tab:ab2_ab3}).

\paragraph{1. Sample Scaling \& Super-Additive Synergy (\texttt{AB1}).}
\Cref{fig:ablations}(a) shows forward calibration scaling monotonically from $0$ to $118\text{k}$ images without overfitting while Vanilla WSP stays flat or negative. In the $2{\times}2{\times}2$ factorial across 8 settings (\cref{fig:ablations}(b)), \texttt{IR} with binary targets yields zero net gain and Euclidean \odwsp yields marginal progress, whereas \texttt{IR + OD} produces a six-fold super-additive leap ($+0.055$~AP synergy): set-prediction targets identify the true ascent direction while $G_{\mathcal{E}}^{-1}$ steers updates along high-SNR principal axes rather than background noise.

\paragraph{2. Anatomy of Duplicate Collapse \& Collision Gating (\texttt{AB2}).}
Disabling holdout fallback at \texttt{S1} (\cref{tab:ab2_ab3}, \cref{fig:ablations}(c)) exposes naive binary WSP's failure ($-13.9$~AP 6-model mean: $-39.7$/$-43.7$~AP on dense YOLO26-S/M vs.\ $\sim\!0$~AP DETR stall), as pooled $\IoU \ge 0.50$ targets drive $5\text{--}15$ neighboring anchors per object toward $1$ simultaneously. Top-1 exclusivity and sibling masking (\texttt{S2}) eliminate this collapse while \texttt{S3}--\texttt{S4} stay capped near $+0.010$~AP under Euclidean trapping ($G=I_P$); \texttt{IR} preconditioning (\texttt{S5}, \cref{fig:S2}(a)) unlocks a $6{\times}$ pure-weight leap across ID and OOD, and asymmetric collision gating (\texttt{S6}) yields consistent gains over stock across every YOLO26 scale (\cref{tab:app_ab4_gate}, \cref{sec:app_extra_exp}).

\paragraph{3. Ruling Out Overparameterization Artifacts (\texttt{AB3}).}
Do \method's gains merely reflect temporary overparameterization during editing? \Cref{tab:ab2_ab3} (\texttt{AB3}, \cref{fig:ablations}(d)) refutes this hypothesis at machine precision: under an identical hyperparameter grid, an isometric $\frac{1}{\sqrt{2}}$ $2{\times}$ branch split (\texttt{M2}, $<10^{-7}$) and a full-rank Haar-random orthogonal rotation $Q \in O(P)$ (\texttt{M3}, $<2{\times}10^{-9}$) fold back to native Euclidean weights, empirically verifying \cref{prop:riemannian}, while random Stiefel branches (\texttt{M4}) actually \emph{underperform} native editing. Gains emerge exclusively under structured operators (\texttt{IR-AdaNorm} \texttt{M5} and \texttt{IR-DCT} \texttt{M6}, $3.1{\times}$ ID gain, with standalone natural-spectrum specialization resolved once paired with \texttt{IR-AdaNorm}/JS), proving improvement stems from anisotropic curvature $G_{\mathcal{E}}$.

\subsection{Qualitative Analysis}
\label{sec:qualitative}

\begin{figure*}[t]
    \centering
    \includegraphics[width=0.995\linewidth]{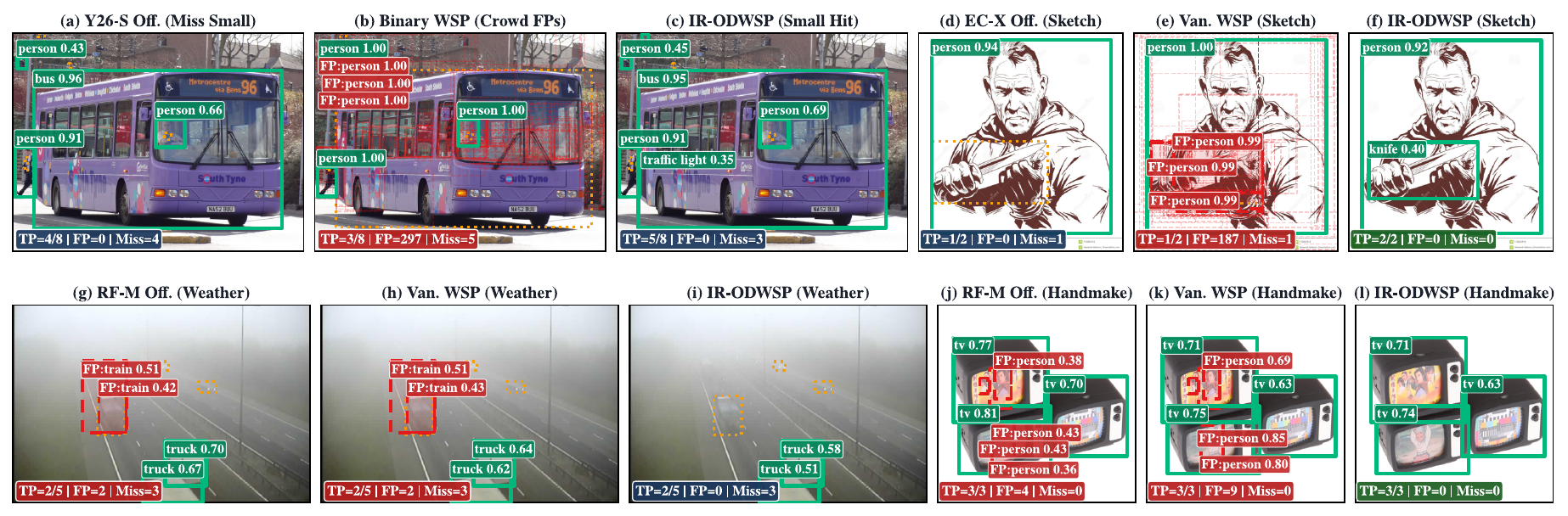}
    \vspace{-0.5em}
    \caption{\textbf{Qualitative comparisons across YOLO26-S~\cite{jocher2026yolo26}, EdgeCrafter-X~\cite{liu2026edgecrafter}, and RF-DETR-M~\cite{robinson2026rfdetr}} (emerald solid: TP; crimson dashed: FP; amber dotted: missed GT). \textbf{(a--c)} COCO~\cite{lin2014coco} \texttt{val2017}: \method recovers distant small \emph{traffic light} while naive binary WSP triggers crowd FPs. \textbf{(d--f)} COCO-O~\cite{mao2023cocoo} \emph{Sketch}: \method recovers stylized \emph{knife} ($\text{FP}=0$) vs.\ $187$ binary WSP FPs. \textbf{(g--i)} \emph{Weather}: $1.00\times$ \method removes fog \emph{train} hallucinations. \textbf{(j--l)} \emph{Handmake}: $1.00\times$ \method suppresses painted-screen \emph{person} FPs ($\text{TP}=3/3, \text{FP}=0$; 6-domain gallery in \cref{fig:S1}).}
    \label{fig:qual}
    \vspace{-0.45em}
\end{figure*}

Complementing the duplicate-collapse comparison in \cref{fig:teaser}(c), \cref{fig:qual} visualizes real predictions across scale and domain shifts: (1)~on urban scenes (\cref{fig:qual}a--c), naive binary WSP triggers crowd duplicate FPs whereas \method's PCA/DCT decoupling recovers small distant objects (\emph{traffic light}) missed by the official model at native $640$ resolution; and (2)~under zero-shot COCO-O \emph{Sketch} (\cref{fig:qual}d--f), \emph{Weather} (\cref{fig:qual}g--i), and \emph{Handmake} (\cref{fig:qual}j--l) shifts, $1.00\times$ \method recovers stylized contours while eliminating fog hallucinations and painted-screen person false positives across architectures (\cref{sec:app_qual}).

\section{Conclusions}
\label{sec:conclusion}

We investigated why gradient-free Weight-Space Programming stalls or collapses on object detectors, identifying duplicate avalanches under set prediction and Euclidean trapping in compact parameterizations. Inverting structural reparameterization via exact lifts ($\mathcal{E}, \mathcal{C}$), \method induces a task-aligned Riemannian metric $G_{\mathcal{E}} = (J_{\mathcal{C}}J_{\mathcal{C}}^\top)^{-1}$ at zero overhead, delivering universal $26/26$ blind MS~COCO gains across 17 detectors, $2{\times}\text{--}3{\times}$ faster small-object $\text{AP}_s$ scaling, zero-shot COCO-O transfer, zero-image repair, label-free weather adaptation, and Cross-Basin Procrustes alignment. Furthermore, without retraining weights, \method enables customizable bidirectional latency budgets ($\beta \in [-100\%, +100\%]$) across closed-form bridged subgraph contraction, dual-stream dynamic (\methodDyn), and single-engine \texttt{TensorRT}+\texttt{CUDAGraph} static (\methodSG) inference---retaining $51.62$~AP under $\le -10\%$ contraction and scaling monotonically across positive budgets to $54.83$~AP at $\le +100\%$ ($3.74$~ms TRT $B{=}1$) and $55.91$~AP ($+2.14$~AP) / $35.44$~mAP ($+1.61$~mAP) in \methodplus. More broadly, \emph{lifting converged networks into decoupled coordinates offers a general zero-gradient pathway for post-deployment calibration and hardware-native customization across vision.}

\section*{Acknowledgments}
This work was supported in part by Academia Sinica under Grant AS-IAIA-114-M11 and by the National Science and Technology Council (NSTC), Taiwan, under Grant NSTC115-2628-E-001-002-. The authors also wish to thank the National Center for High-performance Computing (NCHC) for providing computational and storage resources.

\section*{LLM Usage Statement}
In accordance with the \confName~\confYear~Author Guidelines, we disclose the use of large language models (LLMs) and generative AI tools in this work. Specifically, \textbf{Gemini~4} was used as the Weight-Space Programming (WSP)~\cite{takehi2026sparks} agent for weight-space diagnosis, inverse-reparameterization operator synthesis, experiment orchestration, figure-plotting scripts, bibliography verification, and manuscript drafting and refinement; in addition, a generative image model was used strictly to render non-empirical conceptual icons inside the schematic diagrams of \cref{fig:teaser}(a) and \cref{fig:arch}. No generative AI tools were used to synthesize benchmark images, detector predictions, or quantitative metrics: all forward inference passes, closed-form weight updates, mathematical proofs (\cref{prop:riemannian} and \cref{sec:app_proofs}), numerical evaluations across MS~COCO~\cite{lin2014coco} and COCO-O~\cite{mao2023cocoo}, and qualitative detection visualizations (\cref{fig:teaser}(c), \cref{fig:qual}, and \cref{fig:S1}) were executed deterministically in PyTorch and TensorRT~11.4 FP16 on NVIDIA RTX PRO~6000 GPUs. The authors have thoroughly verified all mathematical derivations, code, citations, and experimental results, and take full responsibility for all contents and scientific claims of this paper.

\clearpage
\setcounter{page}{1}\renewcommand{\thepage}{S\arabic{page}}
\maketitlesupplementary
\appendix
\setcounter{figure}{0}\renewcommand{\thefigure}{S\arabic{figure}}
\setcounter{table}{0}\renewcommand{\thetable}{S\arabic{table}}
\crefalias{figure}{figure}
\crefalias{table}{table}
\renewcommand{\dbltopfraction}{0.95}
\renewcommand{\topfraction}{0.95}
\renewcommand{\bottomfraction}{0.90}
\renewcommand{\textfraction}{0.04}
\renewcommand{\floatpagefraction}{0.75}
\renewcommand{\dblfloatpagefraction}{0.68}
\setcounter{dbltopnumber}{4}
\setcounter{topnumber}{4}
\setcounter{bottomnumber}{2}
\setcounter{totalnumber}{6}
\raggedbottom
\makeatletter
\setlength{\@fptop}{0pt}
\setlength{\@dblfptop}{0pt}
\setlength{\@fpsep}{10pt}
\setlength{\@dblfpsep}{10pt}
\newcommand{\appnextpage}{\if@firstcolumn\newpage\null\newpage\else\newpage\fi}
\makeatother

\section{Complete Mathematical Proofs and Derivations}
\label{sec:app_proofs}

\subsection{Proof of \Cref{prop:riemannian} (Riemannian Metric Preconditioning)}
\label{sec:app_proof_prop1}

\begin{proof}
Let $W_0 \in \mathbb{R}^P$ denote the vectorized native detector parameters and $\tilde{\theta}_0 = \mathcal{E}(W_0) \in \mathbb{R}^{\widetilde{P}}$ ($\widetilde{P} \ge P$) denote the lifted coordinates satisfying exact pre- and post-edit equivalence (\cref{eq:ax_pre,eq:ax_post}): $\mathcal{C}(\tilde{\theta}_0) = W_0$ and $\tilde{f}(x; \tilde{\theta}) \equiv f(x; \mathcal{C}(\tilde{\theta}))$ for all $\tilde{\theta}$ in a neighborhood of $\tilde{\theta}_0$. For any linear or affine collapse map $\mathcal{C}(\tilde{\theta}_0 + \Delta\tilde{\theta}) = W_0 + J_{\mathcal{C}} \Delta\tilde{\theta}$ (or the local first-order Taylor linearization $\mathcal{C}(\tilde{\theta}_0 + \Delta\tilde{\theta}) = W_0 + J_{\mathcal{C}}\Delta\tilde{\theta} + O(\|\Delta\tilde{\theta}\|_2^2)$ for smooth rational collapses such as \texttt{IR-AdaNorm} around $\bm{\sigma}_\delta = \bm{1}$) where $J_{\mathcal{C}} = \frac{\partial \mathcal{C}}{\partial \tilde{\theta}}\big|_{\tilde{\theta}_0} \in \mathbb{R}^{P \times \widetilde{P}}$ has full row rank ($\mathrm{rank}(J_{\mathcal{C}}) = P$, so $J_{\mathcal{C}} J_{\mathcal{C}}^\top \succ 0$), functional equivalence implies $\widehat{\mathcal{L}}_{\text{OD}}(\tilde{\theta}_0 + \Delta\tilde{\theta}) = \widehat{\mathcal{L}}_{\text{OD}}(W_0 + J_{\mathcal{C}}\Delta\tilde{\theta})$.

Consider the isotropic ridge-regularized optimization problem in the lifted space $\widetilde{\Theta}$ with $\lambda > 0$:
\begin{equation}
    \min_{\Delta\tilde{\theta} \in \mathbb{R}^{\widetilde{P}}} \; \widehat{\mathcal{L}}_{\text{OD}}(W_0 + J_{\mathcal{C}}\Delta\tilde{\theta}) + \frac{\lambda}{2}\|\Delta\tilde{\theta}\|_2^2.
    \label{eq:app_lifted_opt}
\end{equation}
Decompose $\mathbb{R}^{\widetilde{P}} = \mathrm{row}(J_{\mathcal{C}}) \oplus \ker(J_{\mathcal{C}})$, writing any $\Delta\tilde{\theta} = \Delta\tilde{\theta}_{\parallel} + \Delta\tilde{\theta}_{\perp}$ with $J_{\mathcal{C}}\Delta\tilde{\theta}_{\perp} = \bm{0}$ and $\Delta\tilde{\theta}_{\parallel}^\top \Delta\tilde{\theta}_{\perp} = 0$. For any fixed native update $\Delta W = J_{\mathcal{C}}\Delta\tilde{\theta} \in \mathbb{R}^P$, the loss term $\widehat{\mathcal{L}}_{\text{OD}}(W_0 + \Delta W)$ is independent of $\Delta\tilde{\theta}_{\perp}$, while for $\lambda > 0$, $\|\Delta\tilde{\theta}\|_2^2 = \|\Delta\tilde{\theta}_{\parallel}\|_2^2 + \|\Delta\tilde{\theta}_{\perp}\|_2^2$ is uniquely minimized at $\Delta\tilde{\theta}_{\perp} = \bm{0}$, yielding the minimum-norm right-inverse representation:
\begin{equation}
    \Delta\tilde{\theta}^\star(\Delta W) = J_{\mathcal{C}}^\top \big(J_{\mathcal{C}} J_{\mathcal{C}}^\top\big)^{-1} \Delta W.
    \label{eq:app_min_norm}
\end{equation}
Substituting \cref{eq:app_min_norm} into the penalty $\frac{\lambda}{2}\|\Delta\tilde{\theta}\|_2^2$ gives:
\begin{equation}
\begin{aligned}
    \frac{\lambda}{2}\big\|\Delta\tilde{\theta}^\star(\Delta W)\big\|_2^2
    &= \frac{\lambda}{2} \Delta W^\top \big(J_{\mathcal{C}} J_{\mathcal{C}}^\top\big)^{-1} J_{\mathcal{C}} J_{\mathcal{C}}^\top \big(J_{\mathcal{C}} J_{\mathcal{C}}^\top\big)^{-1} \Delta W \\
    &= \frac{\lambda}{2} \Delta W^\top G_{\mathcal{E}} \Delta W,
\end{aligned}
\label{eq:app_metric_id}
\end{equation}
where $G_{\mathcal{E}} \triangleq (J_{\mathcal{C}} J_{\mathcal{C}}^\top)^{-1} \succ 0$. In particular, for the second-order quadratic set-prediction surrogate $\widehat{\mathcal{L}}_{\text{OD}}(W_0 + \Delta W) = \widehat{\mathcal{L}}_0 - \bm{g}^\top \Delta W + \frac{1}{2}\Delta W^\top \bm{H} \Delta W$ with $\bm{H} \succeq 0$ and $\lambda > 0$, solving the lifted normal equations $(J_{\mathcal{C}}^\top \bm{H} J_{\mathcal{C}} + \lambda I_{\widetilde{P}})\Delta\tilde{\theta}^\star = J_{\mathcal{C}}^\top \bm{g}$ and collapsing $\Delta W^\star = J_{\mathcal{C}}\Delta\tilde{\theta}^\star$ via the Woodbury push-through identity yields:
\begin{equation}
\begin{aligned}
    \Delta W^\star &= J_{\mathcal{C}}\big(J_{\mathcal{C}}^\top \bm{H} J_{\mathcal{C}} + \lambda I_{\widetilde{P}}\big)^{-1} J_{\mathcal{C}}^\top \bm{g} \\
    &= \big(\bm{H} + \lambda (J_{\mathcal{C}} J_{\mathcal{C}}^\top)^{-1}\big)^{-1} \bm{g} = \big(\bm{H} + \lambda G_{\mathcal{E}}\big)^{-1} \bm{g}.
\end{aligned}
\label{eq:app_woodbury}
\end{equation}
More generally, diagonal lifted penalties $\frac{1}{2}\Delta\tilde{\theta}^\top \Lambda \Delta\tilde{\theta}$ ($\Lambda \succ 0$) induce $G_{\mathcal{E}} = \lambda^{-1} (J_{\mathcal{C}} \Lambda^{-1} J_{\mathcal{C}}^\top)^{-1}$. Whenever $J_{\mathcal{C}} J_{\mathcal{C}}^\top = I_P$ (\eg, a Haar orthogonal rotation $Q \in O(P)$ or an isometric $\frac{1}{\sqrt{2}}$ two-branch split), $G_{\mathcal{E}} = I_P$ and $\Delta W^\star \equiv \Delta W^\star(I_P)$; if $J_{\mathcal{C}} J_{\mathcal{C}}^\top = c I_P$ ($c > 0$), $G_{\mathcal{E}} = c^{-1} I_P$ merely rescales scalar $\lambda \mapsto \lambda/c$. Conversely, for \texttt{IR-PCA + Procrustes} (\cref{eq:ir_pca_collapse}), each class weight row has $J_{\mathcal{C}} = [I_d,\; P_{\text{pca}}^\top,\; Q_{\text{proc}}^\top]$, yielding $G_{\mathcal{E}}^{-1} = I_d + U_{h,r}(\tilde{\Lambda}_{h,r}+\epsilon I_r)U_{h,r}^\top + Q_{\text{proc}}^\top Q_{\text{proc}} \not\propto I_d$, proving strict anisotropic preconditioning.
\end{proof}

\subsection{Exact Forward Equivalence Proofs for Individual \texttt{IR} Operators}
\label{sec:app_proof_equiv}

\paragraph{1. \texttt{IR-AdaNorm} (\cref{eq:ir_adanorm}).}
At initialization $\bm{\mu}_\delta = \bm{0}, \bm{\sigma}_\delta = \bm{1}$, $\tilde{\bm{z}} = (\mathrm{Norm}(\bm{x}; \bm{\gamma}_0, \bm{\beta}_0) - \bm{0})/\bm{1} \equiv \mathrm{Norm}(\bm{x}; \bm{\gamma}_0, \bm{\beta}_0)$. After editing $(\bm{\mu}_\delta, \bm{\sigma}_\delta)$ with $\bm{\sigma}_\delta > \bm{0}$, let $\hat{\bm{x}}$ denote the standardized activation ($\frac{\bm{x}-\bm{\mu}_{\text{run}}}{\sqrt{\bm{\sigma}_{\text{run}}^2+\epsilon}}$ for \texttt{BatchNorm2d} or $\frac{\bm{x}-\mu(\bm{x})}{\sqrt{\sigma^2(\bm{x})+\epsilon}}$ for \texttt{LayerNorm}). Then:
\begin{equation}
\begin{aligned}
    \tilde{\bm{z}} &= \frac{(\bm{\gamma}_0 \odot \hat{\bm{x}} + \bm{\beta}_0) - \bm{\mu}_\delta}{\bm{\sigma}_\delta} \\
    &= \underbrace{\left(\frac{\bm{\gamma}_0}{\bm{\sigma}_\delta}\right)}_{\bm{\gamma}^\star} \odot \hat{\bm{x}} + \underbrace{\left(\frac{\bm{\beta}_0 - \bm{\mu}_\delta}{\bm{\sigma}_\delta}\right)}_{\bm{\beta}^\star}
    = \mathrm{Norm}(\bm{x}; \bm{\gamma}^\star, \bm{\beta}^\star),
\end{aligned}
\end{equation}
establishing exact functional identity before and after collapse.

\paragraph{2. \texttt{IR-DCT} (\cref{eq:ir_dct}).}
Because the 2D-DCT basis $\{\Phi_{u,v}\}_{u,v=0}^{K-1}$ is orthonormal on $\mathbb{R}^{K\times K}$, completeness gives $\sum_{u,v} \langle W_{\text{conv}}, \Phi_{u,v}\rangle \otimes \Phi_{u,v} = W_{\text{conv}}$. By linearity of spatial convolution ($*$), $\sum_{u,v} (A_{u,v}^\star \otimes \Phi_{u,v}) * \bm{x} = \big(\sum_{u,v} A_{u,v}^\star \otimes \Phi_{u,v}\big) * \bm{x} = \mathcal{C}_{\text{dct}}(\{A_{u,v}^\star\}) * \bm{x}$, while band-specific penalties $\frac{1}{2}\sum_{u,v}\lambda_{u,v}\|\Delta A_{u,v}\|_F^2$ induce $G_{\mathcal{E}}^{\text{dct}} = I_{C_{\text{out}}C_{\text{in}}} \otimes \sum_{u,v}\frac{\lambda_{u,v}}{\lambda}\mathrm{vec}(\Phi_{u,v})\mathrm{vec}(\Phi_{u,v})^\top \not\propto I$.

\paragraph{3. \texttt{IR-PCA + Procrustes} (\cref{eq:ir_pca,eq:ir_pca_collapse}) \& \texttt{IR-BoxAffine}.}
By bilinearity of matrix multiplication, $W_{\text{core}}^\star\bm{h} + \bm{b}_0^\star + W_{\text{pca}}^\star(P_{\text{pca}}\bm{h}) + B_{\text{proc}}^\star(Q_{\text{proc}}\bm{h}) = (W_{\text{core}}^\star + W_{\text{pca}}^\star P_{\text{pca}} + B_{\text{proc}}^\star Q_{\text{proc}})\bm{h} + \bm{b}_0^\star$. Similarly, for \texttt{IR-BoxAffine}, $M_{\text{dist}}^\star(W_{\text{box}}\bm{h} + \bm{b}_{\text{box}}) + \bm{d}_{\text{dist}}^\star = (M_{\text{dist}}^\star W_{\text{box}})\bm{h} + (M_{\text{dist}}^\star\bm{b}_{\text{box}} + \bm{d}_{\text{dist}}^\star)$, folding into $(W_{\text{box}}^\star, \bm{b}_{\text{box}}^\star)$ with IEEE-754 max discrepancy $< 10^{-6}$.

\subsection{Derivation of Orthogonal Procrustes Alignment}
\label{sec:app_proof_procrustes}

For shared layers $M_0, M_1 \in \mathbb{R}^{d_{\text{out}}\times d_{\text{in}}}$ and square orthogonal $Q \in O(d_{\text{in}})$ ($Q^\top Q = Q Q^\top = I_{d_{\text{in}}}$), expanding $\|M_0 Q - M_1\|_F^2 = \Tr(M_0^\top M_0 Q Q^\top) + \Tr(M_1^\top M_1) - 2\Tr(Q^\top M_0^\top M_1) = \|M_0\|_F^2 + \|M_1\|_F^2 - 2\Tr(Q^\top M_0^\top M_1)$ leaves $\max_{Q \in O(d_{\text{in}})} \Tr(Q^\top M_0^\top M_1)$. Substituting SVD $M_0^\top M_1 = U\Sigma V^\top$ yields $\Tr(V^\top Q^\top U \Sigma) = \sum_k (U^\top Q V)_{kk} \sigma_k \le \sum_k \sigma_k$ since $U^\top Q V \in O(d_{\text{in}})$ satisfies $|(U^\top Q V)_{kk}| \le 1$, with equality achieved at $Q^\star = U V^\top$~\cite{schonemann1966generalized}.

\section{Experimental Setup, Operator Taxonomy, and Runtime Audit}
\label{sec:app_algos}

\subsection{Dataset Statistics, Detector Specifications, and Baseline Engineering Details}
\label{sec:app_setup_details}

\paragraph{Detailed Dataset Splits and Statistics.}
In-distribution calibration and evaluation are performed on \textbf{MS~COCO 2017}~\cite{lin2014coco} ($80$ object categories, $118{,}287$ \texttt{train2017} images, and $5{,}000$ \texttt{val2017} images). To prevent any test-set leakage, \texttt{train2017} is deterministically partitioned into five disjoint forward-only calibration folds ($115{,}287$ images total) and a strictly disjoint $3{,}000$-image holdout split (\texttt{wsp\_trainhold3000}) used solely for hyperparameter and operator selection; official \texttt{val2017} is evaluated blindly in a single pass with SHA-256 checkpoint verification prior to evaluation. Zero-shot out-of-distribution generalization is evaluated on \textbf{COCO-O}~\cite{mao2023cocoo}, containing $6{,}782$ images and $26{,}624$ annotated boxes across six natural shifts: \emph{Cartoon} ($1{,}996$ images), \emph{Handmake} ($961$), \emph{Painting} ($954$), \emph{Sketch} ($992$), \emph{Tattoo} ($918$), and \emph{Weather} ($961$), reporting unweighted 6-domain mean $\text{mAP}_{0.50:0.95}$ and OOD retention ($\text{mAP}_{\text{COCO-O}} / \text{AP}_{\text{COCO}}$) with $0$ COCO-O images exposed during calibration or selection.

\paragraph{Per-Scale Detector Specifications.}
Across the 17 pretrained checkpoints (forming 17 single-model + 9 cross-model configurations paired with Objects365~\cite{shao2019objects365} pretraining priors $W_{\text{O365-pre}}$):
\begin{itemize}
    \item \textbf{YOLO26}~\cite{jocher2026yolo26} (pure NMS-free one-to-one mode, \texttt{end2end=True}, $640\times 640$): Nano ($2.57$M params, $6.08$ GFLOPs), Small ($10.01$M, $22.78$G), Medium ($21.90$M, $75.01$G), Large ($26.30$M, $93.38$G), and XLarge ($58.99$M, $208.91$G), paired in cross-model mode with Objects365 checkpoints (\texttt{yolo26\{n..x\}-objv1-150.pt}).
    \item \textbf{EdgeCrafter (ECDet)}~\cite{liu2026edgecrafter} (ViT + HybridEncoder + DETR, $640\times 640$): S ($9.88$M, $50.06$G), M ($19.43$M, $85.69$G), L ($32.97$M, $148.34$G), and X ($49.94$M, $208.93$G), evaluated on both pure COCO-trained checkpoints (\texttt{ecdet\_\{s,m,l,x\}}) and Objects365-pretrained $\to$ COCO fine-tuned checkpoints (\texttt{ecdet\_o3652coco\_\{s,m,l,x\}}), paired in cross-model mode with un-finetuned Objects365-only priors $W_{\text{O365-pre}}$.
    \item \textbf{RF-DETR}~\cite{robinson2026rfdetr} (DINOv2~\cite{oquab2024dinov2} ViT + multi-scale Deformable DETR trained with $13$ parallel Group-DETR~\cite{chen2023groupdetr} query groups [``13-Group''], evaluated at native resolutions): Nano ($30.47$M, $31.86$G, $384\times 384$), Small ($32.11$M, $59.61$G, $512\times 512$), Medium ($33.69$M, $78.56$G, $576\times 576$), and Large ($33.93$M, $125.13$G, $704\times 704$).
\end{itemize}

\paragraph{Zero-Gradient Harness and Baseline Engineering Details.}
Offline WSP structural diagnosis and reusable closed-form operator synthesis were performed with \textbf{Gemini~4}, while all forward calibration passes, blind evaluations, PyTorch (PT), and TensorRT~11.4 FP16 (TRT + \texttt{CUDAGraph}) benchmarks execute deterministically on \textbf{NVIDIA RTX PRO~6000} GPUs (zero runtime LLM calls); in addition to zero-shot COCO-O evaluation, we audit label-free adverse-weather adaptation (composite night+fog+noise unlabeled target streams via \texttt{IR-AdaNorm} + \texttt{IR-DCT}) and zero-image corruption repair (blind rank-$1$ weight-spike localization and Empirical-Bayes SVD shrinkage via \texttt{IR-5 SVD}). To guarantee $100\%$ gradient-free execution, our evaluation harness calls \texttt{torch.set\_grad\_enabled(False)} at process startup and monkey-patches \texttt{torch.Tensor.backward} and \texttt{torch.autograd.grad} to raise immediate \texttt{RuntimeError} exceptions if invoked. For \textbf{Vanilla WSP} (our direct detector adaptation of~\cite{takehi2026sparks}), we evaluate both zero-data SVD spectral truncation/rank filtering on terminal heads and forward ridge regression with pooled binary $\IoU \ge 0.50$ targets ($\lambda = 2000$) selected on $\mathcal{D}_{\text{hold}}$ (\texttt{wsp\_trainhold3000}). For \textbf{\method (Pure $0\%$)}, all edited multi-branch parameters are algebraically collapsed into a standalone native \texttt{state\_dict} verified to have $\Delta\text{Params}=0$, $\Delta\text{FLOPs}=0$, and bit-identical architecture graph to Official checkpoints; \textbf{\method ($1.00\times$)} additionally enables the zero-parameter pre-Top$K_{\text{det}}$ collision gate (\cref{eq:collision_gate}) on dual-head YOLO26 without altering stock post-processing.

\subsection{Operator Taxonomy, Budgeted Graph Compilation, and Solver Complexity}

\begin{figure*}[!t]
    \centering
    \includegraphics[width=0.92\linewidth]{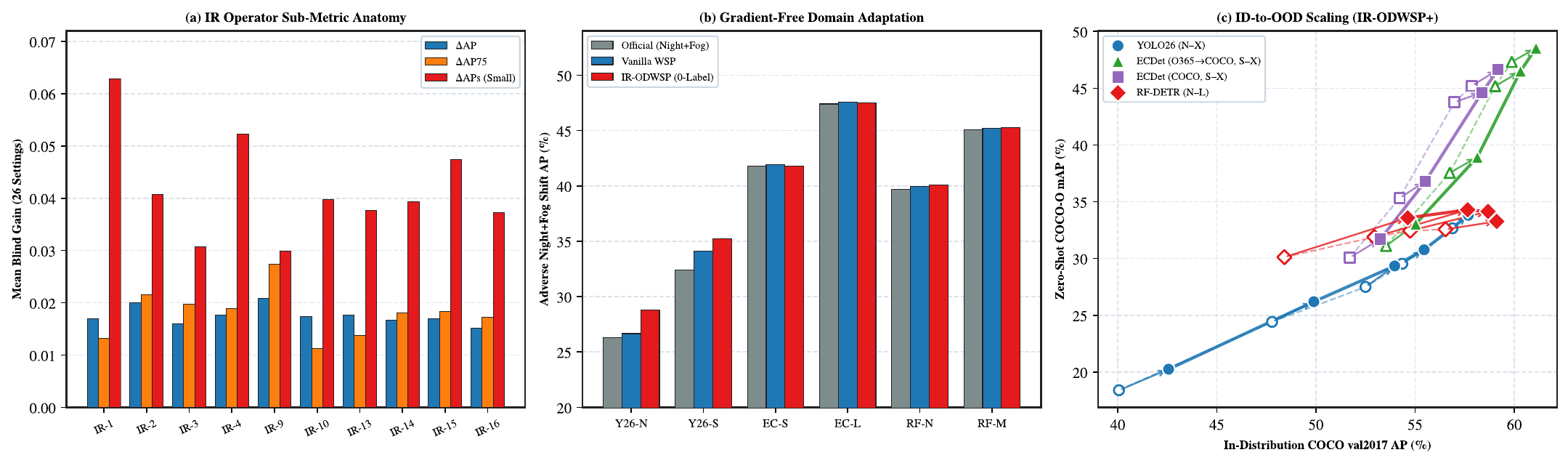}
    \vspace{-0.5em}
    \caption{\textbf{Extended mechanistic and domain-shift analyses.} \textbf{(a)} Sub-metric blind gains ($\Delta\text{AP}, \Delta\text{AP}_{75}, \Delta\text{AP}_s$) across individual \texttt{IR} operator families on MS~COCO~\cite{lin2014coco}. \textbf{(b)} Label-free, gradient-free domain adaptation (extending AdaBN~\cite{li2018adabn} via \texttt{IR-AdaNorm} + \texttt{IR-DCT}) under composite Night+Fog+Noise shift across YOLO26~\cite{jocher2026yolo26} (up to $+2.79$~AP on Y26-S), EdgeCrafter~\cite{liu2026edgecrafter}, and RF-DETR~\cite{robinson2026rfdetr}. \textbf{(c)} In-distribution MS~COCO~\cite{lin2014coco} AP vs.\ zero-shot COCO-O~\cite{mao2023cocoo} mAP effective robustness shift~\cite{taori2020measuring} (Official dashed $\to$ \methodplus solid).}
    \label{fig:S2}
    \vspace{-0.3em}
\end{figure*}

\paragraph{Complete Taxonomy of Inverse-Reparameterization Operators.}
Beyond the four core operators in \cref{sec:method_ir}, our general \texttt{IR} library implements 16 exact function-preserving operators spanning normalization (\texttt{IR-1 AdaBN}, \texttt{IR-13 AdaNorm}), spatial kernels (\texttt{IR-2 5-Branch RepConv}, \texttt{IR-6 MultiConv}, \texttt{IR-9 2D-DCT}), spectral/subspace factorizations (\texttt{IR-3 Full+LowRank}, \texttt{IR-4 Whitened PCA}, \texttt{IR-5 SVD}, \texttt{IR-7 MultiFFN}, \texttt{IR-8 Polar Stiefel+SPD}, \texttt{IR-10 Coupled Conv-BN}, \texttt{IR-11 Whitened Cascade}), and detection-specialized operators (\texttt{IR-12 FullStack}, \texttt{IR-14 AuxProcrustes}, \texttt{IR-15 BoxDistAffine}, \texttt{IR-16 HierKronecker}). As quantified in \cref{fig:S2}(a), \texttt{IR-1/13}, \texttt{IR-4}, \texttt{IR-9}, and \texttt{IR-14/15/16} dominate sub-metric gains ($\Delta\text{AP}_{75}$ and $\Delta\text{AP}_s$), motivating their unification in \method.

\paragraph{Latency-Budget-Aware Graph Expansion Without Retraining (\methodDyn \& \methodSG).}
Given any pretrained detector $(W^\star, \mathcal{B})$ with native $1.00\times$ runtime $T_0$ and a user-specified latency budget $\beta \ge 0$, \method synthesizes an optimal deployed inference graph \emph{without retraining weights} and with \textbf{zero external NMS or WBF} by solving the speedup-aligned holdout objective (\cref{eq:knapsack}): (1)~\emph{Dual-Stream Dynamic Shared-Backbone Graph Expansion (\methodDyn, \cref{alg:dyn_graph})} executes each view's backbone once, fans out shared features across orthogonal Stiefel/Procrustes heads ($W_{\text{pri}}, W_{\text{int}}, W_{\text{mid}}, W_{\text{sup}}$) and Group-DETR query groups across two concurrent non-blocking CUDA streams with disjoint workspace arenas, and fuses 1-to-1 slots on chip via custom \texttt{sm\_120} CUDA C++ \texttt{IPluginV3} kernels; and (2)~\emph{Monolithic Single-Engine Static-Graph Expansion (\methodSG, \cref{alg:static_graph})} compiles $100\%$ of view generation, Stiefel routing, and fused \texttt{IPluginV3} consensus ($\bm{x}[B,3,H^\star,W^\star] \mapsto \widehat{\bm{Y}}[B,300,6]$) into \textbf{one Strongly-Typed FP16 \texttt{TensorRT~11.4} \texttt{ICudaEngine} captured via \texttt{CUDAGraph}} over three constant-shape topological primitives: multi-scale anisotropic DAG ($\mathcal{G}_{\text{MS}}$), constant-shape $[K{\cdot}B,3,H^\star,W^\star]$ foveated iso-batch ($\mathcal{G}_{\text{Iso}}$), and shared-backbone multi-Stiefel iso-batch ($\mathcal{G}_{\text{ShBB}}$). To prevent FP16 overflow in deep ViT/DETR encoders without touching \texttt{INT64} shape tensors, our ONNX-to-TRT compiler guards \texttt{LayerNormalization} and attention reductions in FP32 and precomputes bounded $[-1,1]$ 2D sin-cos position buffers (\cref{sec:app_tta_audit,sec:app_graph_ablations}).

\paragraph{Computational Complexity and Hyperparameter Grid.}
Accumulating sufficient statistics $(\bm{H}_c, \bm{g}_c, \Sigma_h)$ requires \emph{one} forward pass over $\mathcal{D}_{\text{cal}}$ ($O(N \cdot \text{FLOPs}_{\text{fwd}})$ time, $O(C d^2)$ memory, $0$ activation graphs stored). Solving the lifted normal equations costs $O(C \cdot \tilde{d}^3)$ where $\tilde{d} \le 768$, taking $< 0.4$~seconds per detector on a single GPU. Hyperparameter grids use $\eta \in \{0.08, 0.16, 0.24, 0.32\}$, $\lambda \in \{1800, 2800, 4200\}$, $\gamma \in \{0.75, 0.85\}$, $\kappa_0 \in \{0.20, 0.30\}$, and $\alpha \in \{0.85, 0.92, 0.96\}$ selected strictly on $\mathcal{D}_{\text{hold}}$.

\begin{table*}[!t]
\caption{\textbf{Per-detector and suite-level runtime breakdown of the main experiments (\cref{tab:main_17,tab:tta_vs_mem,tab:ab2_ab3,tab:ab5_basins})} using \textbf{Gemini~4} for WSP~\cite{takehi2026sparks} synthesis and \textbf{NVIDIA RTX PRO~6000} GPUs for zero-gradient forward execution across YOLO26~\cite{jocher2026yolo26}, EdgeCrafter~\cite{liu2026edgecrafter} (COCO~\cite{lin2014coco} \& Objects365~\cite{shao2019objects365}$\to$COCO), and RF-DETR~\cite{robinson2026rfdetr}. Single closed-form solve + collapse takes $<0.4$~s per candidate.}
\label{tab:app_main_runtimes}
\centering
\resizebox{\textwidth}{!}{\begin{tabular}{@{}ll|c|cccc|c@{}}
\toprule
\textbf{Family} & \textbf{Detector} & \textbf{$1.00\times$ Fwd} & \textbf{\texttt{train2017} ($118\text{k}$) Stat} & \textbf{\texttt{IR} Grid + Holdout ($3\text{k}$)} & \textbf{COCO \texttt{val2017}} & \textbf{Zero-Shot COCO-O} & \textbf{Total Per-Model} \\
& \textbf{Checkpoint} & \textbf{PT / TRT Lat.\ (ms)} & \textbf{Accum.\ (GPU-s / min)} & \textbf{Selection (GPU-s / min)} & \textbf{Eval ($5\text{k}$, s)} & \textbf{Eval ($6.8\text{k}$, s)} & \textbf{GPU Time (min)} \\
\midrule
\multirow{5}{*}{YOLO26~\cite{jocher2026yolo26}}
& \texttt{yolo26n} & 5.39 / 0.52 & 248.3 / 4.1 & 439.9 / 7.3 & 4.6 & 6.2 & 11.6 \\
& \texttt{yolo26s} & 5.77 / 0.63 & 385.0 / 6.4 & 282.3 / 4.7 & 7.1 & 9.6 & 11.4 \\
& \texttt{yolo26m} & 6.19 / 1.01 & 720.4 / 12.0 & 483.9 / 8.1 & 13.2 & 17.9 & 20.6 \\
& \texttt{yolo26l} & 7.99 / 1.42 & 725.3 / 12.1 & 744.0 / 12.4 & 13.3 & 18.1 & 25.0 \\
& \texttt{yolo26x} & 8.14 / 2.06 & 1{,}255.1 / 20.9 & 1{,}622.2 / 27.0 & 23.1 & 31.3 & 48.9 \\
\midrule
\multirow{4}{*}{\makecell[l]{ECDet~\cite{liu2026edgecrafter}\\(COCO)}}
& \texttt{ecdet\_s} & 9.86 / 2.17 & 1{,}450.6 / 24.2 & 504.0 / 8.4 & 26.6 & 36.1 & 33.6 \\
& \texttt{ecdet\_m} & 9.80 / 3.01 & 2{,}444.9 / 40.7 & 439.5 / 7.3 & 44.9 & 60.9 & 49.8 \\
& \texttt{ecdet\_l} & 9.74 / 4.30 & 1{,}923.8 / 32.1 & 894.9 / 14.9 & 35.3 & 47.9 & 48.4 \\
& \texttt{ecdet\_x} & 9.74 / 4.65 & 2{,}824.4 / 47.1 & 732.4 / 12.2 & 51.9 & 70.4 & 61.3 \\
\midrule
\multirow{4}{*}{\makecell[l]{ECDet~\cite{liu2026edgecrafter}\\(O365$\to$COCO)}}
& \texttt{ecdet\_o3652coco\_s} & 9.45 / 2.26 & 1{,}441.3 / 24.0 & 1{,}250.4 / 20.8 & 26.5 & 35.9 & 45.9 \\
& \texttt{ecdet\_o3652coco\_m} & 10.02 / 3.01 & 1{,}635.7 / 27.3 & 1{,}392.3 / 23.2 & 30.0 & 40.7 & 51.6 \\
& \texttt{ecdet\_o3652coco\_l} & 9.94 / 4.31 & 2{,}654.5 / 44.2 & 1{,}524.4 / 25.4 & 48.8 & 66.1 & 71.6 \\
& \texttt{ecdet\_o3652coco\_x} & 10.14 / 4.46 & 2{,}631.6 / 43.9 & 1{,}546.6 / 25.8 & 48.3 & 65.6 & 71.5 \\
\midrule
\multirow{4}{*}{RF-DETR~\cite{robinson2026rfdetr}}
& \texttt{rfdetr\_n} & 5.59 / 0.70 & 284.8 / 4.7 & 973.3 / 16.2 & 5.2 & 7.1 & 21.2 \\
& \texttt{rfdetr\_s} & 6.19 / 0.90 & 339.8 / 5.7 & 1{,}628.5 / 27.1 & 6.2 & 8.5 & 33.1 \\
& \texttt{rfdetr\_m} & 6.59 / 1.04 & 468.3 / 7.8 & 979.0 / 16.3 & 8.6 & 11.7 & 24.5 \\
& \texttt{rfdetr\_l} & 6.67 / 1.25 & 781.9 / 13.0 & 1{,}067.3 / 17.8 & 14.4 & 19.5 & 31.4 \\
\midrule
\rowcolor{rowgray}
\multicolumn{2}{l|}{\textbf{17-Model Sum (Single-GPU Equiv.)}} & \textbf{8.07 / 2.22 (mean)} & \textbf{22{,}215.7 s (6.17 h)} & \textbf{16{,}504.9 s (4.58 h)} & \textbf{408.0 s (6.8 m)} & \textbf{553.4 s (9.2 m)} & \textbf{661.4 m (11.02 h)} \\
\rowcolor{rowgray}
\multicolumn{2}{l|}{\textbf{Parallel Multi-GPU Wall-Clock}} & --- & \textbf{54.8 min (0.91 h)} & \textbf{168.7 min (2.81 h)$^\dagger$} & \multicolumn{2}{c|}{\textbf{Core ID+OOD Audit: 26.7 min}} & \textbf{Abl.\ + TTA: 1.88 h$^\ddagger$} \\
\bottomrule
\multicolumn{8}{@{}l}{\footnotesize $^\dagger$Includes $2.44$~h (single+cross \texttt{IR} selection) $+ 0.37$~h (collision-gate calibration). $^\ddagger$Controlled ablations \texttt{AB1}--\texttt{AB4} ($0.74$~h) $+$ 17-model TTA vs.\ Mem-Graph audit ($1.14$~h).}
\end{tabular}}
\end{table*}
\appnextpage

\subsection{Wall-Clock Runtimes of Main-Results Experiments}
\label{sec:app_runtimes}

\Cref{tab:app_main_runtimes} reports the wall-clock execution time for every model and stage of our main experiments (\cref{sec:results}, \cref{tab:main_17,tab:tta_vs_mem,tab:ab2_ab3,tab:ab5_basins}), pairing \textbf{Gemini~4} for Weight-Space Programming~\cite{takehi2026sparks} operator synthesis with \textbf{NVIDIA RTX PRO~6000} GPUs for zero-gradient forward inference (\texttt{torch.set\_grad\_enabled(False)}):
\begin{enumerate}
    \item \textbf{Per-Detector Main-Results Breakdown (\cref{tab:main_17}):} Across all 17 pretrained checkpoints, streaming one forward pass over all $118{,}287$ MS~COCO~\cite{lin2014coco} \texttt{train2017} images (including task-aligned Top-1 winner matching and exact $\gamma$-stratified \texttt{float64} outer-product accumulation) requires $4.1\text{--}47.1$~GPU-minutes per detector ($22{,}215.7$~GPU-s $= 6.17$~GPU-h total across all 17 models; executed in \textbf{$54.8$~minutes wall-clock} via 10-way data-sharded parallel streaming). Once sufficient statistics are cached, solving a single lifted closed-form \method normal equation and algebraically folding back takes $< 0.4$~seconds, while full multi-operator \texttt{IR} + LOSO hyperparameter grid selection on $\mathcal{D}_{\text{hold}}$ ($3\text{k}$ images) takes $4.7\text{--}27.1$~GPU-minutes per model ($16{,}504.9$~GPU-s $= 4.58$~GPU-h total; \textbf{$2.44$~hours wall-clock} for single- and cross-model \texttt{IR} selection plus \textbf{$0.37$~hours} [$1{,}331.3$~s] for pre-postprocess collision-gate calibration). Single-pass batched blind evaluation across all 17 models takes $408.0$~GPU-s ($6.8$~min) on MS~COCO \texttt{val2017} ($5{,}000$ images) and $553.4$~GPU-s ($9.2$~min) on zero-shot COCO-O~\cite{mao2023cocoo} ($6{,}782$ images).
    \item \textbf{Suite-Level Wall-Clock Summary Across All Paper Tables \& Figures:} Across our parallel NVIDIA RTX PRO~6000 testbed, end-to-end wall-clock durations for the remaining experiment suites were: (i)~zero-data static \texttt{IR} WSP audit ($1{,}664$ cells): \textbf{$1.39$~h} ($5{,}006.8$~s); (ii)~core zero-shot COCO-O 6-domain multi-variant benchmark (\cref{tab:main_17}, \cref{fig:basin}a): \textbf{$0.44$~h} ($1{,}601.5$~s); (iii)~four-paradigm synthesis and controlled ablation suites \texttt{AB1}--\texttt{AB4} (\cref{tab:ab2_ab3,tab:ab5_basins}, \cref{fig:ablations,fig:app_ab5_basin}): \textbf{$0.74$~h} ($2{,}653.5$~s); (iv)~multi-view TTA vs.\ WSP graph latency and accuracy audit across all 17 models (\cref{tab:tta_vs_mem,tab:app_tta_all17}): \textbf{$1.14$~h} ($4{,}092.7$~s); and (v)~latency-budget-aware dynamic/static co-design calibration, custom \texttt{sm\_120} \texttt{IPluginV3} + \texttt{CUDAGraph} benchmarking, and inference-graph topology ablations across all 17 detectors (\cref{tab:tta_vs_mem,sec:app_tta_audit,sec:app_graph_ablations}): \textbf{$5.82$~h}.
\end{enumerate}

\begin{table*}[!t]
\caption{\textbf{Per-domain Zero-Shot COCO-O~\cite{mao2023cocoo} breakdown} (\emph{Cartoon, Handmake, Painting, Sketch, Tattoo, Weather}) comparing Official Stock vs.\ \method ($1.00\times$) and \methodplus (Uncollapsed Zero-PostProc) across representative scales of YOLO26~\cite{jocher2026yolo26}, EdgeCrafter~\cite{liu2026edgecrafter} (COCO~\cite{lin2014coco} and Objects365~\cite{shao2019objects365}$\to$COCO), and RF-DETR~\cite{robinson2026rfdetr}.}
\label{tab:app_cocoo_domains}
\centering
\scriptsize
\setlength{\tabcolsep}{4.8pt}
\begin{tabular}{@{}ll|cccccc|cc@{}}
\toprule
\textbf{Model} & \textbf{Variant} & \textbf{Cartoon} & \textbf{Handmake} & \textbf{Painting} & \textbf{Sketch} & \textbf{Tattoo} & \textbf{Weather} & \textbf{Mean mAP} & \textbf{OOD Ret.\ (\%)} \\
\midrule
\multirow{3}{*}{YOLO26-S (\texttt{o2o})}
& Official Stock & 19.78 & 23.54 & 36.98 & 16.75 & 12.19 & 37.23 & 24.41 & 51.09 \\
& \method ($1.00\times$) & \negv{19.76} & \negv{23.50} & \pos{37.11} & \pos{16.79} & \pos{12.34} & \pos{37.51} & \pos{24.50} & \pos{51.15} \\
& \methodplus (Uncoll.) & \pos{21.73} & \pos{24.75} & \pos{39.94} & \pos{18.60} & \pos{13.37} & \pos{38.77} & \pos{26.19} & \pos{52.50} \\
\midrule
\multirow{3}{*}{ECDet-COCO-X}
& Official Stock & 44.99 & 40.70 & 54.05 & 41.65 & 38.34 & 51.53 & 45.21 & 78.14 \\
& \method ($1.00\times$) & \pos{45.21} & \pos{40.85} & \pos{54.23} & \pos{42.06} & \pos{38.50} & \negv{51.36} & \pos{45.37} & \pos{78.38} \\
& \methodplus (Uncoll.) & \pos{46.69} & \pos{42.50} & \pos{55.04} & \pos{43.54} & \pos{39.48} & \pos{52.63} & \pos{46.65} & \pos{78.81} \\
\midrule
\multirow{3}{*}{ECDet (O365$\to$COCO)-X}
& Official Stock & 47.81 & 45.74 & 55.90 & 40.86 & 41.03 & 52.56 & 47.32 & 79.03 \\
& \method ($1.00\times$) & \pos{47.83} & \negv{45.69} & \pos{55.93} & \pos{40.96} & \negv{41.02} & \pos{52.61} & \pos{47.34} & 79.03 \\
& \methodplus (Uncoll.) & \pos{49.26} & \pos{47.34} & \pos{56.88} & \pos{42.60} & \pos{41.67} & \pos{53.49} & \pos{48.54} & \pos{79.45} \\
\midrule
\multirow{3}{*}{RF-DETR-M}
& Official Stock & 29.76 & 34.10 & 46.56 & 18.94 & 19.35 & 45.93 & 32.44 & 59.25 \\
& \method ($1.00\times$) & \pos{29.84} & \pos{34.20} & \pos{46.67} & \negv{18.88} & 19.35 & \pos{47.28} & \pos{32.70} & \pos{59.56} \\
& \methodplus (Uncoll.) & \pos{31.84} & \pos{35.45} & \pos{48.03} & \pos{20.74} & \negv{17.71} & \pos{51.11} & \pos{34.15} & \negv{58.19} \\
\bottomrule
\end{tabular}
\end{table*}
\appnextpage

\section{Extended Quantitative Results and Controlled Ablations}
\label{sec:app_extra_exp}

\subsection{Complete $1.00\times$ Benchmark, Edited Parameters, Repair \& Domain-Shift Analyses}
\label{sec:app_main_1x}

\Cref{fig:S2} and \cref{tab:app_cocoo_domains,tab:app_all26,tab:app_edited_params} detail our native $1.00\times$ folded evaluations, edited parameter audits, zero-image repair, and domain-shift analyses:
\begin{enumerate}
    \item \textbf{26-Setting $1.00\times$ Benchmark \& Edited Parameter Ratios (\cref{tab:app_all26,tab:app_edited_params}):} \Cref{tab:app_all26} reports complete blind metrics across all \textbf{26 Strict $1.00\times$ configurations} (17 single-model + 9 cross-model), confirming $26/26$ ($100\%$) positive blind AP gains on MS~COCO \texttt{val2017} (mean $+0.0618$~AP, $+0.127$~$\text{AP}_s$) and $20/26$ positive zero-shot gains on COCO-O ($+0.050$~mAP mean; on O365-finetuned checkpoints, $\mathcal{D}_{\text{hold}}$ sets $\alpha \to 0.96\text{--}1.0$ so $1.00\times$ Cross matches Single while preventing fixed-$\alpha$ WiSE-FT collapse in \cref{tab:ab5_basins}). \Cref{tab:app_edited_params} breaks down exact edited tensor/parameter counts across all 17 models ($3.38\%$ mean, $0.21\%$ median edited parameters; $\Delta\text{Params} = 0$ after algebraic re-collapse $\mathcal{C}$).
    \item \textbf{16-Operator Sub-Metrics \& 6-Domain Zero-Shot COCO-O (\cref{fig:S2}a,c, \cref{tab:app_cocoo_domains}):} \Cref{fig:S2}(a) confirms disproportionate $\Delta\text{AP}_{75}$ and $\Delta\text{AP}_s$ improvements across \texttt{IR} operator families, while \cref{tab:app_cocoo_domains} and \cref{fig:S2}(c) detail per-domain zero-shot COCO-O AP (\emph{Cartoon, Handmake, Painting, Sketch, Tattoo, Weather}) and ID-to-OOD effective robustness scaling~\cite{taori2020measuring}.
    \item \textbf{Zero-Image Repair \& Label-Free Weather Adaptation (\cref{fig:S2}b):} Injecting blind rank-$1$ spectral spikes into internal weights degrades 13-model COCO AP to $47.77$ and 17-model COCO-O mAP to $30.87$; zero-pass \texttt{IR-5 SVD} ($\sigma_1/\sigma_2$) localization and Empirical-Bayes shrinkage with \emph{zero images} recover $52.62$~AP on COCO (\textbf{$+4.85$~AP}, $99.3\%$ recovery) and $33.82$~mAP on COCO-O (\textbf{$+2.95$~mAP}). Under composite Night+Fog+Noise shift using unlabeled target images via \texttt{IR-AdaNorm} (\cref{eq:ir_adanorm}) + \texttt{IR-DCT} Nyquist damping at $1.00\times$ cost (\cref{fig:S2}b), detectors gain \textbf{$+2.50$~AP} on YOLO26-N ($26.32 \to 28.81$) and \textbf{$+2.79$~AP} on YOLO26-S ($32.43 \to 35.22$; $+0.81$~AP 13-model mean).
\end{enumerate}

\begin{table*}[!t]
\caption{\textbf{Complete blind evaluation across all 26 Strict $1.00\times$ configurations (17 Single-Model + 9 Cross-Model)} spanning YOLO26~\cite{jocher2026yolo26}, EdgeCrafter~\cite{liu2026edgecrafter}, and RF-DETR~\cite{robinson2026rfdetr} at $\Delta\text{Params}=0.00\%, \Delta\text{GFLOPs}=0.00\%$ on MS~COCO~\cite{lin2014coco} \texttt{val2017} and Zero-Shot COCO-O~\cite{mao2023cocoo} ($6,782$ images).}
\label{tab:app_all26}
\centering
\resizebox{\textwidth}{!}{\begin{tabular}{@{}ll|cccccc|ccccccc|ccc@{}}
\toprule
\textbf{Scope} & \textbf{Model} & \multicolumn{6}{c|}{\textbf{Official Stock MS~COCO \texttt{val2017}}} & \multicolumn{7}{c|}{\textbf{\method ($1.00\times$, $\Delta\text{Params}=0\%$) MS~COCO \texttt{val2017}}} & \multicolumn{3}{c}{\textbf{Zero-Shot COCO-O}} \\
& & AP & $\text{AP}_{50}$ & $\text{AP}_{75}$ & $\text{AP}_s$ & $\text{AP}_m$ & $\text{AP}_l$ & AP & $\Delta$AP & $\text{AP}_{50}$ & $\text{AP}_{75}$ & $\text{AP}_s$ & $\text{AP}_m$ & $\text{AP}_l$ & Off. & \method & $\Delta$OOD \\
\midrule
Single & \texttt{yolo26n} & 40.065 & 55.687 & 43.497 & 19.737 & 44.055 & 58.096 & 40.151 & \pos{+.086} & 55.693 & 43.566 & 20.028 & 44.095 & 57.997 & 18.392 & 18.415 & \pos{+.023} \\
Single & \texttt{yolo26s} & 47.786 & 64.556 & 52.152 & 29.155 & 52.505 & 64.331 & 47.904 & \pos{+.118} & 64.608 & 52.251 & 29.465 & 52.670 & 64.343 & 24.412 & 24.504 & \pos{+.092} \\
Single & \texttt{yolo26m} & 52.499 & 69.793 & 57.295 & 36.249 & 56.918 & 68.477 & 52.594 & \pos{+.095} & 69.834 & 57.349 & 36.471 & 57.042 & 68.587 & 27.495 & 27.539 & \pos{+.044} \\
Single & \texttt{yolo26l} & 54.351 & 71.501 & 59.413 & 37.881 & 58.591 & 70.339 & 54.509 & \pos{+.158} & 71.587 & 59.575 & 38.148 & 58.777 & 70.380 & 29.539 & 29.622 & \pos{+.083} \\
Single & \texttt{yolo26x} & 56.884 & 74.146 & 62.151 & 41.436 & 61.186 & 72.742 & 56.919 & \pos{+.035} & 74.086 & 62.152 & 41.539 & 61.297 & 72.733 & 32.657 & 32.749 & \pos{+.092} \\
Single & \texttt{ecdet\_s} & 51.707 & 69.365 & 55.779 & 32.262 & 56.369 & 70.494 & 51.731 & \pos{+.024} & 69.450 & 55.805 & 32.295 & 56.278 & 70.509 & 30.080 & 30.075 & \negv{$-.005$} \\
Single & \texttt{ecdet\_m} & 54.216 & 72.233 & 58.691 & 35.861 & 59.138 & 72.699 & 54.224 & \pos{+.008} & 72.251 & 58.708 & 35.889 & 59.163 & 72.681 & 35.322 & 35.354 & \pos{+.032} \\
Single & \texttt{ecdet\_l} & 56.985 & 75.081 & 61.724 & 38.704 & 62.527 & 75.019 & 56.992 & \pos{+.007} & 75.111 & 61.731 & 38.683 & 62.510 & 75.005 & 43.779 & 43.764 & \negv{$-.015$} \\
Single & \texttt{ecdet\_x} & 57.859 & 76.016 & 62.864 & 38.574 & 63.340 & 76.068 & 57.886 & \pos{+.027} & 76.022 & 62.889 & 38.625 & 63.355 & 76.135 & 45.210 & 45.370 & \pos{+.160} \\
Single & \texttt{ecdet\_o3652coco\_s} & 53.533 & 70.943 & 58.017 & 35.108 & 58.277 & 71.772 & 53.583 & \pos{+.050} & 71.027 & 58.069 & 35.151 & 58.283 & 71.808 & 31.106 & 31.144 & \pos{+.038} \\
Single & \texttt{ecdet\_o3652coco\_m} & 56.745 & 74.255 & 61.785 & 38.934 & 61.587 & 73.967 & 56.760 & \pos{+.015} & 74.296 & 61.824 & 38.918 & 61.594 & 73.981 & 37.547 & 37.550 & \pos{+.003} \\
Single & \texttt{ecdet\_o3652coco\_l} & 59.036 & 77.039 & 64.418 & 42.075 & 64.229 & 75.946 & 59.047 & \pos{+.011} & 77.030 & 64.431 & 42.088 & 64.223 & 75.971 & 45.178 & 45.209 & \pos{+.031} \\
Single & \texttt{ecdet\_o3652coco\_x} & 59.873 & 77.417 & 65.238 & 42.348 & 64.804 & 76.995 & 59.895 & \pos{+.022} & 77.443 & 65.245 & 42.433 & 64.844 & 77.004 & 47.317 & 47.338 & \pos{+.021} \\
Single & \texttt{rfdetr\_n} & 48.407 & 67.529 & 51.748 & 25.315 & 53.648 & 71.295 & 48.491 & \pos{+.084} & 67.469 & 51.867 & 25.434 & 53.839 & 71.490 & 30.115 & 30.096 & \negv{$-.019$} \\
Single & \texttt{rfdetr\_s} & 52.942 & 72.080 & 57.093 & 31.736 & 58.367 & 73.038 & 52.998 & \pos{+.056} & 72.133 & 57.151 & 31.882 & 58.456 & 73.117 & 31.892 & 32.056 & \pos{+.164} \\
Single & \texttt{rfdetr\_m} & 54.749 & 73.561 & 59.213 & 35.954 & 59.860 & 73.713 & 54.906 & \pos{+.157} & 73.642 & 59.383 & 36.191 & 60.067 & 73.838 & 32.440 & 32.704 & \pos{+.264} \\
Single & \texttt{rfdetr\_l} & 56.532 & 75.120 & 61.285 & 38.949 & 61.103 & 73.961 & 56.609 & \pos{+.077} & 75.112 & 61.417 & 39.129 & 61.288 & 74.183 & 32.567 & 32.442 & \negv{$-.125$} \\
\midrule
Cross & \texttt{yolo26n} & 40.065 & 55.687 & 43.497 & 19.737 & 44.055 & 58.096 & 40.151 & \pos{+.086} & 55.693 & 43.566 & 20.028 & 44.095 & 57.997 & 18.392 & 18.415 & \pos{+.023} \\
Cross & \texttt{yolo26s} & 47.786 & 64.556 & 52.152 & 29.155 & 52.505 & 64.331 & 47.904 & \pos{+.118} & 64.608 & 52.251 & 29.465 & 52.670 & 64.343 & 24.412 & 24.504 & \pos{+.092} \\
Cross & \texttt{yolo26m} & 52.499 & 69.793 & 57.295 & 36.249 & 56.918 & 68.477 & 52.586 & \pos{+.087} & 69.829 & 57.339 & 36.465 & 57.047 & 68.578 & 27.495 & 27.521 & \pos{+.026} \\
Cross & \texttt{yolo26l} & 54.351 & 71.501 & 59.413 & 37.881 & 58.591 & 70.339 & 54.511 & \pos{+.160} & 71.587 & 59.571 & 38.150 & 58.780 & 70.382 & 29.539 & 29.636 & \pos{+.097} \\
Cross & \texttt{yolo26x} & 56.884 & 74.146 & 62.151 & 41.436 & 61.186 & 72.742 & 56.919 & \pos{+.035} & 74.086 & 62.152 & 41.539 & 61.297 & 72.733 & 32.657 & 32.749 & \pos{+.092} \\
Cross & \texttt{ecdet\_o3652coco\_s} & 53.533 & 70.943 & 58.017 & 35.108 & 58.277 & 71.772 & 53.569 & \pos{+.036} & 70.980 & 58.058 & 35.126 & 58.287 & 71.795 & 31.106 & 31.103 & \negv{$-.003$} \\
Cross & \texttt{ecdet\_o3652coco\_m} & 56.745 & 74.255 & 61.785 & 38.934 & 61.587 & 73.967 & 56.759 & \pos{+.014} & 74.267 & 61.809 & 38.813 & 61.592 & 73.982 & 37.547 & 37.545 & \negv{$-.002$} \\
Cross & \texttt{ecdet\_o3652coco\_l} & 59.036 & 77.039 & 64.418 & 42.075 & 64.229 & 75.946 & 59.050 & \pos{+.014} & 77.075 & 64.427 & 42.125 & 64.218 & 75.953 & 45.178 & 45.224 & \pos{+.046} \\
Cross & \texttt{ecdet\_o3652coco\_x} & 59.873 & 77.417 & 65.238 & 42.348 & 64.804 & 76.995 & 59.900 & \pos{+.027} & 77.455 & 65.246 & 42.435 & 64.829 & 77.018 & 47.317 & 47.350 & \pos{+.033} \\
\midrule
\rowcolor{rowgray}
\multicolumn{2}{l|}{\textbf{26-Setting Mean}} & 53.652 & 71.218 & 58.320 & 35.508 & 58.410 & 71.216 & 53.713 & \pos{+.062} & 71.245 & 58.378 & \pos{35.635} & 58.484 & 71.252 & 33.411 & 33.461 & \pos{+.050} \\
\bottomrule
\end{tabular}}
\end{table*}

\begin{table*}[!t]
\caption{\textbf{Exact edited parameter counts, ratios, and internal-trunk vs.\ detection-head breakdown for each of the 17 detectors in \cref{tab:main_17}} (\method at $1.00\times$ / Pure $0\%$; exact element-wise diff $|W^\star - W_0| > 10^{-7}$ over all learnable \texttt{nn.Parameter} tensors). Because all \texttt{IR} subspaces collapse algebraically back into native shapes ($\mathcal{C}$), net added parameters equal $\Delta\text{Params} = 0$ ($0.00\%$) on every model.}
\label{tab:app_edited_params}
\centering
\setlength{\tabcolsep}{4.2pt}
\resizebox{\textwidth}{!}{\begin{tabular}{@{}l|rr|rr|rr|c|l@{}}
\toprule
\textbf{Detector} & \textbf{Total Params} & \textbf{Edited Tensors} & \textbf{Internal Trunk Edited} & \textbf{Head Edited} & \textbf{Total Edited Params} & \textbf{Edited Ratio} & \textbf{Net $\Delta$Params} & \textbf{Holdout-Selected Scope \& \texttt{IR} Operator} \\
\midrule
\texttt{yolo26n} & 2,572,280 & 12/366 (3.3\%) & 0 (0.00\%) & 19,644 (0.76\%) & \textbf{19,644} & \pos{0.76\%} & $0$ ($0.00\%$) & Detect Head (\texttt{IR-PCA/DCT}) \\
\texttt{yolo26s} & 10,009,784 & 71/366 (19.4\%) & 4,383,072 (43.79\%) & 625,788 (6.25\%) & \textbf{5,008,860} & \pos{50.04\%} & $0$ ($0.00\%$) & Full FPN/PAN Neck + Head (\texttt{IR-12}) \\
\texttt{yolo26m} & 21,896,248 & 12/396 (3.0\%) & 0 (0.00\%) & 62,460 (0.29\%) & \textbf{62,460} & \pos{0.29\%} & $0$ ($0.00\%$) & Detect Head (\texttt{IR-PCA/DCT}) \\
\texttt{yolo26l} & 26,299,704 & 12/558 (2.2\%) & 0 (0.00\%) & 62,460 (0.24\%) & \textbf{62,460} & \pos{0.24\%} & $0$ ($0.00\%$) & Detect Head (\texttt{IR-PCA/DCT}) \\
\texttt{yolo26x} & 58,993,368 & 12/558 (2.2\%) & 0 (0.00\%) & 93,564 (0.16\%) & \textbf{93,564} & \pos{0.16\%} & $0$ ($0.00\%$) & Detect Head (\texttt{IR-PCA/DCT}) \\
\rowcolor{rowgray}
\textit{YOLO26 Mean} & 23,954,277 & 6.0\% & 876,614 (8.76\%) & 172,783 (1.54\%) & \textbf{1,049,398} & \pos{10.30\%} & $0$ ($0.00\%$) & --- \\
\midrule
\texttt{ecdet\_s} & 9,884,102 & 2/484 (0.4\%) & 0 (0.00\%) & 15,440 (0.16\%) & \textbf{15,440} & \pos{0.16\%} & $0$ ($0.00\%$) & Decoder Score Head (\texttt{IR-AdaNorm}) \\
\texttt{ecdet\_m} & 19,431,502 & 117/484 (24.2\%) & 19,392 (0.10\%) & 20,480 (0.11\%) & \textbf{39,872} & \pos{0.21\%} & $0$ ($0.00\%$) & HybridEncoder Norms + Head (\texttt{IR-AdaBN}) \\
\texttt{ecdet\_l} & 32,966,110 & 4/538 (0.7\%) & 0 (0.00\%) & 54,484 (0.17\%) & \textbf{54,484} & \pos{0.17\%} & $0$ ($0.00\%$) & Decoder Score + BBox Heads (\texttt{IR-PCA}) \\
\texttt{ecdet\_x} & 49,941,982 & 5/538 (0.9\%) & 1,203,984 (2.41\%) & 0 (0.00\%) & \textbf{1,203,984} & \pos{2.41\%} & $0$ ($0.00\%$) & ViT Patch-Embed / Stem Convs (\texttt{IR-12}) \\
\rowcolor{rowgray}
\textit{ECDet-COCO Mean} & 28,055,924 & 6.6\% & 305,844 (0.63\%) & 22,601 (0.11\%) & \textbf{328,445} & \pos{0.73\%} & $0$ ($0.00\%$) & --- \\
\midrule
\texttt{ecdet\_o3652coco\_s} & 9,884,102 & 194/484 (40.1\%) & 24,912 (0.25\%) & 15,824 (0.16\%) & \textbf{40,736} & \pos{0.41\%} & $0$ ($0.00\%$) & Backbone/Enc/Dec Norms + Head (\texttt{IR-AdaNorm}) \\
\texttt{ecdet\_o3652coco\_m} & 19,431,502 & 193/484 (39.9\%) & 38,336 (0.20\%) & 20,992 (0.11\%) & \textbf{59,328} & \pos{0.31\%} & $0$ ($0.00\%$) & Backbone/Enc/Dec Norms + Head (\texttt{IR-AdaNorm}) \\
\texttt{ecdet\_o3652coco\_l} & 32,966,110 & 229/538 (42.6\%) & 48,800 (0.15\%) & 20,992 (0.06\%) & \textbf{69,792} & \pos{0.21\%} & $0$ ($0.00\%$) & Backbone/Enc/Dec Norms + Head (\texttt{IR-AdaNorm}) \\
\texttt{ecdet\_o3652coco\_x} & 49,941,982 & 229/538 (42.6\%) & 61,088 (0.12\%) & 20,992 (0.04\%) & \textbf{82,080} & \pos{0.16\%} & $0$ ($0.00\%$) & Backbone/Enc/Dec Norms + Head (\texttt{IR-AdaNorm}) \\
\rowcolor{rowgray}
\textit{ECDet-O365$\to$C Mean} & 28,055,924 & 41.3\% & 43,284 (0.18\%) & 19,700 (0.09\%) & \textbf{62,984} & \pos{0.27\%} & $0$ ($0.00\%$) & --- \\
\midrule
\texttt{rfdetr\_n} & 30,467,298 & 2/465 (0.4\%) & 0 (0.00\%) & 23,387 (0.08\%) & \textbf{23,387} & \pos{0.08\%} & $0$ ($0.00\%$) & Shared Decoder \texttt{class\_embed} (\texttt{IR-DCT}) \\
\texttt{rfdetr\_s} & 32,111,170 & 2/487 (0.4\%) & 0 (0.00\%) & 23,387 (0.07\%) & \textbf{23,387} & \pos{0.07\%} & $0$ ($0.00\%$) & Shared Decoder \texttt{class\_embed} (\texttt{IR-PCA}) \\
\texttt{rfdetr\_m} & 33,687,458 & 2/509 (0.4\%) & 0 (0.00\%) & 23,387 (0.07\%) & \textbf{23,387} & \pos{0.07\%} & $0$ ($0.00\%$) & Shared Decoder \texttt{class\_embed} (\texttt{IR-RepConv}) \\
\texttt{rfdetr\_l} & 33,933,218 & 6/509 (1.2\%) & 589,824 (1.74\%) & 23,387 (0.07\%) & \textbf{613,211} & \pos{1.81\%} & $0$ ($0.00\%$) & NAS Multi-Scale Projector + Head (\texttt{IR-LowRank}) \\
\rowcolor{rowgray}
\textit{RF-DETR Mean} & 32,549,786 & 0.6\% & 147,456 (0.43\%) & 23,387 (0.07\%) & \textbf{170,843} & \pos{0.51\%} & $0$ ($0.00\%$) & --- \\
\midrule
\rowcolor{rowgray}
\textbf{17-Model Mean} & 27,906,936 & 13.2\% & 374,671 (2.87\%) & 66,275 (0.52\%) & \textbf{440,946} & \pos{3.38\%} & $0$ ($0.00\%$) & Mean across 17 checkpoints \\
\textbf{17-Model Median} & 30,467,298 & 2.2\% & --- & --- & \textbf{59,328} & \pos{0.21\%} & $0$ ($0.00\%$) & Median across 17 checkpoints \\
\bottomrule
\end{tabular}}
\end{table*}
\clearpage

\begin{strip}
\noindent\begin{minipage}[t]{\columnwidth}
\subsection{Controlled $1.00\times$ Mechanistic Ablations: Collision Gate (\texttt{AB2-S6}) \& Cross-Basin Symmetry (\texttt{AB4})}
\label{sec:app_ab5}

Grouping our $1.00\times$ mechanistic ablations side-by-side, \cref{tab:app_ab4_gate}, \cref{fig:app_ab5_basin}, and \cref{tab:ab5_basins} isolate dual-head collision gating on YOLO26 and cross-basin symmetry alignment across O365/COCO checkpoints:
\begin{enumerate}
    \item \textbf{Per-Scale Dual-Head Collision-Gate Breakdown (\texttt{AB2-S6}, \cref{tab:app_ab4_gate}):} Isolating each term of \cref{eq:collision_gate} on stock YOLO26 (N--X) verifies consistent gains over \texttt{G0} across every scale: ungated \texttt{o2o+o2m} mixing (\texttt{G1}) leaks duplicate false positives on crowded instances, whereas downward suppression (\texttt{G2}), spatial IoU overlap gating (\texttt{G3}), and local peak protection (\texttt{G4}) progressively eliminate collisions before \texttt{G5} (+collapsed \texttt{IR} weights) achieves the highest ID AP across all five scales while improving OOD mAP over \texttt{G0} on every scale at $1.00\times$ cost.
    \item \textbf{Cross-Basin Symmetry Cosine Law \& Fine-Tuning Drift (\texttt{AB4}, \cref{fig:app_ab5_basin}, \cref{tab:ab5_basins}):} Evaluating cross-model fusion (\cref{sec:method_cross}, $\alpha=0.85$, $1.00\times$ cost) uncovers a phase transition governed by trunk cosine $\cos_{\text{raw}}$ between $W_{\text{O365-pre}}$ and $W_0$: (i)~in the \emph{linearly connected regime} ($\cos > 0.95$, YOLO26-X and ECDet-O365$\to$COCO-S), linear interpolation preserves coherence; (ii)~in \emph{disjoint or fine-tuning-drifted basins} ($\cos < 0.70$), multi-epoch fine-tuning rotates ECDet (O365$\to$COCO)-X away from its ancestor ($\cos=0.35$ vs.\ $0.98$ in Small), causing naive WiSE-FT~\cite{wortsman2022wiseft} to plunge by $-52.4$~AP. Zero-gradient diagonal (\texttt{Diag$_+$}) and Orthogonal Procrustes (\texttt{Proc.\ $O(d)$}, \cref{eq:procrustes_align}) alignment turn $-40.2$~AP collapse on disjoint ECDet-COCO-X into $+1.42$~mAP ($46.6$) and $+0.94$~mAP ($46.2$) zero-shot OOD gains at fixed $\alpha=0.85$, while holdout-selected \method preserves ID AP with $+0.68$~mAP OOD gain ($45.9$).
\end{enumerate}
\end{minipage}\hfill
\begin{minipage}[t]{\columnwidth}
\centering
\vspace{0pt}
\captionof{table}{\textbf{\texttt{AB2} (\texttt{S6} Breakdown): Per-scale YOLO26~\cite{jocher2026yolo26} dual-head collision-gate ablation} (MS~COCO~\cite{lin2014coco} \texttt{val2017} AP / COCO-O~\cite{mao2023cocoo} mAP; \texttt{G1}--\texttt{G4} isolate \cref{eq:collision_gate} terms on stock \texttt{o2o} prior to combining with collapsed \texttt{IR} weights in \texttt{G5}).}
\label{tab:app_ab4_gate}
\vspace{0.3em}
\resizebox{\columnwidth}{!}{\begin{tabular}{@{}l|ccccc@{}}
\toprule
\textbf{Gate Variant} & \textbf{Y26-N} & \textbf{Y26-S} & \textbf{Y26-M} & \textbf{Y26-L} & \textbf{Y26-X} \\
\midrule
\texttt{G0}: Stock \texttt{o2o} & 40.07/18.39 & 47.79/24.41 & 52.50/27.50 & 54.35/29.54 & 56.88/32.66 \\
\texttt{G1}: Ungated & 40.10/18.40 & 47.82/24.49 & 52.55/27.48 & 54.45/29.61 & 56.90/32.73 \\
\texttt{G2}: Down-Only & 40.11/18.41 & 47.81/24.49 & 52.55/27.49 & 54.47/29.58 & 56.89/32.69 \\
\texttt{G3}: +IoU Gate & 40.10/18.42 & 47.84/24.52 & 52.58/27.49 & 54.49/29.63 & 56.90/32.76 \\
\texttt{G4}: +Peak-Dom. & 40.11/18.42 & 47.85/24.52 & 52.58/27.50 & 54.50/29.64 & 56.90/32.77 \\
\texttt{G5}: +\texttt{IR} Weights & \pos{40.15/18.42} & \pos{47.90/24.50} & \pos{52.59/27.54} & \pos{54.51/29.62} & \pos{56.92/32.75} \\
\bottomrule
\end{tabular}}
\par\vspace{1.0em}
\includegraphics[width=0.92\columnwidth]{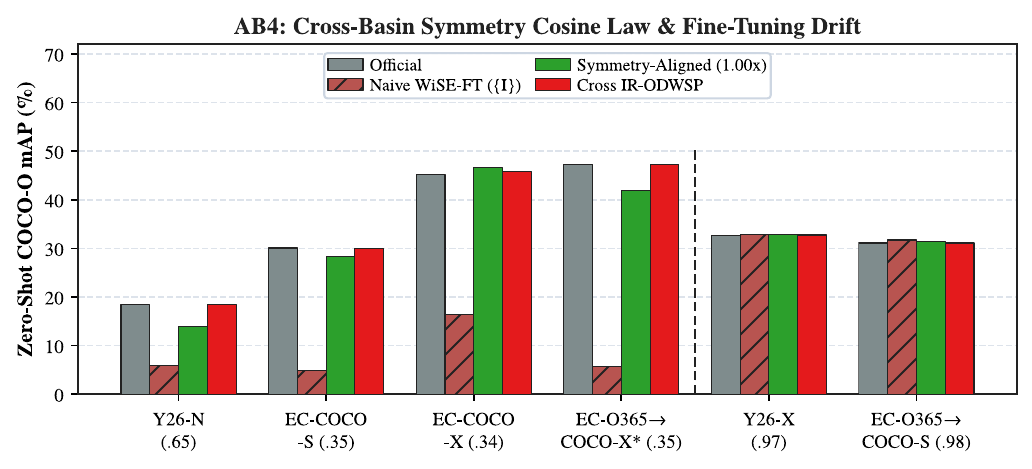}
\vspace{-0.3em}
\captionof{figure}{\textbf{\texttt{AB4} Symmetry-Basin Cosine Law.} The vertical dashed line separates disjoint or fine-tuning-drifted basins ($\cos < 0.70$, left four pairs, including ECDet~\cite{liu2026edgecrafter} (O365~\cite{shao2019objects365}$\to$COCO~\cite{lin2014coco})-X with $\cos=0.35^*$), where naive WiSE-FT~\cite{wortsman2022wiseft} collapses and diagonal/Procrustes~\cite{schonemann1966generalized} symmetry alignment recovers gains, from linearly connected basins ($\cos > 0.95$, right two pairs).}
\label{fig:app_ab5_basin}
\par\vspace{1.0em}
\captionof{table}{\textbf{\texttt{AB4}: Single- vs.\ cross-model symmetry-basin alignment ($\alpha=0.85$, $1.00\times$ cost)} on MS~COCO~\cite{lin2014coco} \texttt{val2017} / COCO-O~\cite{mao2023cocoo} mAP comparing Naive WiSE-FT~\cite{wortsman2022wiseft} against Diagonal and Procrustes~\cite{schonemann1966generalized} alignment. $^*$EC-O365$\to$COCO-X~\cite{liu2026edgecrafter} exhibits large-scale fine-tuning basin drift ($\cos=0.35$) from its Objects365~\cite{shao2019objects365} pretraining prior $W_{\text{O365-pre}}$.}
\label{tab:ab5_basins}
\vspace{0.3em}
\resizebox{\columnwidth}{!}{\begin{tabular}{@{}lc|ccccc@{}}
\toprule
\textbf{Pair ($W_{\text{O365-pre}}, W_0$)} & $\cos_{\text{raw}}{\to}\cos_{Q}$ & \textbf{Off.} & \textbf{Naive $\{I\}$} & \textbf{Diag$_+$} & \textbf{Proc.\ $O(d)$} & \textbf{\method} \\
\midrule
Y26-N (Disj.) & $.65{\to}.90$ & 40.1/18.4 & \negv{18.3/5.8} & 28.8/11.1 & 33.8/13.9 & \pos{40.2/18.4} \\
EC-COCO-S (Disj.) & $.35{\to}.75$ & 51.7/30.1 & \negv{7.3/4.8} & 47.8/28.3 & 48.5/28.1 & 51.2/30.0 \\
EC-COCO-X (Disj.) & $.34{\to}.61$ & 57.9/45.2 & \negv{17.7/16.5} & 57.1/\pos{46.6} & 57.1/\pos{46.2} & 57.8/\pos{45.9} \\
EC-O365$\to$COCO-X$^*$ & $.35{\to}.62$ & 59.9/47.3 & \negv{7.5/5.6} & 55.1/41.9 & 48.4/40.5 & \pos{59.9/47.4} \\
\midrule
Y26-X (Conn.) & $.97{\to}.98$ & 56.9/32.7 & 56.5/\pos{32.8} & 56.5/\pos{32.8} & 56.6/32.7 & \pos{56.9/32.7} \\
EC-O365$\to$COCO-S & $.98{\to}.99$ & 53.5/31.1 & 53.3/\pos{31.7} & 53.4/\pos{31.3} & 52.6/30.5 & \pos{53.6}/31.1 \\
\bottomrule
\end{tabular}}
\end{minipage}
\end{strip}

\begin{figure*}[!t]
    \centering
    \includegraphics[width=0.79\linewidth]{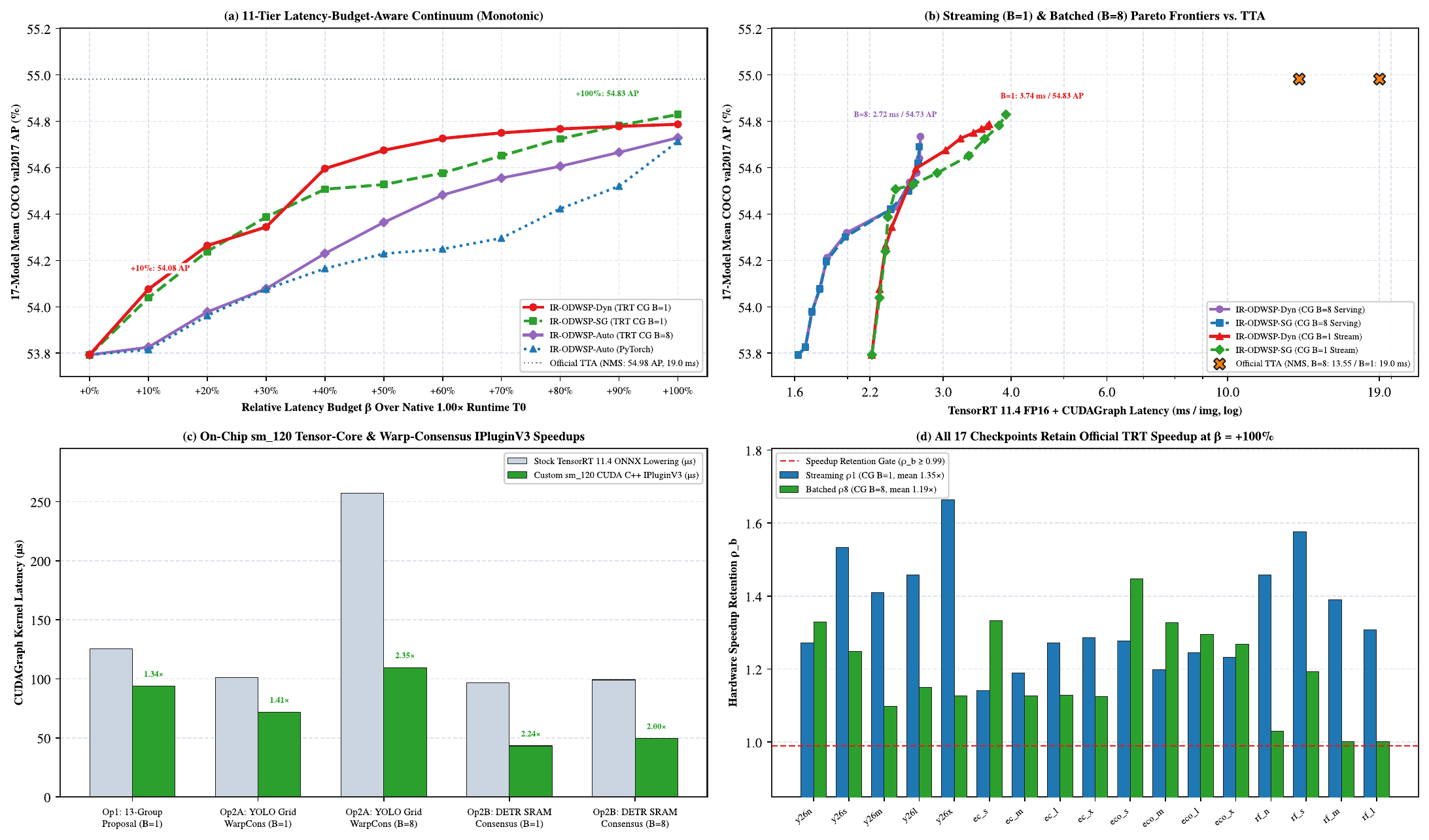}
    \vspace{-0.6em}
    \caption{\textbf{Latency-budget-aware continuums, batched serving frontiers, and hardware speedup retention across all 17 pretrained detectors (supplementing \cref{tab:tta_vs_mem}).} \textbf{(a)} Complete 11-tier continuum ($\beta \in [0\%, +100\%]$) for \methodDyn (\cref{alg:dyn_graph}), \methodSG (\cref{alg:static_graph}), and \methodAuto (strictly monotonic transitions across all budgets). \textbf{(b)} Streaming ($B{=}1$) and batched serving ($B{=}8$) \texttt{CUDAGraph} Pareto frontiers vs.\ Official TTA (NMS). \textbf{(c)} On-chip \texttt{sm\_120} \texttt{IPluginV3} kernel microbenchmarks ($1.34\times$--$2.35\times$). \textbf{(d)} Individual checkpoint speedup retention ($\rho_1, \rho_8 \ge 0.99$) across all 17 models at $\beta = +100\%$.}
    \label{fig:app_continuum}
    \vspace{-0.4em}
\end{figure*}

\begin{table*}[!t]
\caption{\textbf{Complete 11-tier latency-budget-aware continuums ($\beta \in \{0\%, +10\%, \dots, +100\%\}$) and unconstrained upper bound (\methodplus) across all 17 detectors}, supplementing \cref{tab:tta_vs_mem} with $B{=}1$ PyTorch latencies (PT1, ms/img), batched \texttt{CUDAGraph} $B{=}8$ latencies (\methodAuto reports both $B{=}1$ and $B{=}8$-budgeted envelopes), and speedup-retention ratios $\overline{\rho}_b(\min_{17})$ ($0.022\text{--}0.069$~AP FP16 parity, $0$ retraining, $0$ NMS/WBF; $\beta{=}0\%$ single-head floor $d_0$ reports pure collapsed weights [$53.79/33.87$] without YOLO26's \texttt{o2m} gate [$53.84/33.88$ in \cref{tab:main_17,tab:tta_vs_mem}]).}
\label{tab:app_continuum}
\centering
\resizebox{\textwidth}{!}{\begin{tabular}{@{}l|cccc|cccc|cccccc@{}}
\toprule
& \multicolumn{4}{c|}{\textbf{\methodDyn (\cref{alg:dyn_graph})}} & \multicolumn{4}{c|}{\textbf{\methodSG (\cref{alg:static_graph})}} & \multicolumn{6}{c}{\textbf{\methodAuto (Holdout-Selected Envelope)}} \\
\textbf{Budget $\beta$} & \textbf{PT1} & \textbf{CG1} & \textbf{CG8} & \textbf{ID / OOD} & \textbf{PT1} & \textbf{CG1} & \textbf{CG8} & \textbf{ID / OOD} & \textbf{PT1} & \textbf{CG1 / CG8} & \textbf{$\overline{\rho}_1$($\min$)} & \textbf{$\overline{\rho}_8$($\min$)} & \textbf{AP$_s$} & \textbf{ID / OOD} \\
\midrule
\textbf{+0\%} & 8.07 & 2.22 & 1.62 & 53.79/33.87 & 8.07 & 2.22 & 1.62 & 53.79/33.87 & 8.07 & \textbf{2.22 / 1.62} & 1.00(1.00) & 1.00(1.00) & 35.40 & 53.79/33.87 \\
\textbf{+10\%} & 9.26 & 2.29 & 1.69 & \pos{54.08}/\pos{33.94} & 8.97 & 2.29 & 1.69 & \pos{54.04}/\pos{33.94} & 9.26 & \textbf{2.29 / 1.67} & \pos{1.12}(1.00) & \pos{1.02}(1.00) & \pos{35.76} & \pos{54.08}/\pos{33.94} \\
\textbf{+20\%} & 9.96 & 2.35 & 1.72 & \pos{54.26}/\pos{34.01} & 9.52 & 2.35 & 1.72 & \pos{54.24}/\pos{33.98} & 10.13 & \textbf{2.36 / 1.72} & \pos{1.17}(1.01) & \pos{1.04}(1.00) & \pos{36.12} & \pos{54.31}/\pos{34.02} \\
\textbf{+30\%} & 10.90 & 2.41 & 1.79 & \pos{54.34}/\pos{34.08} & 10.27 & 2.37 & 1.75 & \pos{54.39}/\pos{33.98} & 11.80 & \textbf{2.43 / 1.78} & \pos{1.28}(1.01) & \pos{1.05}(1.00) & \pos{36.50} & \pos{54.47}/\pos{34.07} \\
\textbf{+40\%} & 12.98 & 2.67 & 2.01 & \pos{54.60}/\pos{34.10} & 9.86 & 2.45 & 1.79 & \pos{54.51}/33.87 & 12.99 & \textbf{2.67 / 1.84} & \pos{1.31}(1.01) & \pos{1.14}(1.00) & \pos{36.75} & \pos{54.63}/\pos{34.07} \\
\textbf{+50\%} & 15.20 & 3.03 & 2.41 & \pos{54.68}/\pos{34.27} & 11.02 & 2.64 & 1.91 & \pos{54.53}/\pos{33.95} & 14.89 & \textbf{3.05 / 2.00} & \pos{1.36}(1.03) & \pos{1.15}(1.00) & \pos{36.93} & \pos{54.69}/\pos{34.21} \\
\textbf{+60\%} & 17.04 & 3.23 & 2.54 & \pos{54.73}/\pos{34.28} & 12.62 & 2.92 & 2.09 & \pos{54.58}/\pos{34.03} & 16.31 & \textbf{3.25 / 2.46} & \pos{1.39}(1.03) & \pos{1.19}(1.00) & \pos{36.97} & \pos{54.73}/\pos{34.25} \\
\textbf{+70\%} & 16.63 & 3.41 & 2.86 & \pos{54.75}/\pos{34.25} & 13.87 & 3.34 & 2.47 & \pos{54.65}/\pos{34.14} & 15.59 & \textbf{3.45 / 2.60} & \pos{1.41}(1.18) & \pos{1.20}(1.04) & \pos{37.01} & \pos{54.77}/\pos{34.20} \\
\textbf{+80\%} & 15.50 & 3.53 & 3.07 & \pos{54.77}/\pos{34.24} & 15.48 & 3.57 & 2.66 & \pos{54.72}/\pos{34.21} & 13.92 & \textbf{3.58 / 2.68} & \pos{1.41}(1.23) & \pos{1.21}(1.04) & \pos{37.02} & \pos{54.80}/\pos{34.20} \\
\textbf{+90\%} & 14.38 & 3.61 & 3.20 & \pos{54.78}/\pos{34.21} & 13.53 & 3.80 & 3.16 & \pos{54.78}/\pos{34.18} & 12.80 & \textbf{3.66 / 2.71} & \pos{1.39}(1.19) & \pos{1.21}(1.00) & \pos{37.07} & \pos{54.81}/\pos{34.16} \\
\rowcolor{rowgray}
\textbf{+100\%} & 13.84 & 3.65 & 3.25 & \pos{54.79}/\pos{34.21} & 12.69 & 3.91 & 3.40 & \pos{54.83}/\pos{34.18} & 12.69 & \textbf{3.74 / 2.72} & \pos{1.35}(1.14) & \pos{1.19}(1.00) & \pos{37.09} & \pos{54.83}/\pos{34.18} \\
\midrule
Official TTA (NMS) & --- & --- & --- & --- & --- & --- & --- & --- & 57.30 & 19.03 / 13.55 & --- & --- & 38.12 & 54.98/35.77 \\
\textbf{\methodplus} (Unconstrained) & 383.8 & 222.1 & 108.6 & \pos{55.91}/\pos{35.34} & 142.4 & 116.7 & 88.5 & \pos{55.79}/\pos{35.41} & 291.6 & \textbf{208.8 / 126.4} & --- & --- & \pos{39.34} & \pos{55.91}/\pos{35.44} \\
\bottomrule
\end{tabular}}
\vspace{-0.7em}
\end{table*}
\null\newpage\null\newpage

\subsection{Latency-Budget-Aware Continuums, Batched Serving \& Per-Detector Audit}
\label{sec:app_tta_audit}

\begin{figure*}[!t]
    \centering
    \includegraphics[width=0.72\linewidth]{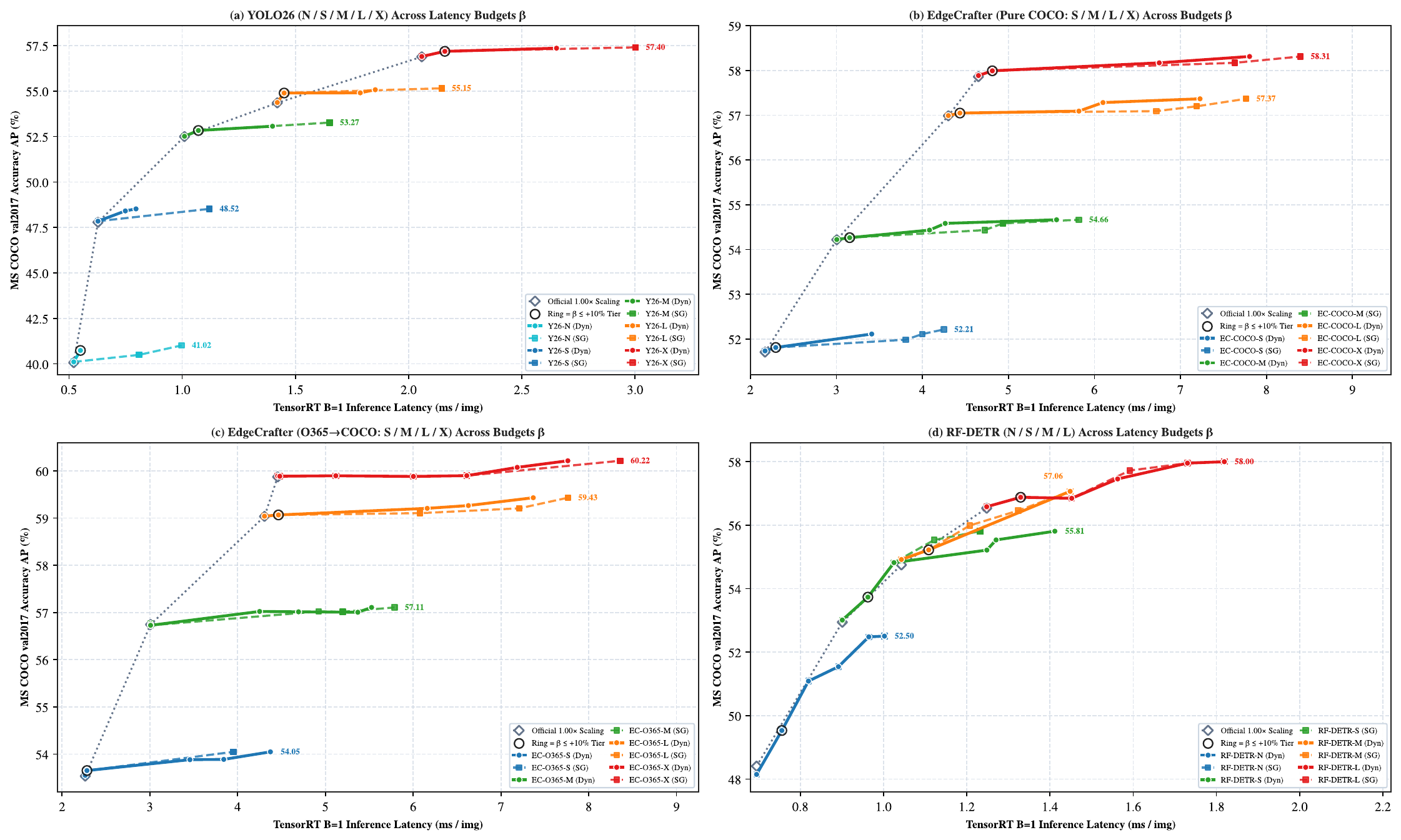}
    \vspace{-0.6em}
    \caption{\textbf{Per-scale accuracy--latency trajectories across all 3 architectures (4 checkpoint tracks, 17 scales) and 11 budgets ($\beta \in \{0\%, +10\%, +20\%, \dots, +100\%\}$), expanding \cref{tab:tta_vs_mem}.} X-axis: streaming \texttt{TensorRT~11.4 FP16} ($B{=}1$, +\texttt{CUDAGraph}) latency (ms/img); Y-axis: MS~COCO \texttt{val2017} AP (\%). \textbf{(a)} YOLO26 (\texttt{N, S, M, L, X}). \textbf{(b)} EdgeCrafter --- Pure COCO (\texttt{S, M, L, X}). \textbf{(c)} EdgeCrafter --- Objects365$\to$COCO (\texttt{S, M, L, X}). \textbf{(d)} RF-DETR (\texttt{N, S, M, L}). Gray dotted curves connect official $1.00\times$ checkpoints; solid/dashed curves trace \methodDyn / \methodSG across $10\%$-step budgets (black rings mark $\beta \le +10\%$; coordinates in \cref{tab:app_dyn_all17_steps,tab:app_sg_all17_steps}), enabling smaller scales (\texttt{ecdet\_o3652coco\_l}, \texttt{rfdetr\_s}, \texttt{rfdetr\_m}) to match or surpass official next-larger backbones.}
    \label{fig:app_arch_scale_lines}
    \vspace{-0.4em}
\end{figure*}

Directly supplementing \cref{sec:method_graph,sec:tta_vs_mem} and expanding \cref{tab:tta_vs_mem}, \cref{fig:app_continuum,fig:app_arch_scale_lines} and \cref{tab:app_continuum,tab:app_tta_all17,tab:app_dyn_all17_steps,tab:app_sg_all17_steps} report the complete 11-tier positive-budget continuums, batched $B{=}8$ serving frontiers, and per-detector $10\%$-step trajectories across all 3 architectures and 17 scales:
\begin{enumerate}
    \item \textbf{11-Tier Continuum, Batched $B{=}8$ Serving \& Speedup Retention (\cref{fig:app_continuum}a--c, \cref{tab:app_continuum}):} Starting from compiled single-head $d_0$ ($53.79$~AP without YOLO26's \texttt{o2m} gate; $53.84$~AP with gate in \cref{tab:main_17,tab:tta_vs_mem}), across PyTorch ($53.79 \to 54.71$~AP), batched \texttt{CUDAGraph} $B{=}8$ ($1.62 \to 2.72$~ms/img, $53.79 \to 54.73$~AP), and streaming \texttt{CUDAGraph} $B{=}1$ ($2.22 \to 3.74$~ms/img, $53.79 \to 54.83$~AP), all transitions are strictly non-decreasing without retraining. On-chip \texttt{sm\_120} \texttt{IPluginV3} fusion ($1.34\times$--$2.35\times$ faster kernels, \cref{fig:app_continuum}c) and 2-stream fork-join \texttt{cudaGraph\_t} capture raise mean speedup retention to $\overline{\rho}_1{=}1.35\times$ and $\overline{\rho}_8{=}1.19\times$ ($\min_{17}\rho_b \ge 0.99$; $54.71$~AP under $\min_{17}\rho_b \ge 1.002$), running $5.0\times$ faster at $B{=}8$ than Official TTA ($13.55$~ms).
\end{enumerate}

\begin{table*}[!t]
\caption{\textbf{Per-detector expansion of \cref{tab:tta_vs_mem} focusing on detailed $\beta \le +10\%$ budget metrics across all 17 checkpoints and 4 detector families} (contrasted against $1.00\times$, Official TTA [NMS], and unconstrained \methodplus). At $\beta \le +10\%$, we report streaming $B{=}1$ PyTorch latency (ms), streaming $B{=}1$ TensorRT+\texttt{CUDAGraph} latency (ms and relative overhead $\Delta$TRT\% over native $T_0$), MS~COCO \texttt{val2017} AP ($\Delta$AP vs.\ Official $1.00\times$), small-object $\text{AP}_s$, and zero-shot COCO-O mAP (when dual-head overhead exceeds $+10\%$, \methodSG falls back to compiled single-head $d_0$).}
\label{tab:app_tta_all17}
\centering
\setlength{\tabcolsep}{3.2pt}
\resizebox{\textwidth}{!}{\begin{tabular}{@{}l|cccc|ccccc|ccccc|cc@{}}
\toprule
& \multicolumn{4}{c|}{\textbf{$1.00\times$ ($\beta{=}0\%$)}} & \multicolumn{5}{c|}{\textbf{\methodDyn (\cref{alg:dyn_graph}, $\beta \le +10\%$)}} & \multicolumn{5}{c|}{\textbf{\methodSG (\cref{alg:static_graph}, $\beta \le +10\%$)}} & \multicolumn{2}{c}{\textbf{Ref.\ \& Upper Bound}} \\
\textbf{Detector} & \textbf{PT1} & \textbf{TRT1} & \textbf{Off/Ours ID} & \textbf{Off/Ours OOD} & \textbf{PT1} & \textbf{TRT1 ($\Delta\%$)} & \textbf{ID ($\Delta$AP)} & \textbf{AP$_s$} & \textbf{OOD} & \textbf{PT1} & \textbf{TRT1 ($\Delta\%$)} & \textbf{ID ($\Delta$AP)} & \textbf{AP$_s$} & \textbf{OOD} & \textbf{Off.\ TTA (TRT1/ID)} & \textbf{\methodplus (ID/OOD)} \\
\midrule
\texttt{yolo26n} & 5.39 & 0.52 & 40.07/\pos{40.15} & 18.39/\pos{18.42} & 10.41 & \textbf{0.55} ($+5.7\%$) & \pos{40.74} (+0.67) & 20.61 & 18.56 & 5.39 & \textbf{0.52} ($+0.0\%$) & \pos{40.11} (+0.05) & 19.81 & 18.45 & 4.5/42.68 & \pos{42.57/20.25} \\
\texttt{yolo26s} & 5.77 & 0.63 & 47.79/\pos{47.90} & 24.41/\pos{24.50} & 5.77 & \textbf{0.63} ($+0.0\%$) & \pos{47.85} (+0.06) & 29.43 & 24.40 & 5.77 & \textbf{0.63} ($+0.0\%$) & \pos{47.85} (+0.06) & 29.43 & 24.40 & 6.0/49.96 & \pos{49.89/26.19} \\
\texttt{yolo26m} & 6.19 & 1.01 & 52.50/\pos{52.59} & 27.50/\pos{27.54} & 6.19 & \textbf{1.07} ($+6.1\%$) & \pos{52.83} (+0.33) & 36.52 & 27.33 & 6.19 & \textbf{1.07} ($+6.1\%$) & \pos{52.83} (+0.33) & 36.52 & 27.33 & 7.6/54.02 & \pos{53.97/29.34} \\
\texttt{yolo26l} & 7.99 & 1.42 & 54.35/\pos{54.51} & 29.54/\pos{29.62} & 7.99 & \textbf{1.45} ($+2.1\%$) & \pos{54.89} (+0.54) & 37.98 & 29.63 & 7.99 & \textbf{1.45} ($+2.1\%$) & \pos{54.89} (+0.54) & 37.98 & 29.63 & 10.4/55.49 & \pos{55.45/30.77} \\
\texttt{yolo26x} & 8.14 & 2.06 & 56.88/\pos{56.92} & 32.66/\pos{32.75} & 8.14 & \textbf{2.16} ($+4.9\%$) & \pos{57.18} (+0.30) & 41.32 & 32.84 & 8.14 & \textbf{2.16} ($+4.9\%$) & \pos{57.18} (+0.30) & 41.32 & 32.84 & 14.6/57.79 & \pos{57.68/33.80} \\
\rowcolor{rowgray}
\textit{YOLO26 Mean} & 6.70 & 1.13 & 50.32/\pos{50.42} & 26.50/\pos{26.57} & 7.70 & \textbf{1.17} ($+3.8\%$) & \pos{50.70} (+0.38) & 33.17 & \pos{26.55} & 6.70 & \textbf{1.17} ($+2.6\%$) & \pos{50.57} (+0.26) & 33.01 & \pos{26.53} & 8.6/51.99 & \pos{51.91/28.07} \\
\midrule
\texttt{ecdet\_s} & 9.86 & 2.17 & 51.71/\pos{51.73} & 30.08/\negv{30.08} & 10.72 & \textbf{2.30} ($+5.7\%$) & \pos{51.81} (+0.10) & 32.82 & 30.12 & 10.72 & \textbf{2.30} ($+5.7\%$) & \pos{51.81} (+0.10) & 32.82 & 30.12 & 18.1/52.53 & \pos{53.25/31.71} \\
\texttt{ecdet\_m} & 9.80 & 3.01 & 54.22/\pos{54.22} & 35.32/\pos{35.35} & 10.80 & \textbf{3.15} ($+5.0\%$) & \pos{54.26} (+0.05) & 35.94 & 35.41 & 10.80 & \textbf{3.15} ($+5.0\%$) & \pos{54.26} (+0.05) & 35.94 & 35.41 & 24.2/54.92 & \pos{55.51/36.79} \\
\texttt{ecdet\_l} & 9.74 & 4.30 & 56.99/\pos{56.99} & 43.78/\negv{43.76} & 11.40 & \textbf{4.44} ($+3.1\%$) & \pos{57.05} (+0.06) & 38.84 & 43.82 & 11.40 & \textbf{4.44} ($+3.1\%$) & \pos{57.05} (+0.06) & 38.84 & 43.82 & 33.0/57.63 & \pos{58.37/44.62} \\
\texttt{ecdet\_x} & 9.74 & 4.65 & 57.86/\pos{57.89} & 45.21/\pos{45.37} & 11.68 & \textbf{4.81} ($+3.5\%$) & \pos{57.99} (+0.13) & 38.84 & 45.50 & 11.68 & \textbf{4.81} ($+3.5\%$) & \pos{57.99} (+0.13) & 38.84 & 45.50 & 35.6/58.39 & \pos{59.19/46.65} \\
\rowcolor{rowgray}
\textit{ECDet-COCO Mean} & 9.79 & 3.53 & 55.19/\pos{55.21} & 38.60/\pos{38.64} & 11.15 & \textbf{3.67} ($+4.3\%$) & \pos{55.28} (+0.09) & 36.61 & \pos{38.71} & 11.15 & \textbf{3.67} ($+4.3\%$) & \pos{55.28} (+0.09) & 36.61 & \pos{38.71} & 27.7/55.87 & \pos{56.58/39.94} \\
\midrule
\texttt{ecdet\_o3652coco\_s} & 9.45 & 2.26 & 53.53/\pos{53.58} & 31.11/\pos{31.14} & 10.38 & \textbf{2.28} ($+0.9\%$) & \pos{53.65} (+0.12) & 35.20 & 31.22 & 10.38 & \textbf{2.28} ($+0.9\%$) & \pos{53.65} (+0.12) & 35.20 & 31.22 & 18.0/54.19 & \pos{55.03/33.04} \\
\texttt{ecdet\_o3652coco\_m} & 10.02 & 3.01 & 56.75/\pos{56.76} & 37.55/\pos{37.55} & 10.41 & \textbf{3.01} ($+0.0\%$) & \negv{56.73} (-0.02) & 38.64 & 37.51 & 10.41 & \textbf{3.01} ($+0.0\%$) & \negv{56.73} (-0.02) & 38.64 & 37.51 & 27.9/57.29 & \pos{58.14/38.93} \\
\texttt{ecdet\_o3652coco\_l} & 9.94 & 4.31 & 59.04/\pos{59.05} & 45.18/\pos{45.21} & 11.46 & \textbf{4.47} ($+3.7\%$) & \pos{59.07} (+0.03) & 42.27 & 45.34 & 11.46 & \textbf{4.47} ($+3.7\%$) & \pos{59.07} (+0.03) & 42.27 & 45.34 & 33.2/59.70 & \pos{60.31/46.51} \\
\texttt{ecdet\_o3652coco\_x} & 10.14 & 4.46 & 59.87/\pos{59.90} & 47.32/\pos{47.34} & 11.92 & \textbf{4.48} ($+0.5\%$) & \pos{59.89} (+0.01) & 42.35 & 47.32 & 11.92 & \textbf{4.48} ($+0.5\%$) & \pos{59.89} (+0.01) & 42.35 & 47.32 & 37.8/60.37 & \pos{61.10/48.54} \\
\rowcolor{rowgray}
\textit{ECDet-O365$\to$C Mean} & 9.89 & 3.51 & 57.30/\pos{57.32} & 40.29/\pos{40.31} & 11.04 & \textbf{3.56} ($+1.3\%$) & \pos{57.33} (+0.04) & 39.62 & \pos{40.35} & 11.04 & \textbf{3.56} ($+1.3\%$) & \pos{57.33} (+0.04) & 39.62 & \pos{40.35} & 29.2/57.89 & \pos{58.64/41.75} \\
\midrule
\texttt{rfdetr\_n} & 5.59 & 0.70 & 48.41/\pos{48.49} & 30.12/\negv{30.10} & 7.11 & \textbf{0.76} ($+8.7\%$) & \pos{49.53} (+1.12) & 27.13 & 30.47 & 7.11 & \textbf{0.76} ($+8.7\%$) & \pos{49.53} (+1.12) & 27.13 & 30.47 & 14.0/51.18 & \pos{54.63/33.58} \\
\texttt{rfdetr\_s} & 6.19 & 0.90 & 52.94/\pos{53.00} & 31.89/\pos{32.06} & 7.77 & \textbf{0.96} ($+6.8\%$) & \pos{53.73} (+0.79) & 34.33 & 32.15 & 7.77 & \textbf{0.96} ($+6.8\%$) & \pos{53.73} (+0.79) & 34.33 & 32.15 & 12.0/54.79 & \pos{57.65/34.29} \\
\texttt{rfdetr\_m} & 6.59 & 1.04 & 54.75/\pos{54.91} & 32.44/\pos{32.70} & 8.15 & \textbf{1.11} ($+6.3\%$) & \pos{55.22} (+0.47) & 36.52 & 32.78 & 8.15 & \textbf{1.11} ($+6.3\%$) & \pos{55.22} (+0.47) & 36.52 & 32.78 & 13.7/56.10 & \pos{58.68/34.15} \\
\texttt{rfdetr\_l} & 6.67 & 1.25 & 56.53/\pos{56.61} & 32.57/\negv{32.44} & 7.18 & \textbf{1.33} ($+6.5\%$) & \pos{56.88} (+0.35) & 39.28 & 32.67 & 7.18 & \textbf{1.33} ($+6.5\%$) & \pos{56.88} (+0.35) & 39.28 & 32.67 & 13.1/57.68 & \pos{59.11/33.27} \\
\rowcolor{rowgray}
\textit{RF-DETR Mean} & 6.26 & 0.97 & 53.16/\pos{53.25} & 31.75/\pos{31.82} & 7.55 & \textbf{1.04} ($+7.1\%$) & \pos{53.84} (+0.68) & 34.31 & \pos{32.02} & 7.55 & \textbf{1.04} ($+7.1\%$) & \pos{53.84} (+0.68) & 34.31 & \pos{32.02} & 13.2/54.94 & \pos{57.52/33.82} \\
\midrule
\rowcolor{rowgray}
\textbf{17-Model Mean} & 8.07 & \textbf{2.22} & 53.78/\pos{53.84} & 33.83/\pos{33.88} & 9.26 & \textbf{2.29} ($+4.1\%$) & \pos{54.08} ($+0.30$) & \pos{35.76} & \pos{33.94} & 8.97 & \textbf{2.29} ($+3.8\%$) & \pos{54.04} ($+0.26$) & \pos{35.72} & \pos{33.94} & 19.03/54.98 & \pos{55.91/35.44} \\
\bottomrule
\end{tabular}}
\vspace{-0.7em}
\end{table*}
\appnextpage

\begin{enumerate}
\setcounter{enumi}{1}
    \item \textbf{Per-Detector $\beta \le +10\%$ Audit \& Per-Scale $10\%$-Step Trajectories (\cref{fig:app_continuum}d, \cref{fig:app_arch_scale_lines}, \cref{tab:app_tta_all17,tab:app_dyn_all17_steps,tab:app_sg_all17_steps}):} \Cref{tab:app_tta_all17} reports per-checkpoint metrics at $\beta \le +10\%$ (PT1 ms, TRT $B{=}1$ ms/$\Delta\%$, COCO AP, $\text{AP}_s$, COCO-O mAP, and family sub-means), while \cref{fig:app_arch_scale_lines} and \cref{tab:app_dyn_all17_steps,tab:app_sg_all17_steps} trace every scale of YOLO26, EdgeCrafter (COCO \& O365$\to$COCO), and RF-DETR across $0\%$--$+100\%$. Every checkpoint ($0.52$--$0.70$~ms \texttt{yolo26n}/\texttt{rfdetr\_n} to $4.65$~ms \texttt{ecdet\_x}) obeys $T(g) \le (1+\beta)T_0$ and $\rho_b \ge 0.99$, letting smaller scales match or surpass next-larger backbones without retraining.
\end{enumerate}

\begin{table*}[!t]
\caption{\textbf{Complete per-detector $10\%$-step budget continuum ($\beta \in \{0\%, +10\%, +20\%, \dots, +100\%\}$) for \methodDyn (\cref{alg:dyn_graph}) across all 17 checkpoints and 4 detector families}, expanding \cref{tab:tta_vs_mem} and providing the exact numerical coordinates (\texttt{TRT B=1 ms / MS~COCO AP}) plotted in \cref{fig:app_arch_scale_lines} (\textbf{+0\%} reports the compiled single-head FP16 engine floor $d_0$; rightmost column lists holdout-selected \methodAuto at $\beta \le +100\%$).}
\label{tab:app_dyn_all17_steps}
\centering
\setlength{\tabcolsep}{3.0pt}
\resizebox{\textwidth}{!}{\begin{tabular}{@{}l|ccccccccccc|c@{}}
\toprule
\textbf{Detector} & \textbf{+0\%} & \textbf{+10\%} & \textbf{+20\%} & \textbf{+30\%} & \textbf{+40\%} & \textbf{+50\%} & \textbf{+60\%} & \textbf{+70\%} & \textbf{+80\%} & \textbf{+90\%} & \textbf{+100\%} & \textbf{Auto (+100\%)} \\
\midrule
\texttt{yolo26n} & 0.52/40.11 & \textbf{0.55}/\pos{40.74} & 0.55/\pos{40.74} & 0.55/\pos{40.74} & 0.55/\pos{40.74} & \textbf{0.55}/\pos{40.74} & 0.55/\pos{40.74} & 0.55/\pos{40.74} & 0.55/\pos{40.74} & 0.55/\pos{40.74} & \textbf{0.55}/\pos{40.74} & \textbf{1.00}/\pos{41.02} \\
\texttt{yolo26s} & 0.63/47.85 & 0.63/47.85 & 0.75/\pos{48.41} & 0.80/\pos{48.52} & 0.80/\pos{48.52} & \textbf{0.80}/\pos{48.52} & 0.80/\pos{48.52} & 0.80/\pos{48.52} & 0.80/\pos{48.52} & 0.80/\pos{48.52} & \textbf{0.80}/\pos{48.52} & \textbf{0.80}/\pos{48.52} \\
\texttt{yolo26m} & 1.01/52.52 & \textbf{1.07}/\pos{52.83} & 1.07/\pos{52.83} & 1.07/\pos{52.83} & 1.40/\pos{53.07} & \textbf{1.40}/\pos{53.07} & 1.40/\pos{53.07} & 1.40/\pos{53.07} & 1.40/\pos{53.07} & 1.40/\pos{53.07} & \textbf{1.40}/\pos{53.07} & \textbf{1.65}/\pos{53.27} \\
\texttt{yolo26l} & 1.42/54.37 & \textbf{1.45}/\pos{54.89} & 1.45/\pos{54.89} & 1.79/\pos{54.90} & 1.85/\pos{55.08} & \textbf{1.85}/\pos{55.08} & 1.85/\pos{55.08} & 1.85/\pos{55.08} & 1.85/\pos{55.08} & 1.85/\pos{55.08} & \textbf{1.85}/\pos{55.08} & \textbf{2.15}/\pos{55.15} \\
\texttt{yolo26x} & 2.06/56.90 & \textbf{2.16}/\pos{57.18} & 2.16/\pos{57.18} & 2.65/\pos{57.36} & 2.65/\pos{57.36} & \textbf{2.65}/\pos{57.36} & 2.65/\pos{57.36} & 2.65/\pos{57.36} & 2.65/\pos{57.36} & 2.65/\pos{57.36} & \textbf{2.65}/\pos{57.36} & \textbf{3.00}/\pos{57.40} \\
\rowcolor{rowgray}
\textit{YOLO26 Mean} & 1.13/50.35 & \textbf{1.17}/\pos{50.70} & \textbf{1.20}/\pos{50.81} & \textbf{1.37}/\pos{50.87} & \textbf{1.45}/\pos{50.95} & \textbf{1.45}/\pos{50.95} & \textbf{1.45}/\pos{50.95} & \textbf{1.45}/\pos{50.95} & \textbf{1.45}/\pos{50.95} & \textbf{1.45}/\pos{50.95} & \textbf{1.45}/\pos{50.95} & \textbf{1.72}/\pos{51.07} \\
\midrule
\texttt{ecdet\_s} & 2.17/51.73 & \textbf{2.30}/\pos{51.81} & 2.30/\pos{51.81} & 2.30/\pos{51.81} & 2.30/\pos{51.81} & \textbf{2.30}/\pos{51.81} & 3.41/\pos{52.11} & 3.41/\pos{52.11} & 3.41/\pos{52.11} & 3.41/\pos{52.11} & \textbf{3.41}/\pos{52.11} & \textbf{4.25}/\pos{52.21} \\
\texttt{ecdet\_m} & 3.01/54.22 & \textbf{3.15}/\pos{54.26} & 3.15/\pos{54.26} & 3.15/\pos{54.26} & 4.08/\pos{54.43} & \textbf{4.27}/\pos{54.58} & 4.27/\pos{54.58} & 4.27/\pos{54.58} & 4.27/\pos{54.58} & 5.56/\pos{54.66} & \textbf{5.56}/\pos{54.66} & \textbf{5.56}/\pos{54.66} \\
\texttt{ecdet\_l} & 4.30/56.99 & \textbf{4.44}/\pos{57.05} & 4.44/\pos{57.05} & 4.44/\pos{57.05} & 5.82/\pos{57.09} & \textbf{6.10}/\pos{57.28} & 6.10/\pos{57.28} & 7.23/\pos{57.37} & 7.23/\pos{57.37} & 7.23/\pos{57.37} & \textbf{7.23}/\pos{57.37} & \textbf{7.23}/\pos{57.37} \\
\texttt{ecdet\_x} & 4.65/57.88 & \textbf{4.81}/\pos{57.99} & 4.81/\pos{57.99} & 4.81/\pos{57.99} & 4.81/\pos{57.99} & \textbf{6.75}/\pos{58.17} & 6.75/\pos{58.17} & 7.80/\pos{58.31} & 7.80/\pos{58.31} & 7.80/\pos{58.31} & \textbf{7.80}/\pos{58.31} & \textbf{7.80}/\pos{58.31} \\
\rowcolor{rowgray}
\textit{ECDet-COCO Mean} & 3.53/55.21 & \textbf{3.67}/\pos{55.28} & \textbf{3.67}/\pos{55.28} & \textbf{3.67}/\pos{55.28} & \textbf{4.25}/\pos{55.33} & \textbf{4.85}/\pos{55.46} & \textbf{5.13}/\pos{55.54} & \textbf{5.68}/\pos{55.59} & \textbf{5.68}/\pos{55.59} & \textbf{6.00}/\pos{55.61} & \textbf{6.00}/\pos{55.61} & \textbf{6.21}/\pos{55.64} \\
\midrule
\texttt{ecdet\_o3652coco\_s} & 2.26/53.55 & \textbf{2.28}/\pos{53.65} & 2.28/\pos{53.65} & 2.28/\pos{53.65} & 2.28/\pos{53.65} & \textbf{2.28}/\pos{53.65} & 3.46/\pos{53.88} & 3.84/\pos{53.89} & 3.84/\pos{53.89} & 3.84/\pos{53.89} & \textbf{4.37}/\pos{54.05} & \textbf{3.95}/\pos{54.05} \\
\texttt{ecdet\_o3652coco\_m} & 3.01/56.76 & 3.01/\negv{56.73} & 3.01/\negv{56.73} & 3.01/\negv{56.73} & 3.01/\negv{56.73} & \textbf{4.25}/\pos{57.02} & 4.69/\pos{57.01} & 4.69/\pos{57.01} & 5.37/\pos{57.01} & 5.53/\pos{57.11} & \textbf{5.53}/\pos{57.11} & \textbf{5.53}/\pos{57.11} \\
\texttt{ecdet\_o3652coco\_l} & 4.31/59.04 & \textbf{4.47}/\pos{59.07} & 4.47/\pos{59.07} & 4.47/\pos{59.07} & 4.47/\pos{59.07} & \textbf{6.16}/\pos{59.21} & 6.63/\pos{59.27} & 6.63/\pos{59.27} & 7.37/\pos{59.43} & 7.37/\pos{59.43} & \textbf{7.37}/\pos{59.43} & \textbf{7.37}/\pos{59.43} \\
\texttt{ecdet\_o3652coco\_x} & 4.46/59.89 & 4.48/59.89 & 5.12/\pos{59.90} & 5.12/\pos{59.90} & 6.00/\negv{59.88} & \textbf{6.61}/\pos{59.90} & 6.61/\pos{59.90} & 7.19/\pos{60.08} & 7.76/\pos{60.22} & 7.76/\pos{60.22} & \textbf{7.76}/\pos{60.22} & \textbf{7.76}/\pos{60.22} \\
\rowcolor{rowgray}
\textit{ECDet-O365$\to$C Mean} & 3.51/57.31 & \textbf{3.56}/\pos{57.33} & \textbf{3.72}/\pos{57.34} & \textbf{3.72}/\pos{57.34} & \textbf{3.94}/\pos{57.33} & \textbf{4.83}/\pos{57.44} & \textbf{5.35}/\pos{57.52} & \textbf{5.59}/\pos{57.56} & \textbf{6.09}/\pos{57.63} & \textbf{6.12}/\pos{57.66} & \textbf{6.26}/\pos{57.70} & \textbf{6.15}/\pos{57.70} \\
\midrule
\texttt{rfdetr\_n} & 0.70/48.15 & \textbf{0.76}/\pos{49.53} & 0.82/\pos{51.09} & 0.89/\pos{51.54} & 0.96/\pos{52.48} & \textbf{1.00}/\pos{52.50} & 1.00/\pos{52.50} & 1.00/\pos{52.50} & 1.00/\pos{52.50} & 1.00/\pos{52.50} & \textbf{1.00}/\pos{52.50} & \textbf{1.00}/\pos{52.50} \\
\texttt{rfdetr\_s} & 0.90/53.01 & \textbf{0.96}/\pos{53.73} & 1.03/\pos{54.82} & 1.03/\pos{54.82} & 1.25/\pos{55.21} & \textbf{1.27}/\pos{55.53} & 1.41/\pos{55.81} & 1.41/\pos{55.81} & 1.41/\pos{55.81} & 1.41/\pos{55.81} & \textbf{1.41}/\pos{55.81} & \textbf{1.23}/\pos{55.81} \\
\texttt{rfdetr\_m} & 1.04/54.92 & \textbf{1.11}/\pos{55.22} & 1.11/\pos{55.22} & 1.11/\pos{55.22} & 1.45/\pos{57.07} & \textbf{1.45}/\pos{57.07} & 1.45/\pos{57.07} & 1.45/\pos{57.07} & 1.45/\pos{57.07} & 1.45/\pos{57.07} & \textbf{1.45}/\pos{57.07} & \textbf{1.45}/\pos{57.07} \\
\texttt{rfdetr\_l} & 1.25/56.58 & \textbf{1.33}/\pos{56.88} & 1.45/\pos{56.85} & 1.56/\pos{57.46} & 1.73/\pos{57.95} & \textbf{1.82}/\pos{58.00} & 1.82/\pos{58.00} & 1.82/\pos{58.00} & 1.82/\pos{58.00} & 1.82/\pos{58.00} & \textbf{1.82}/\pos{58.00} & \textbf{1.82}/\pos{58.00} \\
\rowcolor{rowgray}
\textit{RF-DETR Mean} & 0.97/53.16 & \textbf{1.04}/\pos{53.84} & \textbf{1.10}/\pos{54.49} & \textbf{1.15}/\pos{54.76} & \textbf{1.35}/\pos{55.68} & \textbf{1.39}/\pos{55.78} & \textbf{1.42}/\pos{55.84} & \textbf{1.42}/\pos{55.84} & \textbf{1.42}/\pos{55.84} & \textbf{1.42}/\pos{55.84} & \textbf{1.42}/\pos{55.84} & \textbf{1.38}/\pos{55.84} \\
\midrule
\rowcolor{rowgray}
\textbf{17-Model Mean} & \textbf{2.22}/53.79 & \textbf{2.29}/\pos{54.08} & \textbf{2.35}/\pos{54.26} & \textbf{2.41}/\pos{54.34} & \textbf{2.67}/\pos{54.60} & \textbf{3.03}/\pos{54.68} & \textbf{3.23}/\pos{54.73} & \textbf{3.41}/\pos{54.75} & \textbf{3.53}/\pos{54.77} & \textbf{3.61}/\pos{54.78} & \textbf{3.65}/\pos{54.79} & \textbf{3.74}/\pos{54.83} \\
\bottomrule
\end{tabular}}
\end{table*}

\begin{table*}[!t]
\caption{\textbf{Complete per-detector $10\%$-step budget continuum ($\beta \in \{0\%, +10\%, +20\%, \dots, +100\%\}$) for \methodSG (\cref{alg:static_graph}) across all 17 checkpoints and 4 detector families}, expanding \cref{tab:tta_vs_mem} and providing the exact numerical coordinates (\texttt{TRT B=1 ms / MS~COCO AP}) plotted in \cref{fig:app_arch_scale_lines} (\textbf{+0\%} reports the compiled single-head FP16 engine floor $d_0$; rightmost column lists holdout-selected \methodAuto at $\beta \le +100\%$).}
\label{tab:app_sg_all17_steps}
\centering
\setlength{\tabcolsep}{3.0pt}
\resizebox{\textwidth}{!}{\begin{tabular}{@{}l|ccccccccccc|c@{}}
\toprule
\textbf{Detector} & \textbf{+0\%} & \textbf{+10\%} & \textbf{+20\%} & \textbf{+30\%} & \textbf{+40\%} & \textbf{+50\%} & \textbf{+60\%} & \textbf{+70\%} & \textbf{+80\%} & \textbf{+90\%} & \textbf{+100\%} & \textbf{Auto (+100\%)} \\
\midrule
\texttt{yolo26n} & 0.52/40.11 & 0.52/40.11 & 0.52/40.11 & 0.52/40.11 & 0.52/40.11 & 0.52/40.11 & 0.81/\pos{40.50} & 0.81/\pos{40.50} & 0.81/\pos{40.50} & 0.81/\pos{40.50} & \textbf{1.00}/\pos{41.02} & \textbf{1.00}/\pos{41.02} \\
\texttt{yolo26s} & 0.63/47.85 & 0.63/47.85 & 0.63/47.85 & 0.63/47.85 & 0.63/47.85 & 0.63/47.85 & 0.63/47.85 & 0.63/47.85 & 1.12/\pos{48.52} & 1.12/\pos{48.52} & \textbf{1.12}/\pos{48.52} & \textbf{0.80}/\pos{48.52} \\
\texttt{yolo26m} & 1.01/52.52 & \textbf{1.07}/\pos{52.83} & 1.07/\pos{52.83} & 1.07/\pos{52.83} & 1.07/\pos{52.83} & \textbf{1.07}/\pos{52.83} & 1.07/\pos{52.83} & 1.65/\pos{53.27} & 1.65/\pos{53.27} & 1.65/\pos{53.27} & \textbf{1.65}/\pos{53.27} & \textbf{1.65}/\pos{53.27} \\
\texttt{yolo26l} & 1.42/54.37 & \textbf{1.45}/\pos{54.89} & 1.45/\pos{54.89} & 1.45/\pos{54.89} & 1.45/\pos{54.89} & \textbf{1.45}/\pos{54.89} & 2.15/\pos{55.15} & 2.15/\pos{55.15} & 2.15/\pos{55.15} & 2.15/\pos{55.15} & \textbf{2.15}/\pos{55.15} & \textbf{2.15}/\pos{55.15} \\
\texttt{yolo26x} & 2.06/56.90 & \textbf{2.16}/\pos{57.18} & 2.16/\pos{57.18} & 2.16/\pos{57.18} & 2.16/\pos{57.18} & \textbf{3.00}/\pos{57.40} & 3.00/\pos{57.40} & 3.00/\pos{57.40} & 3.00/\pos{57.40} & 3.00/\pos{57.40} & \textbf{3.00}/\pos{57.40} & \textbf{3.00}/\pos{57.40} \\
\rowcolor{rowgray}
\textit{YOLO26 Mean} & 1.13/50.35 & \textbf{1.17}/\pos{50.57} & \textbf{1.17}/\pos{50.57} & \textbf{1.17}/\pos{50.57} & \textbf{1.17}/\pos{50.57} & \textbf{1.33}/\pos{50.62} & \textbf{1.53}/\pos{50.75} & \textbf{1.65}/\pos{50.83} & \textbf{1.75}/\pos{50.97} & \textbf{1.75}/\pos{50.97} & \textbf{1.78}/\pos{51.07} & \textbf{1.72}/\pos{51.07} \\
\midrule
\texttt{ecdet\_s} & 2.17/51.73 & \textbf{2.30}/\pos{51.81} & 2.30/\pos{51.81} & 2.30/\pos{51.81} & 2.30/\pos{51.81} & \textbf{2.30}/\pos{51.81} & 2.30/\pos{51.81} & 2.30/\pos{51.81} & 3.81/\pos{51.98} & 4.00/\pos{52.11} & \textbf{4.25}/\pos{52.21} & \textbf{4.25}/\pos{52.21} \\
\texttt{ecdet\_m} & 3.01/54.22 & \textbf{3.15}/\pos{54.26} & 3.15/\pos{54.26} & 3.15/\pos{54.26} & 3.15/\pos{54.26} & \textbf{3.15}/\pos{54.26} & 4.72/\pos{54.43} & 4.94/\pos{54.58} & 4.94/\pos{54.58} & 4.94/\pos{54.58} & \textbf{5.82}/\pos{54.66} & \textbf{5.56}/\pos{54.66} \\
\texttt{ecdet\_l} & 4.30/56.99 & \textbf{4.44}/\pos{57.05} & 4.44/\pos{57.05} & 4.44/\pos{57.05} & 4.44/\pos{57.05} & \textbf{4.44}/\pos{57.05} & 6.72/\pos{57.09} & 7.19/\pos{57.20} & 7.19/\pos{57.20} & 7.76/\pos{57.37} & \textbf{7.76}/\pos{57.37} & \textbf{7.23}/\pos{57.37} \\
\texttt{ecdet\_x} & 4.65/57.88 & \textbf{4.81}/\pos{57.99} & 4.81/\pos{57.99} & 4.81/\pos{57.99} & 4.81/\pos{57.99} & \textbf{4.81}/\pos{57.99} & 4.81/\pos{57.99} & 7.63/\pos{58.17} & 7.63/\pos{58.17} & 8.40/\pos{58.31} & \textbf{8.40}/\pos{58.31} & \textbf{7.80}/\pos{58.31} \\
\rowcolor{rowgray}
\textit{ECDet-COCO Mean} & 3.53/55.21 & \textbf{3.67}/\pos{55.28} & \textbf{3.67}/\pos{55.28} & \textbf{3.67}/\pos{55.28} & \textbf{3.67}/\pos{55.28} & \textbf{3.67}/\pos{55.28} & \textbf{4.64}/\pos{55.33} & \textbf{5.51}/\pos{55.44} & \textbf{5.89}/\pos{55.48} & \textbf{6.27}/\pos{55.59} & \textbf{6.56}/\pos{55.64} & \textbf{6.21}/\pos{55.64} \\
\midrule
\texttt{ecdet\_o3652coco\_s} & 2.26/53.55 & \textbf{2.28}/\pos{53.65} & 2.28/\pos{53.65} & 2.28/\pos{53.65} & 2.28/\pos{53.65} & \textbf{2.28}/\pos{53.65} & 2.28/\pos{53.65} & 2.28/\pos{53.65} & 3.95/\pos{54.05} & 3.95/\pos{54.05} & \textbf{3.95}/\pos{54.05} & \textbf{3.95}/\pos{54.05} \\
\texttt{ecdet\_o3652coco\_m} & 3.01/56.76 & 3.01/\negv{56.73} & 3.01/\negv{56.73} & 3.01/\negv{56.73} & 3.01/\negv{56.73} & 3.01/\negv{56.73} & 3.01/\negv{56.73} & 4.93/\pos{57.02} & 5.20/\pos{57.01} & 5.20/\pos{57.01} & \textbf{5.79}/\pos{57.11} & \textbf{5.53}/\pos{57.11} \\
\texttt{ecdet\_o3652coco\_l} & 4.31/59.04 & \textbf{4.47}/\pos{59.07} & 4.47/\pos{59.07} & 4.47/\pos{59.07} & 4.47/\pos{59.07} & \textbf{6.08}/\pos{59.11} & 6.08/\pos{59.11} & 7.21/\pos{59.21} & 7.21/\pos{59.21} & 7.76/\pos{59.43} & \textbf{7.76}/\pos{59.43} & \textbf{7.37}/\pos{59.43} \\
\texttt{ecdet\_o3652coco\_x} & 4.46/59.89 & 4.48/59.89 & 5.12/\pos{59.90} & 5.12/\pos{59.90} & 6.00/\negv{59.88} & \textbf{6.61}/\pos{59.90} & 6.61/\pos{59.90} & 6.61/\pos{59.90} & 6.61/\pos{59.90} & 8.36/\pos{60.22} & \textbf{8.36}/\pos{60.22} & \textbf{7.76}/\pos{60.22} \\
\rowcolor{rowgray}
\textit{ECDet-O365$\to$C Mean} & 3.51/57.31 & \textbf{3.56}/\pos{57.33} & \textbf{3.72}/\pos{57.34} & \textbf{3.72}/\pos{57.34} & \textbf{3.94}/\pos{57.33} & \textbf{4.50}/\pos{57.35} & \textbf{4.50}/\pos{57.35} & \textbf{5.26}/\pos{57.44} & \textbf{5.74}/\pos{57.54} & \textbf{6.32}/\pos{57.68} & \textbf{6.47}/\pos{57.70} & \textbf{6.15}/\pos{57.70} \\
\midrule
\texttt{rfdetr\_n} & 0.70/48.15 & \textbf{0.76}/\pos{49.53} & 0.82/\pos{51.09} & 0.89/\pos{51.54} & 0.96/\pos{52.48} & \textbf{1.00}/\pos{52.50} & 1.00/\pos{52.50} & 1.00/\pos{52.50} & 1.00/\pos{52.50} & 1.00/\pos{52.50} & \textbf{1.00}/\pos{52.50} & \textbf{1.00}/\pos{52.50} \\
\texttt{rfdetr\_s} & 0.90/53.01 & \textbf{0.96}/\pos{53.73} & 1.03/\pos{54.82} & 1.12/\pos{55.53} & 1.23/\pos{55.81} & \textbf{1.23}/\pos{55.81} & 1.23/\pos{55.81} & 1.23/\pos{55.81} & 1.23/\pos{55.81} & 1.23/\pos{55.81} & \textbf{1.23}/\pos{55.81} & \textbf{1.23}/\pos{55.81} \\
\texttt{rfdetr\_m} & 1.04/54.92 & \textbf{1.11}/\pos{55.22} & 1.21/\pos{55.99} & 1.32/\pos{56.46} & 1.45/\pos{57.07} & \textbf{1.45}/\pos{57.07} & 1.45/\pos{57.07} & 1.45/\pos{57.07} & 1.45/\pos{57.07} & 1.45/\pos{57.07} & \textbf{1.45}/\pos{57.07} & \textbf{1.45}/\pos{57.07} \\
\texttt{rfdetr\_l} & 1.25/56.58 & \textbf{1.33}/\pos{56.88} & 1.45/\pos{56.85} & 1.59/\pos{57.72} & 1.73/\pos{57.95} & \textbf{1.82}/\pos{58.00} & 1.82/\pos{58.00} & 1.82/\pos{58.00} & 1.82/\pos{58.00} & 1.82/\pos{58.00} & \textbf{1.82}/\pos{58.00} & \textbf{1.82}/\pos{58.00} \\
\rowcolor{rowgray}
\textit{RF-DETR Mean} & 0.97/53.16 & \textbf{1.04}/\pos{53.84} & \textbf{1.13}/\pos{54.69} & \textbf{1.23}/\pos{55.31} & \textbf{1.34}/\pos{55.83} & \textbf{1.38}/\pos{55.84} & \textbf{1.38}/\pos{55.84} & \textbf{1.38}/\pos{55.84} & \textbf{1.38}/\pos{55.84} & \textbf{1.38}/\pos{55.84} & \textbf{1.38}/\pos{55.84} & \textbf{1.38}/\pos{55.84} \\
\midrule
\rowcolor{rowgray}
\textbf{17-Model Mean} & \textbf{2.22}/53.79 & \textbf{2.29}/\pos{54.04} & \textbf{2.35}/\pos{54.24} & \textbf{2.37}/\pos{54.39} & \textbf{2.45}/\pos{54.51} & \textbf{2.64}/\pos{54.53} & \textbf{2.92}/\pos{54.58} & \textbf{3.34}/\pos{54.65} & \textbf{3.57}/\pos{54.72} & \textbf{3.80}/\pos{54.78} & \textbf{3.91}/\pos{54.83} & \textbf{3.74}/\pos{54.83} \\
\bottomrule
\end{tabular}}
\end{table*}
\appnextpage

\begin{figure*}[!t]
    \centering
    \includegraphics[width=0.92\linewidth]{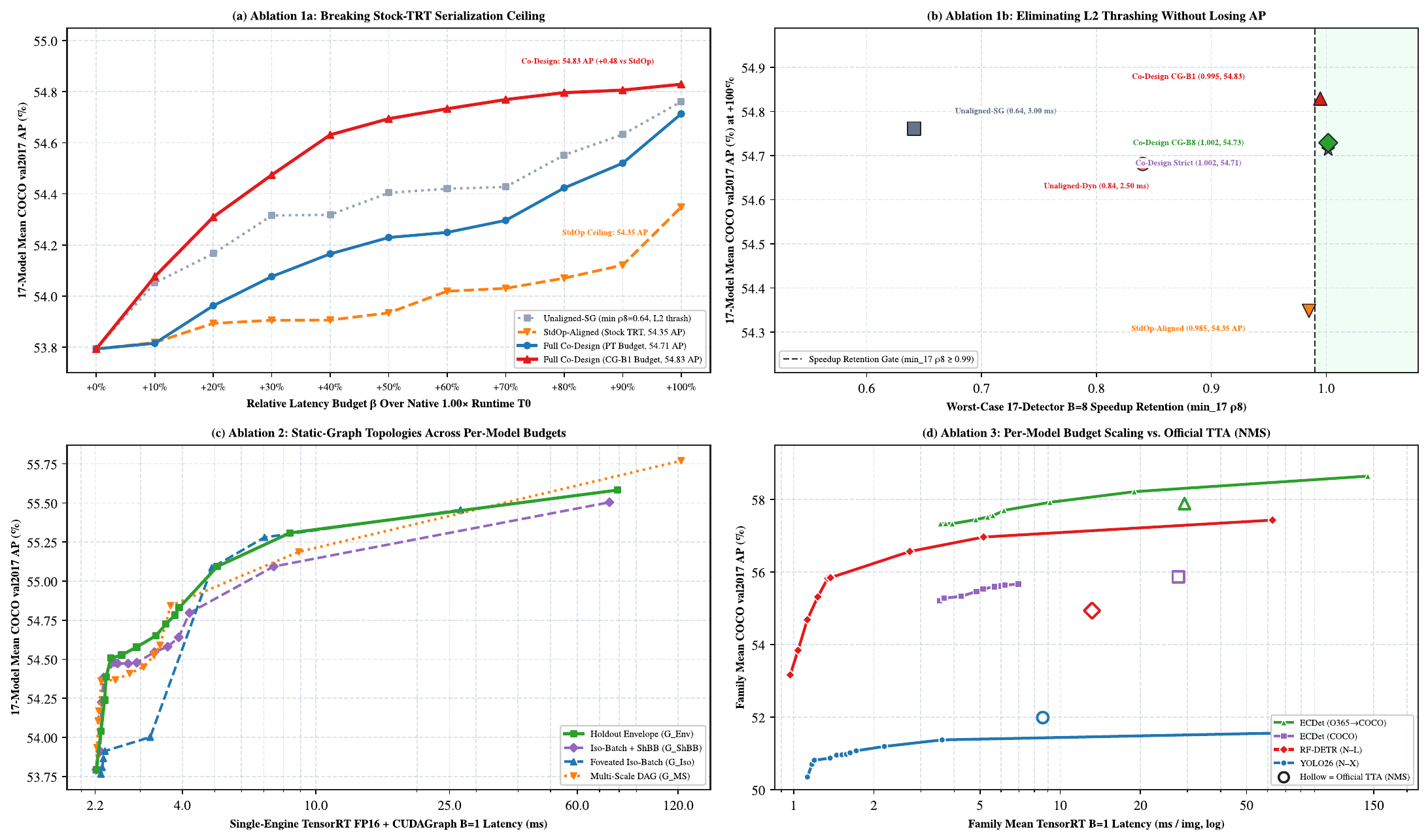}
    \vspace{-0.5em}
    \caption{\textbf{Controlled ablations on dynamic and static inference-graph design under per-model relative budgets (supplementing \cref{sec:method_graph,sec:ablations}).} \textbf{(a)} Breaking stock-TensorRT serialization ceiling across $\beta \in [0\%, +100\%]$ ($54.35 \to 54.83$~AP). \textbf{(b)} Eliminating $B{=}8$ L2-cache thrashing ($\min_{17}\rho_8: 0.64 \to \ge 0.99$) without losing accuracy. \textbf{(c)} Complementarity of the three monolithic static-graph topologies ($\mathcal{G}_{\text{MS}}, \mathcal{G}_{\text{Iso}}, \mathcal{G}_{\text{ShBB}}$) and their envelope ($\mathcal{G}_{\text{Env}}$) across per-model relative budgets $\beta$. \textbf{(d)} Per-model relative-budget scaling across all four detector families vs.\ Official TTA (NMS) on \texttt{TensorRT} $B{=}1$ latency.}
    \label{fig:app_graph_ablations}
    \vspace{-0.3em}
\end{figure*}
\null\newpage\null\newpage

\subsection{Controlled Ablations on Dynamic and Static Inference-Graph Design}
\label{sec:app_graph_ablations}

Supplementing \cref{sec:method_graph,sec:ablations}, \cref{fig:app_graph_ablations} (above) and \cref{tab:app_codesign_ablation,tab:app_sg_topologies} isolate the three core design choices behind \methodDyn and \methodSG under per-model relative latency budgets $\beta$:
\begin{enumerate}
    \item \textbf{Why Algorithm--Toolchain Co-Design Is Required (\cref{fig:app_graph_ablations}a--b, \cref{tab:app_codesign_ablation}):} Comparing three regimes side-by-side across $\beta \in [0\%, +100\%]$ shows that (i)~\emph{Speedup-Unaligned Expansion} reaches $54.68$/$54.76$~AP but suffers $B{=}8$ L2-cache thrashing ($\min_{17}\rho_8 = 0.64$, $3.00$~ms) from unchunked $[K{\cdot}8]$ views and FP32 \texttt{Pow} gates; (ii)~\emph{Stock-Operator Alignment} (\texttt{StdOp}: horizontal Conv fusion, nested \texttt{Sqrt}, and $2{\times}B{=}4$ L2-chunked \texttt{CUDAGraph}) cuts $B{=}8$ latency ($3.00 \to 1.99$~ms, $\min_{17}\rho_8 = 0.985$) but stalls at $54.35$~AP due to serialized multi-group decoder/grid kernels; and (iii)~\emph{Full Co-Design} breaks this serialization ceiling via custom \texttt{IPluginV3} kernels, recovering $+0.48$~AP ($54.83$~AP) with $\min_{17}\rho_b \ge 0.99$.
    \item \textbf{Complementarity of Static-Graph Topologies Across Budgets (\cref{fig:app_graph_ablations}c, \cref{tab:app_sg_topologies}):} Evaluating $\mathcal{G}_{\text{MS}}$ (multi-scale DAG), $\mathcal{G}_{\text{Iso}}$ ($[K{\cdot}B,3,H^\star,W^\star]$ foveated iso-batch), and $\mathcal{G}_{\text{ShBB}}$ (shared-backbone multi-Stiefel fan-out) reveals three hardware regimes: (i)~at tight budgets ($+10\%$--$+50\%$), $\mathcal{G}_{\text{ShBB}}$ dominates ($12/17$ models at $+10\%$, $53.91 \to 54.48$~AP) by running the backbone once; (ii)~at $+60\%$--$+100\%$, $\mathcal{G}_{\text{MS}}$ and $\mathcal{G}_{\text{Iso}}$ join $\mathcal{G}_{\text{ShBB}}$ to lift $\mathcal{G}_{\text{Env}}$ (\methodSG) to $54.83$~AP ($3.91$~ms \texttt{CG1}); and (iii)~at $+200\%$--\texttt{Full}, $\mathcal{G}_{\text{Iso}}$ maximizes L2 weight reuse ($16/17$ models at $+200\%$, $55.14$~AP), $\mathcal{G}_{\text{MS}}$ unlocks the spatial ceiling ($55.77$~AP), $\mathcal{G}_{\text{ShBB}}$ maximizes COCO-O transfer ($35.47$~mAP), and $\mathcal{G}_{\text{Env}}$ reaches $55.79$~AP / $35.41$~mAP.
    \item \textbf{Cross-Family Per-Model Budget Scaling vs.\ Official TTA (\cref{fig:app_graph_ablations}d, \cref{tab:app_sg_topologies}):} Formulating \cref{alg:dyn_graph,alg:static_graph} against each model's own $T_0$ scales all four families without cross-scale subsidization: on \texttt{TensorRT} $B{=}1$, RF-DETR surpasses Official TTA ($13.16$~ms, $54.94$~AP) already at $\beta \le +30\%$ ($1.23$~ms, $55.31$~AP, $10.7\times$ faster; $55.84$~AP at $+50\%$), ECDet-COCO climbs monotonically from $55.19$ to $55.87$~AP at $+200\%$ ($56.58$~AP in \methodplus, $+0.71$~AP over $27.7$~ms Official TTA [$55.87$~AP]), ECDet-O365$\to$COCO overtakes Official TTA ($29.25$~ms, $57.89$~AP) by $+200\%$ ($57.92$~AP; $58.64$~AP in \methodplus), and YOLO26 scales monotonically from $50.35$ ($50.42$ with gate) to $51.07$~AP at $+100\%$ ($51.91$~AP in \methodplus).
\end{enumerate}

\begin{table*}[!t]
\caption{\textbf{Controlled Ablation 1: Speedup-Unaligned vs.\ Stock-Operator Speedup-Aligned vs.\ Full Algorithm--Toolchain Co-Design} across budgets $\beta \in \{0\%, +10\%, +20\%, +30\%, +50\%, +70\%, +100\%\}$ on all 17 detectors. Without $\rho_b$ constraints, \texttt{Unaligned} suffers $B{=}8$ L2-cache thrashing ($\min_{17}\rho_8{=}0.64$, $3.00$~ms); stock-operator alignment (\texttt{StdOp}) cuts $B{=}8$ latency ($1.99$~ms, $\min_{17}\rho_8{=}0.985$) but stalls at $54.35$~AP; full co-design reclaims $54.83$~AP while maintaining $\min_{17}\rho_b \ge 0.99$.}
\label{tab:app_codesign_ablation}
\centering
\resizebox{\textwidth}{!}{\begin{tabular}{@{}l|ccc|ccc|ccc|ccc|ccc@{}}
\toprule
& \multicolumn{3}{c|}{\textbf{(i) \texttt{Unaligned-SG} (No $\rho_b$ Gate)}} & \multicolumn{3}{c|}{\textbf{(i) \texttt{Unaligned-Dyn} (No $\rho_b$ Gate)}} & \multicolumn{3}{c|}{\textbf{(ii) \texttt{StdOp-Auto} (Stock TRT)}} & \multicolumn{3}{c|}{\textbf{(iii) Co-Design (CG-$B{=}8$-Bud.)}} & \multicolumn{3}{c}{\textbf{(iii) Co-Design (CG-$B{=}1$-Bud.)}} \\
\textbf{Budget $\beta$} & \textbf{CG1/CG8} & \textbf{$\min\rho_8$($\overline{\rho}_8$)} & \textbf{ID / OOD} & \textbf{CG1/CG8} & \textbf{$\min\rho_8$($\overline{\rho}_8$)} & \textbf{ID / OOD} & \textbf{CG1/CG8} & \textbf{$\min\rho_8$($\overline{\rho}_8$)} & \textbf{ID / OOD} & \textbf{CG8} & \textbf{$\min\rho_8$($\overline{\rho}_8$)} & \textbf{ID / OOD} & \textbf{CG1} & \textbf{$\min\rho_1$($\overline{\rho}_1$)} & \textbf{ID / OOD} \\
\midrule
\textbf{+0\%} & 2.22/1.62 & 1.00(1.00) & 53.79/33.87 & 2.22/1.62 & 1.00(1.00) & 53.79/33.87 & \textbf{2.22/1.62} & 1.00(1.00) & 53.79/33.87 & \textbf{1.62} & 1.00(1.00) & 53.79/33.87 & \textbf{2.22} & 1.00(1.00) & 53.79/33.87 \\
\textbf{+10\%} & 2.23/1.68 & \negv{0.86}(0.97) & 54.05/34.06 & 2.20/1.65 & \negv{0.91}(0.99) & 54.06/34.06 & \textbf{2.25/1.65} & 1.00(1.01) & 53.82/33.91 & \textbf{1.67} & 1.00(1.02) & \pos{53.83}/\pos{33.92} & \textbf{2.29} & 1.00(1.12) & \pos{54.08}/\pos{33.94} \\
\textbf{+20\%} & 2.29/1.70 & \negv{0.86}(0.98) & 54.17/34.09 & 2.23/1.67 & \negv{0.84}(0.98) & 54.11/34.09 & \textbf{2.31/1.67} & 1.00(1.03) & 53.89/33.99 & \textbf{1.72} & 1.00(1.04) & \pos{53.98}/\pos{33.91} & \textbf{2.36} & \pos{1.01}(1.17) & \pos{54.31}/\pos{34.02} \\
\textbf{+30\%} & 2.34/1.73 & \negv{0.77}(0.98) & 54.32/34.14 & 2.27/1.69 & \negv{0.84}(0.98) & 54.19/34.10 & \textbf{2.34/1.69} & 1.00(1.04) & 53.91/33.99 & \textbf{1.78} & 1.00(1.05) & \pos{54.08}/\pos{33.94} & \textbf{2.43} & \pos{1.01}(1.28) & \pos{54.47}/\pos{34.07} \\
\textbf{+50\%} & 2.54/1.83 & \negv{0.64}(0.99) & 54.40/34.17 & 2.40/1.75 & \negv{0.84}(0.99) & 54.40/34.15 & \textbf{2.49/1.78} & 1.00(1.07) & 53.93/34.00 & \textbf{2.00} & 1.00(1.15) & \pos{54.36}/\pos{34.14} & \textbf{3.05} & \pos{1.03}(1.36) & \pos{54.69}/\pos{34.21} \\
\textbf{+70\%} & 2.79/2.00 & \negv{0.64}(1.01) & 54.43/34.22 & 2.51/1.79 & \negv{0.84}(1.00) & 54.45/34.14 & \textbf{2.55/1.81} & 0.99(1.08) & 54.03/34.01 & \textbf{2.60} & \pos{1.04}(1.20) & \pos{54.56}/\pos{34.32} & \textbf{3.45} & \pos{1.18}(1.41) & \pos{54.77}/\pos{34.20} \\
\rowcolor{rowgray}
\textbf{+100\%} & 3.76/3.00 & \negv{0.64}(0.96) & 54.76/34.47 & 3.27/2.50 & \negv{0.84}(1.00) & 54.68/34.30 & \textbf{2.82/1.99} & 0.99(1.11) & 54.35/34.08 & \textbf{2.72} & 1.00(1.19) & \pos{54.73}/\pos{34.25} & \textbf{3.74} & \pos{1.14}(1.35) & \pos{54.83}/\pos{34.18} \\
\bottomrule
\end{tabular}}
\end{table*}

\begin{table*}[!t]
\caption{\textbf{Controlled Ablations 2 \& 3: Monolithic Single-Engine Static-Graph Topologies ($\mathcal{G}_{\text{MS}}, \mathcal{G}_{\text{Iso}}, \mathcal{G}_{\text{ShBB}}$) and Holdout Envelope ($\mathcal{G}_{\text{Env}}$) Across Per-Model Relative Latency Budgets $\beta$} on all 17 detectors (from $\beta{=}0\%$ to $+100\%$ and extended multi-view regimes $+200\%, +500\%$, \texttt{Full}; vs.\ Official TTA [NMS]). $\mathcal{G}_{\text{ShBB}}$ dominates tight sub-2-view budgets ($+10\%$--$+50\%$) via single-backbone Stiefel fan-out, $\mathcal{G}_{\text{Iso}}$ dominates multi-view budgets ($+200\%$--$+500\%$) via L2 weight reuse, and $\mathcal{G}_{\text{MS}}$ unlocks the spatial ceiling at \texttt{Full} while $\mathcal{G}_{\text{ShBB}}$ maximizes zero-shot OOD.}
\label{tab:app_sg_topologies}
\centering
\resizebox{\textwidth}{!}{\begin{tabular}{@{}l|cc|cc|cc|ccc|cccc@{}}
\toprule
& \multicolumn{2}{c|}{\textbf{Multi-Scale ($\mathcal{G}_{\text{MS}}$)}} & \multicolumn{2}{c|}{\textbf{Foveated Iso ($\mathcal{G}_{\text{Iso}}$)}} & \multicolumn{2}{c|}{\textbf{Iso+ShBB ($\mathcal{G}_{\text{ShBB}}$)}} & \multicolumn{3}{c|}{\textbf{Static Envelope ($\mathcal{G}_{\text{Env}}$)}} & \multicolumn{4}{c}{\textbf{Family $\mathcal{G}_{\text{Env}}$ COCO AP}} \\
\textbf{Budget $\beta$} & \textbf{CG1/8} & \textbf{ID / OOD} & \textbf{CG1/8} & \textbf{ID / OOD} & \textbf{CG1/8} & \textbf{ID / OOD} & \textbf{CG1/8} & \textbf{ID / OOD} & \textbf{Mix} & \textbf{Y26} & \textbf{EC} & \textbf{EC$_{\text{O}\to\text{C}}$} & \textbf{RF} \\
\midrule
\textbf{+0\%} & 2.22/1.62 & 53.79/33.87 & 2.22/1.62 & 53.79/33.87 & 2.22/1.62 & 53.79/33.87 & \textbf{2.22/1.62} & 53.79/33.87 & 0/0/0 & 50.35 & 55.21 & 57.31 & 53.16 \\
\textbf{+10\%} & 2.23/1.63 & \pos{53.93}/\pos{33.90} & 2.22/1.62 & 53.79/33.87 & 2.28/1.68 & \pos{53.91}/\pos{33.92} & \textbf{2.29/1.69} & \pos{54.04}/\pos{33.94} & 3/0/12 & \pos{50.57} & \pos{55.28} & \pos{57.33} & \pos{53.84} \\
\textbf{+20\%} & 2.25/1.65 & \pos{54.11}/\pos{33.95} & 2.22/1.62 & 53.79/33.87 & 2.34/1.72 & \pos{54.22}/\pos{33.98} & \textbf{2.35/1.72} & \pos{54.24}/\pos{33.98} & 3/0/12 & \pos{50.57} & \pos{55.28} & \pos{57.34} & \pos{54.69} \\
\textbf{+30\%} & 2.26/1.66 & \pos{54.17}/\pos{33.96} & 2.29/1.70 & \negv{53.77}/\pos{34.12} & 2.37/1.75 & \pos{54.38}/\pos{33.98} & \textbf{2.37/1.75} & \pos{54.39}/\pos{33.98} & 1/0/14 & \pos{50.57} & \pos{55.28} & \pos{57.34} & \pos{55.31} \\
\textbf{+40\%} & 2.29/1.70 & \pos{54.36}/33.87 & 2.29/1.70 & \negv{53.77}/\pos{34.12} & 2.45/1.79 & \pos{54.48}/\negv{33.85} & \textbf{2.45/1.79} & \pos{54.51}/33.87 & 3/0/12 & \pos{50.57} & \pos{55.28} & \pos{57.33} & \pos{55.83} \\
\textbf{+50\%} & 2.53/1.80 & \pos{54.37}/\pos{33.95} & 2.31/1.73 & \pos{53.81}/\pos{34.06} & 2.45/1.79 & \pos{54.48}/\negv{33.85} & \textbf{2.64/1.91} & \pos{54.53}/\pos{33.95} & 5/1/9 & \pos{50.62} & \pos{55.28} & \pos{57.35} & \pos{55.84} \\
\textbf{+60\%} & 2.95/2.04 & \pos{54.42}/\pos{34.05} & 2.33/1.75 & \pos{53.86}/\pos{33.99} & 2.51/1.82 & \pos{54.48}/\negv{33.85} & \textbf{2.92/2.09} & \pos{54.58}/\pos{34.03} & 8/2/6 & \pos{50.75} & \pos{55.33} & \pos{57.35} & \pos{55.84} \\
\textbf{+70\%} & 3.19/2.30 & \pos{54.46}/\pos{34.11} & 2.35/1.77 & \pos{53.91}/\pos{33.93} & 2.79/1.98 & \pos{54.50}/\pos{33.91} & \textbf{3.34/2.47} & \pos{54.65}/\pos{34.14} & 10/3/3 & \pos{50.83} & \pos{55.44} & \pos{57.44} & \pos{55.84} \\
\textbf{+80\%} & 3.41/2.41 & \pos{54.54}/\pos{34.19} & 2.35/1.77 & \pos{53.91}/\pos{33.93} & 3.24/2.40 & \pos{54.56}/\pos{34.00} & \textbf{3.57/2.66} & \pos{54.72}/\pos{34.21} & 11/3/3 & \pos{50.97} & \pos{55.48} & \pos{57.54} & \pos{55.84} \\
\textbf{+90\%} & 3.42/2.43 & \pos{54.54}/\pos{34.20} & 3.21/2.70 & \pos{54.00}/\pos{33.99} & 3.69/2.70 & \pos{54.60}/\pos{34.05} & \textbf{3.80/3.16} & \pos{54.78}/\pos{34.18} & 7/7/3 & \pos{50.97} & \pos{55.59} & \pos{57.68} & \pos{55.84} \\
\rowcolor{rowgray}
\textbf{+100\%} & 3.55/2.49 & \pos{54.60}/\pos{34.29} & 3.21/2.70 & \pos{54.00}/\pos{33.99} & 3.90/3.23 & \pos{54.64}/\pos{33.99} & \textbf{3.91/3.40} & \pos{54.83}/\pos{34.18} & 5/7/5 & \pos{51.07} & \pos{55.64} & \pos{57.70} & \pos{55.84} \\
\midrule
\textbf{+200\%} & 4.10/3.06 & \pos{54.86}/\pos{34.49} & 5.47/5.01 & \pos{55.13}/\pos{34.74} & 4.55/3.83 & \pos{54.85}/\pos{34.43} & \textbf{5.50/5.02} & \pos{55.14}/\pos{34.73} & 1/16/0 & \pos{51.18} & \pos{55.87} & \pos{57.92} & \pos{56.57} \\
\textbf{+500\%} & 8.93/7.92 & \pos{55.18}/\pos{34.87} & 9.81/9.40 & \pos{55.40}/\pos{35.03} & 8.74/7.51 & \pos{55.14}/\pos{34.84} & \textbf{10.91/10.54} & \pos{55.42}/\pos{35.03} & 2/14/1 & \pos{51.35} & \pos{56.16} & \pos{58.22} & \pos{56.96} \\
\rowcolor{rowgray}
\textbf{\texttt{Full}} & 122.5/89.8 & \pos{55.77}/\pos{35.28} & 37.0/36.4 & \pos{55.62}/\pos{35.31} & 109.4/83.5 & \pos{55.70}/\pos{35.47} & \textbf{116.7/88.5} & \pos{55.79}/\pos{35.41} & 11/1/5 & \pos{51.57} & \pos{56.59} & \pos{58.64} & \pos{57.43} \\
\bottomrule
\end{tabular}}
\end{table*}
\appnextpage

\begin{figure*}[!t]
    \centering
    \includegraphics[width=0.68\linewidth]{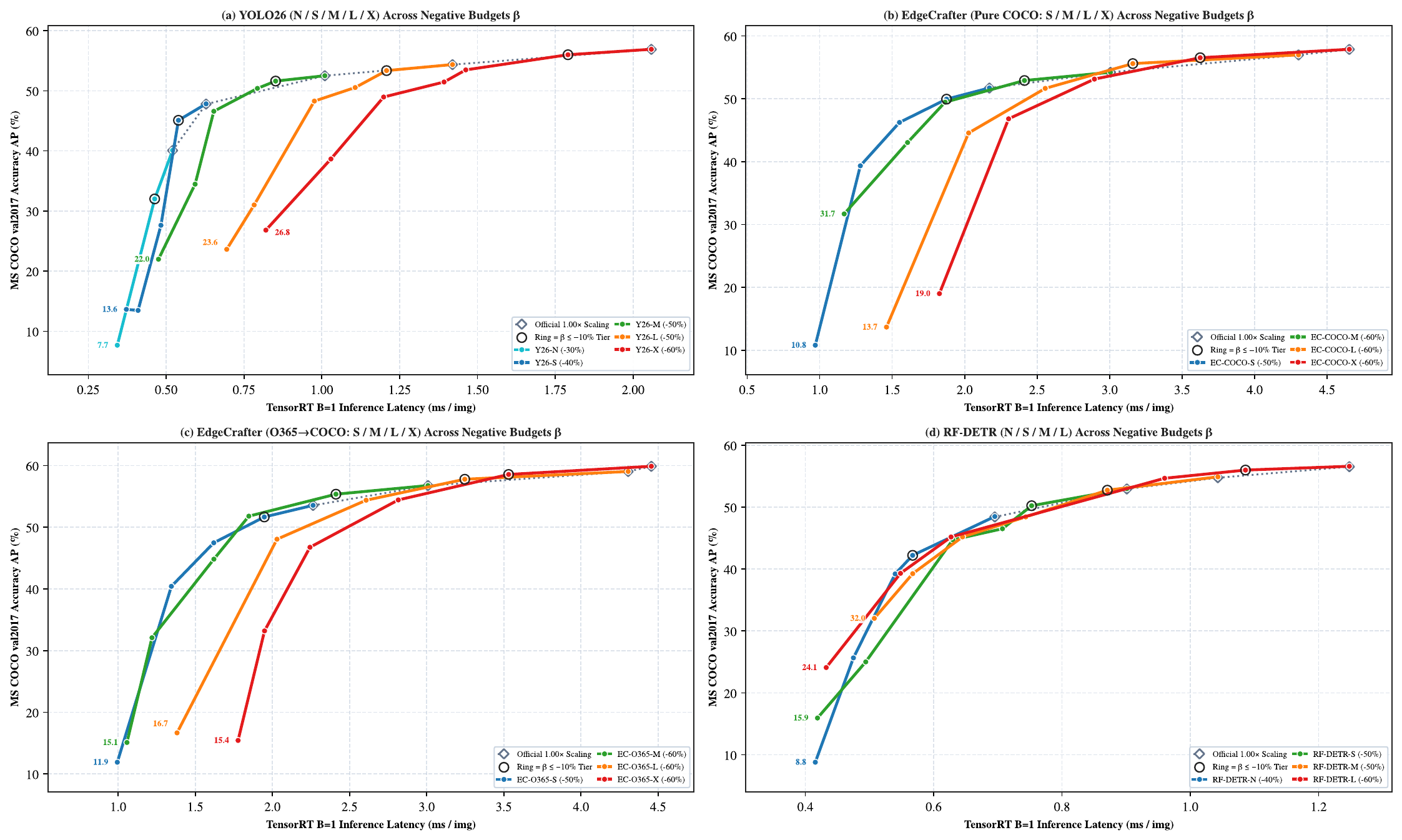}
    \vspace{-0.6em}
    \caption{\textbf{Per-scale accuracy--latency contraction trajectories across all 3 architectures (4 checkpoint tracks, 17 scales) under negative latency budgets ($\beta \in \{0\%, -10\%, -20\%, \dots, -100\%\}$), expanding \cref{tab:tta_vs_mem}.} X-axis: streaming \texttt{TensorRT~11.4 FP16} ($B{=}1$, +\texttt{CUDAGraph}) latency (ms/img); Y-axis: MS~COCO \texttt{val2017} AP (\%). \textbf{(a)} YOLO26 (\texttt{N, S, M, L, X}). \textbf{(b)} EdgeCrafter --- Pure COCO (\texttt{S, M, L, X}). \textbf{(c)} EdgeCrafter --- Objects365$\to$COCO (\texttt{S, M, L, X}). \textbf{(d)} RF-DETR (\texttt{N, S, M, L}). Gray dotted curves connect official $1.00\times$ checkpoints across scales; colored curves trace holdout-selected closed-form bridged subgraph contraction across $10\%$ negative budget steps (black rings mark $\beta \le -10\%$). Collapsed ultra-shallow / $0$-decoder / null regimes ($\text{AP} < 5.0\%$) are omitted from the plot to preserve scale readability; all 11 tiers ($0\%$ to $-100\%$) are reported in full in \cref{tab:app_neg_all17_steps}.}
    \label{fig:app_neg_arch_scale_lines}
    \vspace{-0.3em}
\end{figure*}
\null\newpage\null\newpage

\subsection{Negative-Budget Subgraph Contraction Continuum \& Per-Detector Breakdown}
\label{sec:app_neg_continuum}

Directly expanding the $\beta \le -10\%$ row of \cref{tab:tta_vs_mem}, \cref{fig:app_neg_arch_scale_lines} (above) and \cref{tab:app_neg_all17_steps,tab:app_neg_continuum_ablation} evaluate zero-gradient subgraph contraction across all 17 detectors and 11 negative steps ($\beta \in \{0\%, -10\%, \dots, -100\%\}$):
\begin{enumerate}
    \item \textbf{Per-Scale Non-Collapsed Trajectories (\cref{fig:app_neg_arch_scale_lines}, \cref{tab:app_neg_all17_steps}):} \Cref{fig:app_neg_arch_scale_lines} plots streaming \texttt{TensorRT} $B{=}1$ latency vs.\ COCO AP across YOLO26, EdgeCrafter (Pure COCO and O365$\to$COCO), and RF-DETR (omitting $\text{AP} < 5.0\%$ regimes; all 11 steps in \cref{tab:app_neg_all17_steps}). Compute-dense scales contract smoothly ($\beta \le -50\%$: \texttt{yolo26x} $38.65$~AP in $1.03$~ms, \texttt{ecdet\_x} $46.82$~AP in $2.30$~ms, \texttt{rfdetr\_l} $39.31$~AP in $0.55$~ms), whereas sub-ms scales (\texttt{yolo26n} $0.52$~ms, \texttt{rfdetr\_n} $0.70$~ms) hit sequential launch floors and switch to layer bypass.
    \item \textbf{Hardware Scaling Regimes Across $B{=}1$, Batched $B{=}8$ \& PyTorch $B{=}1$ (\cref{tab:app_neg_continuum_ablation}):} Solving $T_{\text{mode}}(g) \le (1+\beta)T_{0,\text{mode}}$ per runtime reveals Amdahl's law: at $\beta \le -10\%$, streaming \texttt{CUDAGraph} $B{=}1$ cuts 17-model mean latency by $-17.3\%$ ($2.22 \to 1.78$~ms) while retaining $95.6\%$ ID AP ($51.62$~AP) and $98.6\%$ COCO-O mAP ($33.48$~mAP); batched \texttt{CUDAGraph} $B{=}8$ retains $48.84$~AP ($90.7\%$) at $-30\%$, $44.37$~AP ($82.3\%$) at $-50\%$, and $31.05$~AP ($56.7\%$) at $-70\%$.
    \item \textbf{Cross-Scale Dominance: Reduced-Budget Second-Smallest Models Surpass Smallest Models:} When reducing budgets without retraining, \emph{every} family's second-smallest model (\texttt{yolo26s}, \texttt{ecdet\_m}, \texttt{ecdet\_o3652coco\_m}, \texttt{rfdetr\_s}) offers contracted options surpassing its native $1.00\times$ smallest model (\texttt{yolo26n}, \texttt{ecdet\_s}, \texttt{ecdet\_o3652coco\_s}, \texttt{rfdetr\_n}) in \emph{both} speed and AP. Under PyTorch $B{=}1$, all 4/4 tracks achieve dual dominance: \texttt{yolo26s} at $-10\%$ ($4.04$~ms, $47.41$~AP) and $-30\%$ ($3.97$~ms, $45.09$~AP) beats \texttt{yolo26n} ($5.39$~ms, $40.10$~AP; up to $26.4\%$ faster, $+7.31$~AP); \texttt{ecdet\_m} at $-10\%$ ($8.23$~ms, $53.45$~AP) and $-20\%$ ($7.21$~ms, $52.55$~AP) beats \texttt{ecdet\_s} ($9.86$~ms, $51.73$~AP); \texttt{ecdet\_o3652coco\_m} at $-10\%$ ($8.46$~ms, $56.01$~AP) and $-20\%$ ($7.33$~ms, $55.16$~AP) beats \texttt{ecdet\_o3652coco\_s} ($9.45$~ms, $53.55$~AP); and \texttt{rfdetr\_s} at $-10\%$ ($5.23$~ms, $49.49$~AP) beats \texttt{rfdetr\_n} ($5.59$~ms, $48.49$~AP). Under batched \texttt{TRT+CUDAGraph} $B{=}8$, \texttt{yolo26s} ($-40\%$: $0.11 < 0.15$~ms, $40.80 > 40.10$~AP), \texttt{ecdet\_m} ($-10\%$: $1.71 < 1.83$~ms, $52.91 > 51.73$~AP), and \texttt{ecdet\_o3652coco\_m} ($-10\%$: $1.71 < 1.84$~ms, $55.36 > 53.55$~AP) likewise beat their smallest models in both speed and AP.
    \item \textbf{Closed-Form \texttt{MM-FW-Ridge} Subgraph Bridges vs.\ Naive Bypass (\cref{tab:app_neg_continuum_ablation}):} Fitting \texttt{MM-FW-Ridge} bridges strictly on $\mathcal{D}_{\text{cal}}$ rescues $+5.99$~AP ID / $+5.38$~mAP OOD across 17 models on deep contraction and lifts YOLO26 PAN-neck bypass ($384\times 384$) from $1.37$ to $18.95$~AP mean ($+17.58$~AP; $4.55 \to 27.04$~AP on \texttt{yolo26x}).
\end{enumerate}

\begin{table*}[!t]
\caption{\textbf{Complete per-detector $10\%$-step negative-budget contraction continuum ($\beta \in \{0\%, -10\%, -20\%, \dots, -100\%\}$) across all 17 checkpoints and 4 detector families}, expanding \cref{tab:tta_vs_mem} and providing the complete numerical coordinates (\texttt{TRT B=1 ms / MS~COCO AP}) underlying \cref{fig:app_neg_arch_scale_lines} (including extreme $-70\%$--$-100\%$ regimes omitted from the plot; \textbf{0\%} reports the $300$-slot single-head anchor $n_{00}$, and the rightmost column reports compute-bound batched \texttt{CUDAGraph} $B{=}8$ at $\beta \le -50\%$).}
\label{tab:app_neg_all17_steps}
\centering
\setlength{\tabcolsep}{2.8pt}
\resizebox{\textwidth}{!}{\begin{tabular}{@{}l|ccccccccccc|c@{}}
\toprule
\textbf{Detector} & \textbf{0\%} & \textbf{-10\%} & \textbf{-20\%} & \textbf{-30\%} & \textbf{-40\%} & \textbf{-50\%} & \textbf{-60\%} & \textbf{-70\%} & \textbf{-80\%} & \textbf{-90\%} & \textbf{-100\%} & \textbf{CG8 (-50\%)} \\
\midrule
\texttt{yolo26n} & 0.52/40.10 & \textbf{0.46}/32.01 & 0.34/7.67 & 0.34/7.67 & 0.17/0.16 & \textbf{0.17}/0.16 & 0.17/0.16 & 0.15/0.18 & 0.08/0.04 & 0.00/0.00 & \textbf{0.00}/0.00 & \textbf{0.07}/32.01 \\
\texttt{yolo26s} & 0.63/47.83 & \textbf{0.54}/45.09 & 0.48/27.62 & 0.41/13.46 & 0.37/13.64 & \textbf{0.17}/0.33 & 0.17/0.33 & 0.17/0.33 & 0.10/0.05 & 0.00/0.00 & \textbf{0.00}/0.00 & \textbf{0.10}/38.26 \\
\texttt{yolo26m} & 1.01/52.51 & \textbf{0.85}/51.61 & 0.79/50.40 & 0.65/46.59 & 0.59/34.46 & \textbf{0.48}/21.98 & 0.29/0.94 & 0.29/0.94 & 0.11/0.06 & 0.08/0.01 & \textbf{0.00}/0.00 & \textbf{0.19}/46.59 \\
\texttt{yolo26l} & 1.42/54.36 & \textbf{1.21}/53.36 & 1.11/50.54 & 0.98/48.28 & 0.78/31.01 & \textbf{0.69}/23.63 & 0.40/0.94 & 0.40/0.94 & 0.11/0.05 & 0.11/0.05 & \textbf{0.00}/0.00 & \textbf{0.26}/48.28 \\
\texttt{yolo26x} & 2.06/56.90 & \textbf{1.79}/55.99 & 1.46/53.47 & 1.39/51.46 & 1.20/48.96 & \textbf{1.03}/38.65 & 0.82/26.82 & 0.47/1.06 & 0.35/0.80 & 0.10/0.08 & \textbf{0.00}/0.00 & \textbf{0.45}/51.46 \\
\rowcolor{rowgray}
\textit{YOLO26 Mean} & 1.13/50.34 & \textbf{0.97}/47.61 & \textbf{0.84}/37.94 & \textbf{0.75}/33.49 & \textbf{0.62}/25.64 & \textbf{0.51}/16.95 & \textbf{0.37}/5.84 & \textbf{0.30}/0.69 & \textbf{0.15}/0.20 & \textbf{0.06}/0.03 & \textbf{0.00}/0.00 & \textbf{0.22}/43.32 \\
\midrule
\texttt{ecdet\_s} & 2.17/51.73 & \textbf{1.88}/49.95 & 1.55/46.24 & 1.28/39.34 & 1.28/39.34 & \textbf{0.97}/10.80 & 0.73/0.71 & 0.57/0.07 & 0.14/0.00 & 0.14/0.00 & \textbf{0.00}/0.00 & \textbf{0.81}/46.24 \\
\texttt{ecdet\_m} & 3.01/54.22 & \textbf{2.41}/52.91 & 1.87/49.47 & 1.87/49.47 & 1.61/43.05 & \textbf{1.17}/31.69 & 1.17/31.69 & 0.85/1.56 & 0.12/0.00 & 0.12/0.00 & \textbf{0.00}/0.00 & \textbf{1.12}/49.47 \\
\texttt{ecdet\_l} & 4.30/56.99 & \textbf{3.16}/55.60 & 3.16/55.60 & 2.56/51.66 & 2.56/51.66 & \textbf{2.03}/44.57 & 1.46/13.68 & 1.07/1.92 & 0.77/0.09 & 0.00/0.00 & \textbf{0.00}/0.00 & \textbf{1.19}/44.57 \\
\texttt{ecdet\_x} & 4.65/57.88 & \textbf{3.62}/56.53 & 3.62/56.53 & 2.89/53.14 & 2.30/46.82 & \textbf{2.30}/46.82 & 1.83/19.00 & 1.23/2.29 & 0.87/0.24 & 0.00/0.00 & \textbf{0.00}/0.00 & \textbf{1.34}/46.82 \\
\rowcolor{rowgray}
\textit{ECDet-COCO Mean} & 3.53/55.21 & \textbf{2.77}/53.75 & \textbf{2.55}/51.96 & \textbf{2.15}/48.40 & \textbf{1.94}/45.22 & \textbf{1.62}/33.47 & \textbf{1.30}/16.27 & \textbf{0.93}/1.46 & \textbf{0.47}/0.08 & \textbf{0.06}/0.00 & \textbf{0.00}/0.00 & \textbf{1.11}/46.77 \\
\midrule
\texttt{ecdet\_o3652coco\_s} & 2.26/53.55 & \textbf{1.95}/51.66 & 1.62/47.47 & 1.35/40.41 & 1.35/40.41 & \textbf{0.99}/11.90 & 0.76/1.03 & 0.59/0.08 & 0.15/0.00 & 0.15/0.00 & \textbf{0.00}/0.00 & \textbf{0.81}/47.47 \\
\texttt{ecdet\_o3652coco\_m} & 3.01/56.76 & \textbf{2.41}/55.36 & 1.85/51.79 & 1.85/51.79 & 1.62/44.82 & \textbf{1.22}/32.08 & 1.06/15.10 & 0.85/1.60 & 0.12/0.00 & 0.12/0.00 & \textbf{0.00}/0.00 & \textbf{1.13}/51.79 \\
\texttt{ecdet\_o3652coco\_l} & 4.31/59.04 & \textbf{3.25}/57.78 & 3.25/57.78 & 2.61/54.38 & 2.03/48.06 & \textbf{2.03}/48.06 & 1.38/16.66 & 1.10/2.34 & 0.79/0.12 & 0.06/0.00 & \textbf{0.00}/0.00 & \textbf{1.19}/48.06 \\
\texttt{ecdet\_o3652coco\_x} & 4.46/59.88 & \textbf{3.53}/58.56 & 3.53/58.56 & 2.82/54.43 & 2.24/46.75 & \textbf{1.95}/33.19 & 1.78/15.43 & 1.24/1.74 & 0.87/0.08 & 0.14/0.00 & \textbf{0.00}/0.00 & \textbf{1.33}/46.75 \\
\rowcolor{rowgray}
\textit{ECDet-O365$\to$C Mean} & 3.51/57.31 & \textbf{2.78}/55.84 & \textbf{2.56}/53.90 & \textbf{2.15}/50.25 & \textbf{1.81}/45.01 & \textbf{1.55}/31.31 & \textbf{1.24}/12.05 & \textbf{0.94}/1.44 & \textbf{0.48}/0.05 & \textbf{0.12}/0.00 & \textbf{0.00}/0.00 & \textbf{1.11}/48.52 \\
\midrule
\texttt{rfdetr\_n} & 0.70/48.49 & \textbf{0.57}/42.21 & 0.54/39.21 & 0.47/25.65 & 0.42/8.79 & \textbf{0.31}/0.35 & 0.27/0.12 & 0.18/0.01 & 0.10/0.00 & 0.00/0.00 & \textbf{0.00}/0.00 & \textbf{0.11}/30.27 \\
\texttt{rfdetr\_s} & 0.90/52.99 & \textbf{0.75}/50.23 & 0.71/46.51 & 0.63/44.72 & 0.49/25.00 & \textbf{0.42}/15.93 & 0.28/0.04 & 0.24/0.01 & 0.11/0.00 & 0.00/0.00 & \textbf{0.00}/0.00 & \textbf{0.13}/35.87 \\
\texttt{rfdetr\_m} & 1.04/54.90 & \textbf{0.87}/52.74 & 0.74/48.43 & 0.65/45.20 & 0.57/39.26 & \textbf{0.51}/32.04 & 0.29/0.08 & 0.29/0.08 & 0.19/0.00 & 0.00/0.00 & \textbf{0.00}/0.00 & \textbf{0.19}/45.20 \\
\texttt{rfdetr\_l} & 1.25/56.60 & \textbf{1.09}/55.99 & 0.96/54.67 & 0.63/45.18 & 0.63/45.18 & \textbf{0.55}/39.31 & 0.43/24.09 & 0.28/0.13 & 0.24/0.02 & 0.11/0.00 & \textbf{0.00}/0.00 & \textbf{0.20}/45.18 \\
\rowcolor{rowgray}
\textit{RF-DETR Mean} & 0.97/53.25 & \textbf{0.82}/50.29 & \textbf{0.74}/47.20 & \textbf{0.59}/40.19 & \textbf{0.53}/29.56 & \textbf{0.45}/21.91 & \textbf{0.32}/6.08 & \textbf{0.25}/0.06 & \textbf{0.16}/0.01 & \textbf{0.03}/0.00 & \textbf{0.00}/0.00 & \textbf{0.16}/39.13 \\
\midrule
\rowcolor{rowgray}
\textbf{17-Model Mean} & \textbf{2.22}/53.81 & \textbf{1.78}/51.62 & \textbf{1.62}/47.17 & \textbf{1.37}/42.52 & \textbf{1.19}/35.73 & \textbf{1.00}/25.38 & \textbf{0.78}/9.81 & \textbf{0.59}/0.90 & \textbf{0.31}/0.09 & \textbf{0.07}/0.01 & \textbf{0.00}/0.00 & \textbf{0.62}/44.37 \\
\bottomrule
\end{tabular}}
\end{table*}

\begin{table*}[!t]
\caption{\textbf{Multi-regime 11-tier negative-budget contraction continuums ($\beta \in \{0\%, -10\%, \dots, -100\%\}$) across all 17 detectors}, comparing streaming \texttt{CUDAGraph} $B{=}1$, compute-bound batched \texttt{CUDAGraph} $B{=}8$, and standalone PyTorch $B{=}1$ (each independently constrained by $T_{\text{mode}}(g) \le (1+\beta)T_{0,\text{mode}}$ on $\mathcal{D}_{\text{hold}}$; bottom block isolates closed-form \texttt{MM-FW-Ridge} subgraph bridges vs.\ unbridged naive layer bypass across all 17 models).}
\label{tab:app_neg_continuum_ablation}
\centering
\resizebox{\textwidth}{!}{\begin{tabular}{@{}l|cccc|cccc|cccc@{}}
\toprule
& \multicolumn{4}{c|}{\textbf{Streaming TRT+\texttt{CUDAGraph} ($B{=}1$)}} & \multicolumn{4}{c|}{\textbf{Batched TRT+\texttt{CUDAGraph} ($B{=}8$)}} & \multicolumn{4}{c}{\textbf{Standalone PyTorch ($B{=}1$)}} \\
\textbf{Budget $\beta$} & \textbf{ms} & \textbf{$\overline{\Delta}$Time} & \textbf{ID / OOD} & \textbf{Ret.\ ID\%} & \textbf{ms} & \textbf{$\overline{\Delta}$Time} & \textbf{ID / OOD} & \textbf{Ret.\ ID\%} & \textbf{ms} & \textbf{$\overline{\Delta}$Time} & \textbf{ID / OOD} & \textbf{Ret.\ ID\%} \\
\midrule
\textbf{0\%} & \textbf{2.22} & $+0.0\%$ & 53.81/33.86 & 100.0\% & \textbf{1.62} & $+0.0\%$ & 53.81/33.86 & 100.0\% & 8.07 & $+0.0\%$ & 53.81/33.86 & 100.0\% \\
\rowcolor{rowgray}
\textbf{-10\%} & \textbf{1.78} & $-17.3\%$ & 51.62/33.48 & 95.6\% & \textbf{1.19} & $-23.3\%$ & 52.40/33.84 & 97.3\% & 6.47 & $-20.7\%$ & 51.51/32.57 & 95.5\% \\
\textbf{-20\%} & \textbf{1.62} & $-26.6\%$ & 47.17/30.90 & 86.2\% & \textbf{1.17} & $-27.1\%$ & 51.82/33.58 & 96.2\% & 5.92 & $-26.6\%$ & 50.07/31.73 & 92.8\% \\
\textbf{-30\%} & \textbf{1.37} & $-37.1\%$ & 42.52/27.87 & 77.3\% & \textbf{0.83} & $-45.0\%$ & 48.84/31.30 & 90.7\% & 4.58 & $-43.2\%$ & 21.72/14.19 & 41.5\% \\
\textbf{-40\%} & \textbf{1.19} & $-46.1\%$ & 35.73/23.46 & 64.4\% & \textbf{0.80} & $-51.0\%$ & 47.20/30.43 & 87.4\% & 4.39 & $-46.4\%$ & 16.71/11.10 & 31.8\% \\
\rowcolor{rowgray}
\textbf{-50\%} & \textbf{1.00} & $-56.1\%$ & 25.38/16.68 & 45.1\% & \textbf{0.62} & $-58.6\%$ & 44.37/28.42 & 82.3\% & 1.44 & $-78.9\%$ & 2.57/1.33 & 6.1\% \\
\textbf{-60\%} & \textbf{0.78} & $-66.1\%$ & 9.81/6.35 & 17.3\% & \textbf{0.52} & $-67.3\%$ & 39.16/25.41 & 71.7\% & 1.29 & $-81.2\%$ & 2.36/1.23 & 5.8\% \\
\textbf{-70\%} & \textbf{0.59} & $-73.4\%$ & 0.90/0.50 & 1.6\% & \textbf{0.38} & $-75.2\%$ & 31.05/19.83 & 56.7\% & 0.89 & $-87.7\%$ & 0.64/0.38 & 1.5\% \\
\textbf{-80\%} & \textbf{0.31} & $-86.8\%$ & 0.09/0.07 & 0.2\% & \textbf{0.24} & $-84.7\%$ & 15.40/9.68 & 27.6\% & 0.58 & $-92.2\%$ & 0.02/0.02 & 0.1\% \\
\textbf{-90\%} & \textbf{0.07} & $-96.8\%$ & 0.01/0.01 & 0.0\% & \textbf{0.14} & $-91.6\%$ & 0.42/0.35 & 0.7\% & 0.40 & $-95.3\%$ & 0.01/0.01 & 0.0\% \\
\rowcolor{rowgray}
\textbf{-100\%} & \textbf{0.00} & $-100.0\%$ & 0.00/0.00 & 0.0\% & \textbf{0.00} & $-100.0\%$ & 0.00/0.00 & 0.0\% & 0.00 & $-100.0\%$ & 0.00/0.00 & 0.0\% \\
\midrule
\multicolumn{5}{@{}l|}{\textbf{Bridge Ablation A (17-Model Deep Spatial / Partial-Dec):} Naive vs.\ \texttt{MM-FW-Ridge}} & \multicolumn{2}{c|}{Naive: 19.42 AP} & \multicolumn{2}{c|}{\texttt{MM-FW-Ridge}: \pos{25.42} AP} & \multicolumn{4}{c}{Gain: \pos{$+5.99$ ID / $+5.38$ OOD}} \\
\multicolumn{5}{@{}l|}{\textbf{Bridge Ablation B (17-Model Mid-Trunk Layer Bypass):} Naive vs.\ \texttt{MM-FW-Ridge}} & \multicolumn{2}{c|}{Naive: 0.57 AP} & \multicolumn{2}{c|}{\texttt{MM-FW-Ridge}: \pos{6.37} AP} & \multicolumn{4}{c}{Gain: \pos{$+5.80$ ID / $+2.91$ OOD}} \\
\multicolumn{5}{@{}l|}{\textbf{Bridge Ablation C (YOLO26 PAN-Neck Bypass @ 384):} Naive vs.\ \texttt{MM-FW-Ridge}} & \multicolumn{2}{c|}{Naive: 1.37 AP} & \multicolumn{2}{c|}{\texttt{MM-FW-Ridge}: \pos{18.95} AP} & \multicolumn{4}{c}{Gain: \pos{$+17.58$ ID / $+9.04$ OOD}} \\
\bottomrule
\end{tabular}}
\end{table*}
\appnextpage

\begin{table*}[!t]
\caption{\textbf{Effect of pairing our Memory-Graph (\methodDyn / \methodplus) with external post-processing (NMS, Soft-NMS~\cite{bodla2017softnms}, Canonical WBF~\cite{solovyev2021wbf}, and Score-Gated Hybrid WBF-NMS) on MS~COCO~\cite{lin2014coco} \texttt{val2017} AP} across 8 representative detectors spanning YOLO26~\cite{jocher2026yolo26}, EdgeCrafter~\cite{liu2026edgecrafter}, and RF-DETR~\cite{robinson2026rfdetr}.}
\label{tab:app_mem_nms_wbf}
\centering
\resizebox{\textwidth}{!}{\begin{tabular}{@{}l|cc|cc|ccc|c@{}}
\toprule
& \multicolumn{2}{c|}{\textbf{Reference Baselines}} & \multicolumn{2}{c|}{\textbf{Regime A: Shared-Query Mem}} & \multicolumn{3}{c|}{\textbf{Regime B: Unbound / Multi-View Mem + Post-Proc}} & \textbf{In-Graph Full} \\
\textbf{Detector} & \textbf{Official ($1\times$)} & \textbf{Official TTA (NMS)} & \textbf{Compact ($0$-Post)} & \textbf{+Post-NMS ($.65$)} & \textbf{Unbound + WBF} & \textbf{Unbound + NMS} & \textbf{Best Hybrid / Flip-NMS} & \textbf{\methodplus ($0$-Post)} \\
\midrule
\texttt{yolo26s} & 47.79 & 49.96 & 48.88 & \negv{48.65} ($-0.23$) & \pos{49.84} (\texttt{o2o}-6v) & \pos{49.96} (\texttt{o2o}-6v) & \pos{49.87} (\texttt{o2m} Hybrid) & 49.89 \\
\texttt{yolo26l} & 54.35 & 55.49 & 54.95 & \negv{54.67} ($-0.28$) & \pos{55.58} (\texttt{o2o}-6v) & \pos{55.56} (\texttt{o2m}-6v) & \pos{55.57} (\texttt{o2m} Hybrid) & 55.45 \\
\midrule
\texttt{ecdet\_s} (COCO) & 51.71 & 52.53 & 52.26 & \negv{52.07} ($-0.19$) & \negv{50.72} ($-1.54$) & \pos{52.49} & \pos{52.74} ($2\times$Flip + NMS$.70$) & \pos{53.25} \\
\texttt{ecdet\_x} (COCO) & 57.86 & 58.39 & 58.44 & \negv{58.22} ($-0.22$) & \negv{54.82} ($-3.62$) & 58.30 & \pos{58.58} ($2\times$Flip + NMS$.70$) & \pos{59.19} \\
\texttt{ecdet\_o3652coco\_s} & 53.53 & 54.19 & 53.97 & \negv{53.80} ($-0.17$) & \negv{49.00} ($-4.97$) & \pos{54.13} & 53.44 (Unbound Hybrid) & \pos{55.03} \\
\texttt{ecdet\_o3652coco\_x} & 59.87 & 60.37 & 60.34 & \negv{60.08} ($-0.26$) & \negv{55.49} ($-4.85$) & 60.31 & 59.83 (Unbound Hybrid) & \pos{61.10} \\
\midrule
\texttt{rfdetr\_n} ($384$) & 48.41 & 51.18 & 53.15 & \negv{53.04} ($-0.11$) & \negv{52.69} ($-0.46$) & 52.91 & \pos{53.38} (13-Grp Hybrid WBF-NMS) & \pos{54.63} \\
\texttt{rfdetr\_m} ($576$) & 54.75 & 56.10 & 57.80 & \negv{57.57} ($-0.23$) & \negv{57.25} ($-0.55$) & 57.63 & 57.68 (13-Grp Hybrid WBF-NMS) & \pos{58.68} \\
\bottomrule
\end{tabular}}
\end{table*}
\null\newpage\null\newpage

\subsection{Pairing Memory-Graph Routing with External Post-Processing (NMS / WBF)}
\label{sec:app_mem_postproc}

A natural question arising from \cref{sec:tta_vs_mem} is whether pairing our in-graph Memory-Graph routing (\methodDyn / \methodplus) with external NMS, Soft-NMS~\cite{bodla2017softnms}, or WBF~\cite{solovyev2021wbf} yields further improvements. \Cref{tab:app_mem_nms_wbf} evaluates eight representative detectors across all four tracks on MS~COCO \texttt{val2017} under two distinct regimes---(A)~appending post-hoc NMS or Gaussian Soft-NMS directly after a uniform uncalibrated shared-query/anchor Memory-Graph probe (fixing uniform branch weights without $\mathcal{D}_{\text{hold}}$ tuning so post-processing effects are isolated cleanly), and (B)~relaxing the single-view shared-query constraint so each multi-basin/scale/flip memory proposes independent candidates fused via canonical WBF, Batched NMS, or a score-gated \emph{Hybrid WBF-NMS} (NMS cluster selection at $\IoU=0.65$ followed by $s^2\IoU^{2.5}$-weighted box expectation over $\IoU \ge 0.72$ agreeing memories):

\begin{enumerate}
    \item \textbf{Post-Hoc NMS on Shared-Query Memory-Graph Universally Degrades Accuracy ($8/8$ Models, $-0.11$ to $-0.28$~AP):} Because shared-query (DETR) and peak-protected (YOLO26) memory routing already produces 1-to-1 duplicate-free predictions per instance, applying hard NMS ($\IoU=0.65$) deletes legitimate crowded/occluded same-class True Positives rather than false duplicates (and even Gaussian Soft-NMS dips $-0.01\text{--}-0.05$~AP).
    \item \textbf{In-Decoder Shared-Query Fusion Outperforms Box-Space WBF/NMS on High-Capacity DETRs:} Unbound canonical WBF degrades DETRs by $-0.46\text{--}-4.97$~AP due to Top-$300$ low-confidence tail query distortion, and even unbound 6-memory NMS on ECDet-X ($58.30$~AP) and ECDet (O365$\to$COCO)-X ($60.31$~AP) trails in-decoder shared-query \methodplus ($59.19$ and $61.10$~AP) by $0.89$ and $0.79$~AP, confirming that latent per-query trajectory fusion preserves substantially richer geometry than post-hoc box pruning.
    \item \textbf{Targeted Gains When Unlocking Multi-View / \texttt{o2m} Proposals in Recall-Limited Regimes:} Once external post-processing is permitted, relaxing single-view proposal bottlenecks improves three recall-limited regimes: (i)~on \textbf{YOLO26-S/L}, unlocking our Procrustes bi-basin ($W_{\text{O365-pre}}, W_0$) memories across 6 views (\texttt{o2o} / \texttt{o2m}) with WBF or NMS jumps $+0.63\text{--}+1.08$~AP over $0$-PostProc Compact Mem ($55.58$~AP on YOLO26-L, surpassing both Official TTA (NMS, $55.49$) and 6v-WBF ($55.56$~AP) at identical 6-view budget); (ii)~on low-resolution \textbf{RF-DETR-N} ($384\times 384$), unbound 13-Group multi-scale/flip proposals paired with score-gated \emph{Hybrid WBF-NMS} lift Compact Mem from $53.15$ to $\mathbf{53.38}$~\textbf{AP} ($+0.23$~AP; $+2.20$~AP over Official TTA [NMS]); and (iii)~on pure-COCO \textbf{ECDet-COCO-S}, fusing original and horizontal-flip shared-query Memory-Graphs via high-threshold NMS ($\IoU=0.70$) lifts AP from $52.26$ to $\mathbf{52.74}$~\textbf{AP} ($+0.48$~AP), outperforming Official TTA (NMS, $52.53$~AP), while zero-postprocessing \methodplus reaches $53.25$~AP.
\end{enumerate}

\begin{figure*}[!t]
    \centering
    \includegraphics[height=0.54\textheight,keepaspectratio]{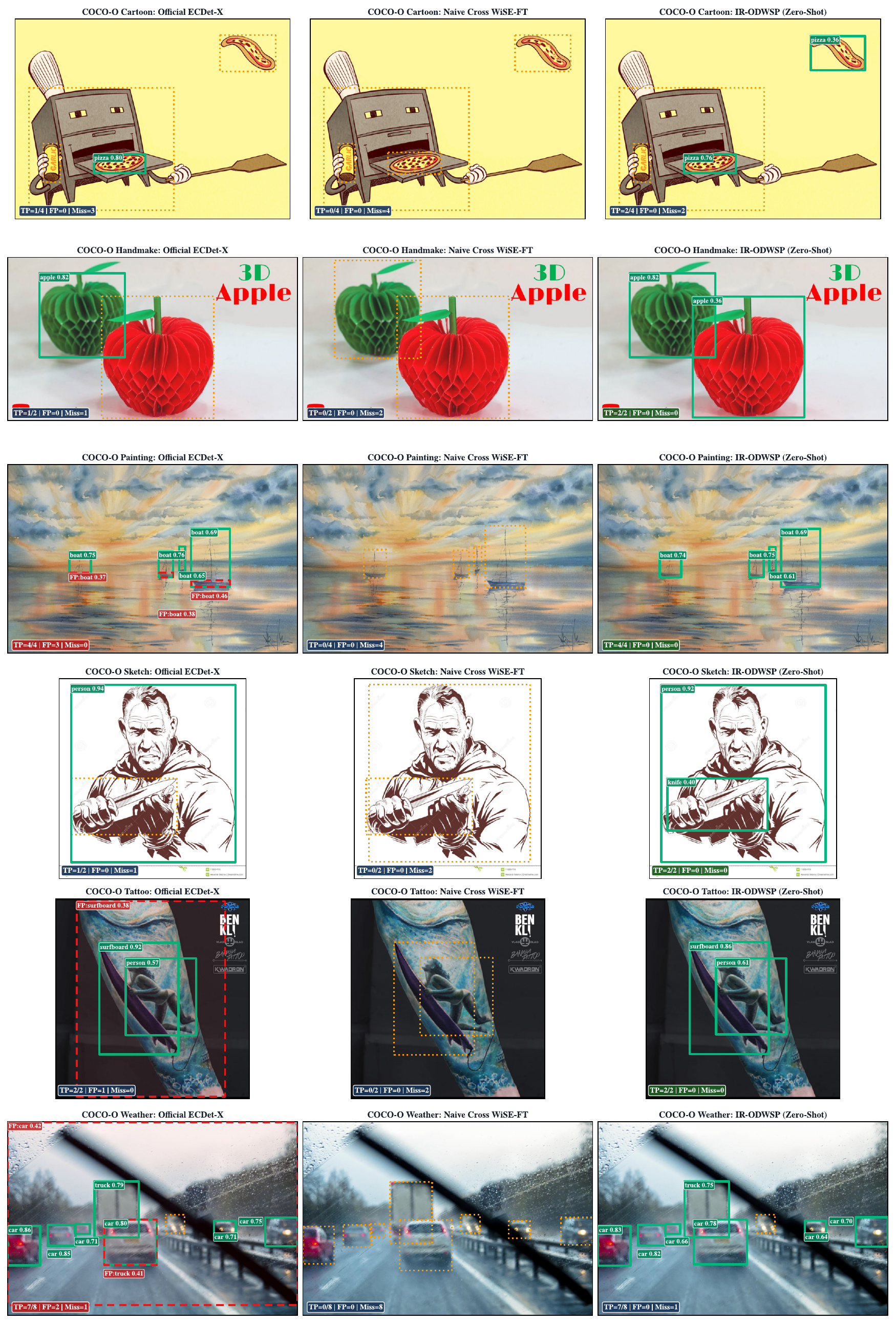}
    \vspace{-0.5em}
    \caption{\textbf{Extended qualitative gallery across all 6 natural OOD domains of COCO-O~\cite{mao2023cocoo}} (\emph{Cartoon, Handmake, Painting, Sketch, Tattoo, Weather}) on EdgeCrafter-X~\cite{liu2026edgecrafter} at $1.00\times$ deployment cost, illustrating cross-basin collapse under naive WiSE-FT~\cite{wortsman2022wiseft} vs.\ accurate zero-shot domain generalization under \method.}
    \label{fig:S1}
\end{figure*}
\appnextpage

\section{Extended Multi-Domain Qualitative Results}
\label{sec:app_qual}

\Cref{fig:S1} presents a six-row qualitative gallery spanning \textbf{all six natural distribution shifts of COCO-O} (\emph{Cartoon, Handmake, Painting, Sketch, Tattoo, Weather}), comparing Official EdgeCrafter-X, Naive Cross-Model WiSE-FT ($\alpha=0.85$), and Symmetry-Aligned Cross-Model \method ($1.00\times$ inference speed). Across every domain, naive Euclidean interpolation across disjoint weight basins ($\cos=0.340$) suffers severe activation cancellation and misses primary foreground objects, whereas \method aligns channel variances and orthogonal bases without gradients to recover missed stylized objects and suppress spurious background detections. Specifically, across \emph{Cartoon} and \emph{Sketch} monochrome contours, \emph{Handmake} craft geometries, \emph{Painting} impressionist brushwork, \emph{Tattoo} skin-ink occlusions, and \emph{Weather} highway snow and glare, folded $1.00\times$ \method preserves tight box boundaries and high-confidence true positives without retraining weights or external post-hoc NMS.
\vfill\eject

\section{Limitations}
\label{sec:app_limitations}

While \method achieves $100\%$ non-regression and strictly positive blind gains across all 26 single- and cross-model detector settings at $1.00\times$ deployment cost ($\beta = 0\%$, $2.22$~ms \texttt{CUDAGraph} at $B{=}1$), the absolute AP gain of pure single-checkpoint $1.00\times$ collapse on already heavily converged detectors is naturally bounded ($+0.01\text{--}+0.16$~AP overall, up to $+0.31$~$\text{AP}_s$), as algebraic re-collapse $\mathcal{C}$ cannot increase the representational rank or spatial resolution of a frozen backbone; unlocking larger gains without retraining weights ($+0.30$~AP already at $\beta \le +10\%$ in \methodDyn, $+1.05$~AP at $\beta \le +100\%$ in \methodSG/\methodAuto, and $+2.14$~AP in unconstrained \methodplus) trades off a user-customizable inference latency budget (\cref{sec:tta_vs_mem,sec:app_tta_audit,sec:app_graph_ablations}). Second, cross-model Orthogonal Procrustes alignment (\cref{eq:procrustes_align}) requires matching shared layer widths ($d_{\text{in}}$) between $W_{\text{aux}}$ and $W_0$; extending exact zero-gradient transport across heterogeneous backbone channel widths remains an open direction.
\end{document}